\documentclass[smallextended]{svjour3}       
\smartqed  
\usepackage{graphicx}
\usepackage[table]{xcolor}
\usepackage{bm}
\usepackage{amsmath, amssymb}
\usepackage{mathrsfs}  
\usepackage{array}
\usepackage{xcolor, soul}
\usepackage{enumerate}
\usepackage{multirow}
\usepackage{bbm}
\usepackage{color}
\usepackage{natbib}
\usepackage{graphicx,subcaption}
 \usepackage{verbatim}
\usepackage[utf8]{inputenc} 
\usepackage{hyperref}
\hypersetup{
    colorlinks=true,
    linkcolor=blue,
    filecolor=magenta,      
    urlcolor=cyan,
}

\usepackage{hhline}
\usepackage{float} 
\usepackage{booktabs}       
\usepackage{amsfonts}       
\usepackage{nicefrac}       
\usepackage{times}
\usepackage{mathptmx} 
\usepackage[ruled,vlined]{algorithm2e}
\usepackage{cellspace}
\usepackage{xcoffins}
\NewCoffin\tablecoffin

\usepackage[table]{xcolor}
\newcommand{\bl[1]}{{\bf #1}}
\newcommand{\bz}{\bl[z]}
\newcommand{\bxi}{\bm{\xi}}
\newcommand{\bgamma}{\bm{\gamma}}
\newcommand{\bx}{{\bf x}}
\newcommand{\Dmat}{W}
\newcommand{\reals}{\mathbb{R}}
\newcommand{\Domega}{D_\Omega}
\newcommand{\bv}{\bl[v]}
\newcommand{\bu}{\bl[u]}
\newcommand{\colvec}[2][.8]{%
  \scalebox{#1}{%
    \renewcommand{\arraystretch}{.8}%
    $\begin{bmatrix}#2\end{bmatrix}$%
  }
}

\begin{document}

\title{A unified framework for closed-form nonparametric regression, classification, preference and mixed problems with Skew Gaussian Processes}

\author{Alessio Benavoli  \and         Dario Azzimonti \and Dario Piga}


\institute{{Alessio Benavoli \at
School of Computer Science and Statistics, Trinity College,\\ Dublin, Ireland
              \email{alessio.benavoli@tcd.ie}           
           \and
           Dario Azzimonti and Dario Piga \at
           Dalle Molle Institute for Artificial Intelligence Research (IDSIA) - USI/SUPSI, Manno, Switzerland. 
}}

\date{Received: date / Accepted: date}

\maketitle

\begin{abstract}
Skew-Gaussian processes (SkewGPs)  extend the multivariate Unified Skew-Normal distributions over finite dimensional vectors to distribution over functions. SkewGPs are more general and flexible than   Gaussian processes, as SkewGPs may also   represent asymmetric distributions. In a   recent contribution we showed that SkewGP and probit likelihood are conjugate, which allows  us  to  compute the exact  posterior for non-parametric binary  classification and  preference learning. In this paper, we generalize previous results and we prove that SkewGP is conjugate with both the normal and affine probit likelihood, and  more in general, with their product. This allows us to (i) handle   classification, preference, numeric and ordinal  regression, and mixed problems in a unified framework; (ii) derive  closed-form expression for the  corresponding posterior distributions. We show empirically that the proposed framework  based on SkewGP  provides better performance than  Gaussian processes in active learning and Bayesian (constrained) optimization. These two tasks are fundamental for design of experiments and in Data Science.
 \end{abstract}

 
\section{Introduction}
\label{sec:intro}
Gaussian Processes (GPs) are powerful nonparametric distributions  over  functions.  For real-valued outputs, we can combine the GP prior with a Gaussian likelihood and perform exact posterior inference in closed form.  However, in other cases, such as classification, preference learning, ordinal regression and mixed problems, the likelihood is no longer conjugate to the GP prior and closed-form expression for the posterior is not available.

	In this paper, we show that is actually possible to derive closed-form expression for the posterior process in all the above cases (not only for regression), and that the posterior process is a Skew Gaussian Process (SkewGP), a stochastic process whose finite dimensional marginals follow a Skew-Normal distribution. 
	
	Consider, for example, a classification task with a probit likelihood and a GP prior on the latent function. We can show that the posterior is a SkewGP and that   the posterior latent function can be computed analytically on the training points. By exploiting the closure properties of SkewGPs, we can compute the distribution  of the latent function at any new test point. We can easily obtain posterior samples at any test point by exploiting an additive representation for Skew-Normal vectors which decomposes a Skew-Normal vector in a linear combination of a normal vector and a truncated-normal vector. This decomposition has two main advantages. First of all, it just requires normal and truncated-normal samples. Truncated-normal samples can be obtained with rejection-free Monte Carlo sampling by using the \emph{lin-ess} method \citep{gessner2019integrals}; this avoids the need of expensive Markov Chain Monte Carlo methods. Secondly, the closure of the SkewGP family requires only sampling of the posterior at the training points. These samples can then be reused for any new test point, thus greatly reducing the computational cost for predicting many test points Such scenario  comes up for instance in Bayesian optimization tasks as we explain in Section~\ref{sec:applications}. 

As prior class, SkewGPs are more general and more flexible nonparametric distributions than GPs, as SkewGPs may also   represent asymmetric distributions. Moreover, SkewGPs include GPs as a particular case. By exploiting the closed-form expression for the posterior and predictive distribution, we show that we can compute inferences for regression, classification, preference and mixed problems with computation complexity of 
$O(n^3)$ and storage demands of $O(n^2)$ (same as for GP regression).

This allows us to provide a uniﬁed framework for nonparametric inference for
a large class of likelihoods and, consequently, supervised learning problems,
as illustrated in Table \ref{tab:1}.



\begin{table*}[htp]
    \centering
    \resizebox{\columnwidth}{!}{%
\setlength{\tabcolsep}{5pt}
\renewcommand{\arraystretch}{3.2}  
    \begin{tabular}{|cc||cc|c|c|c|}

   \multicolumn{1}{c}{}  &\multicolumn{1}{c}{} &\multicolumn{1}{c}{} &\multicolumn{1}{c}{} & \multicolumn{1}{c}{\large \textbf{regression}~~~~} & \multicolumn{1}{c}{\large \textbf{preference}$|$\textbf{classification }} & \multicolumn{1}{c}{\large\textbf{mixed}~~~~~~~}\\
        \hline
 \multirow{3}{*}{\rotatebox[origin=c]{90}{\Large \textbf{Prior}}}&  \multirow{3}{*}{\begin{minipage}{5cm}
\includegraphics[width=5cm,trim={1cm 2.0cm 1cm 1.2cm},clip]{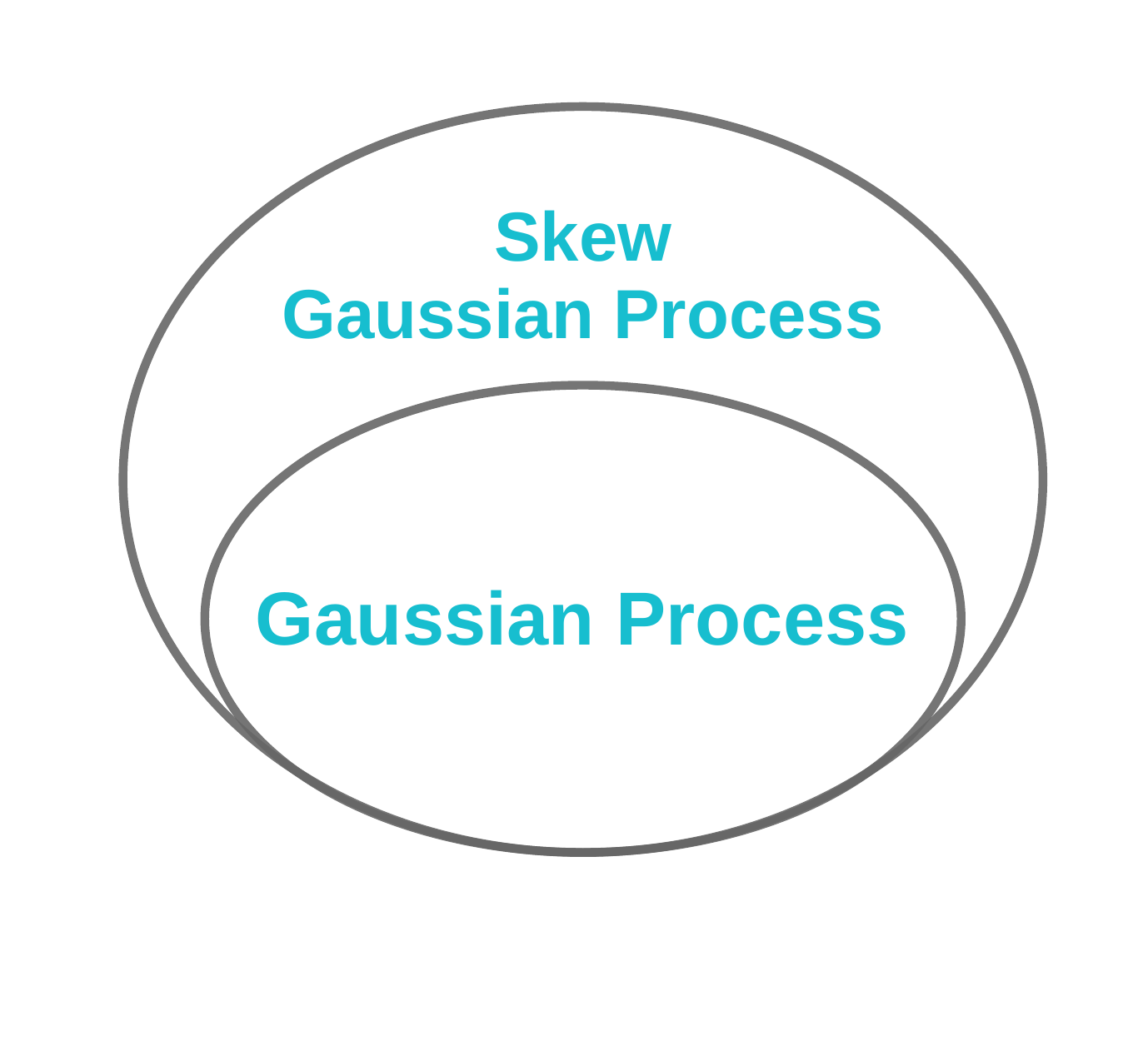}
 \end{minipage}
}~&
 \rotatebox[origin=c]{90}{\Large \textbf{Likelihood}}& & \raisebox{-.4\totalheight}{\includegraphics[width=4cm,trim={2cm 0 0 0},clip]{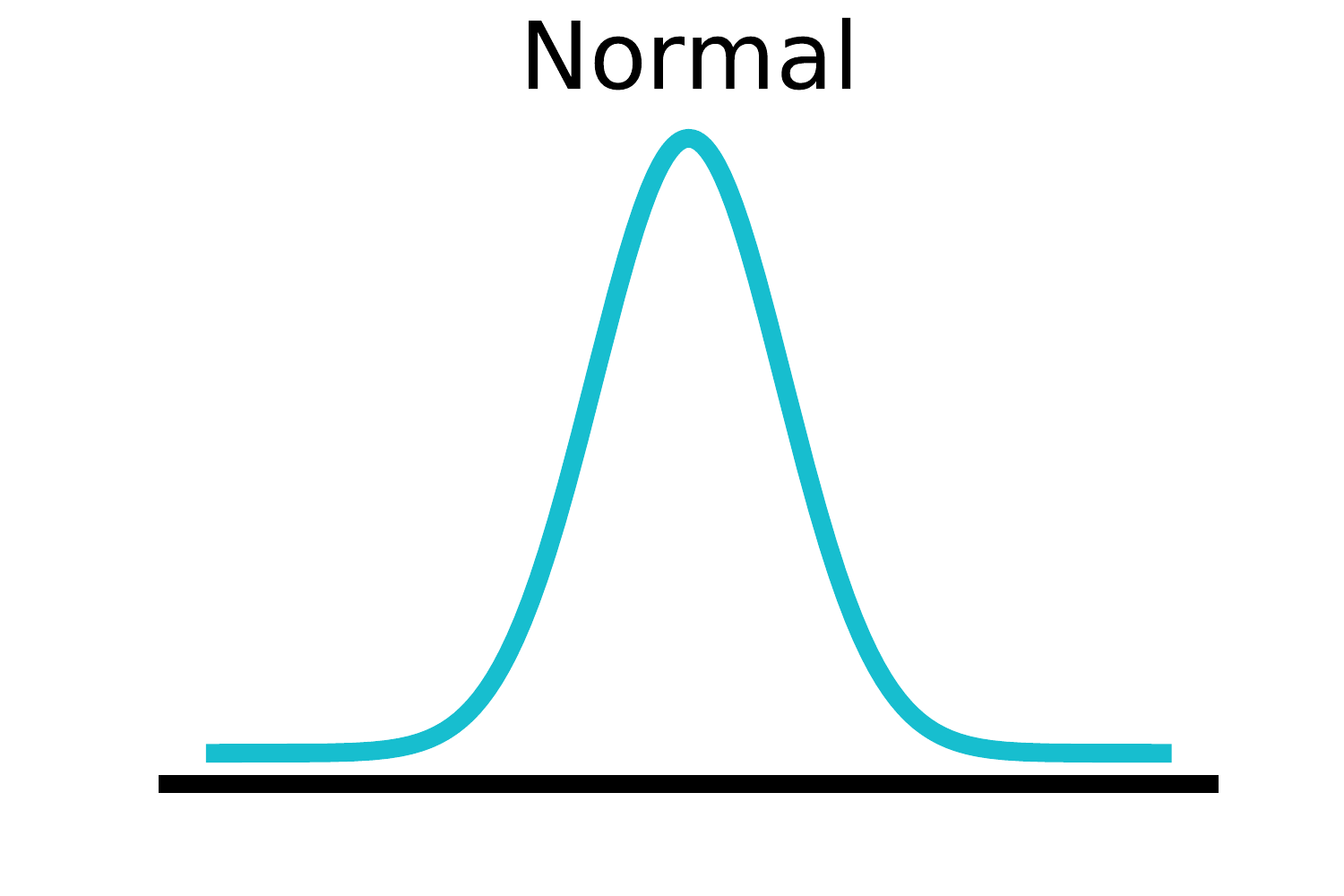}}
 &
  \raisebox{-.4\totalheight}{\includegraphics[width=4cm,trim={2cm 0 0 0},clip]{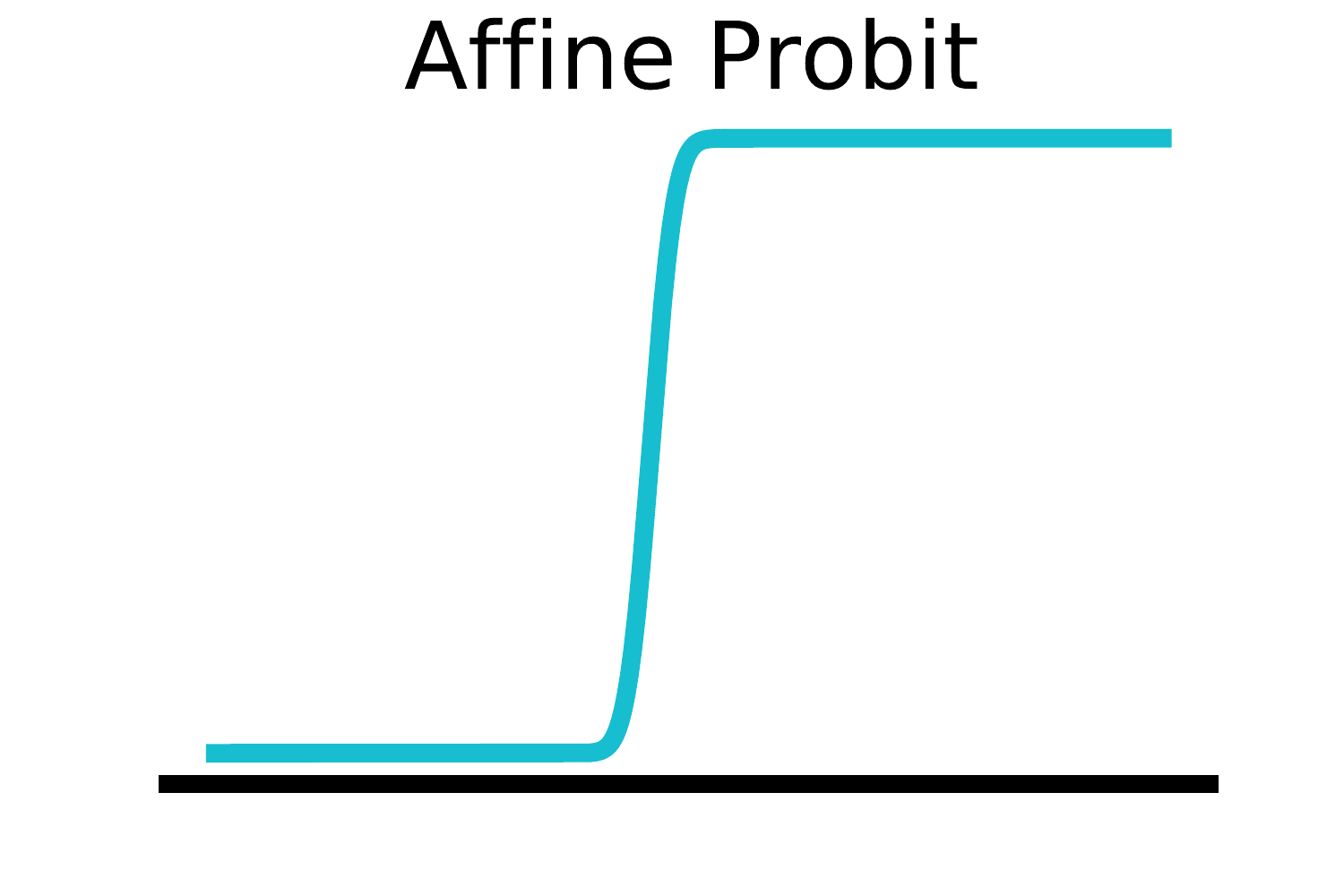}}&
   \raisebox{-.4\totalheight}{\includegraphics[width=4cm,trim={2cm 0 0 0},clip]{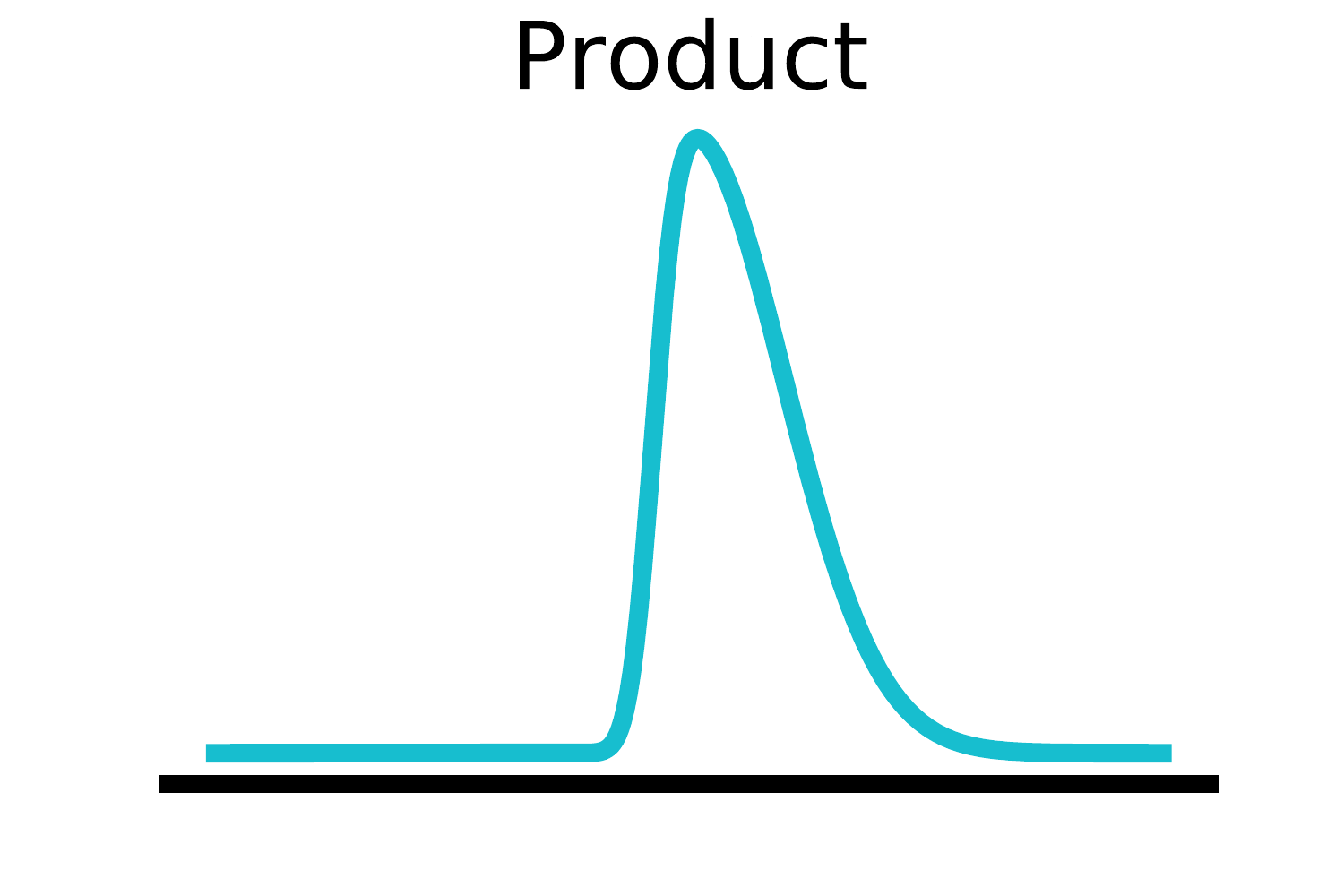}}\\
   \cline{3-7}
 & & \multirow{2}{*}{\rotatebox[origin=c]{90}{\Large \textbf{Posterior}}}& \multicolumn{1}{|c|}{\large Exact} & \large SkewGP\,$|$\,\large GP & \large SkewGP & \large SkewGP\\
 \cline{4-7}
 & & & \multicolumn{1}{|c|}{\large Approx.} & & \large GP & \large GP\\
 \hline
    \end{tabular}}
    \caption{SkewGPs are conjugate to a larger class of likelihoods.}
    \label{tab:1}
\end{table*}

\subsection{Different types of observations and likelihood models}
In supervised learning applications, we deal with the problem
of learning input-output mappings from data.
Consider a dataset consisting of $n$ samples.  Each of the samples is a pair of input vector $\bx_i \in \mathbb{R}^d$ and  output $y_i$. Depending on the type of the output variable, supervised learning problems can be  divided into different categories.

\paragraph{Numeric:} In the regression
setting the outputs are real values $y \in \mathbb{R}$ and
the input-output mapping is usually modelled as
$y_i=f(\bx_i)+v_i$, where $v_i$  is an additive independent, identically distributed Gaussian noise with zero mean and variance $\sigma_v^2$.
The likelihood model is  given by:
\begin{equation}
\label{eq:normlike1}
 p(y_i|f(\bx_i))=\phi\left(\frac{y_i-f(\bx_i)}{\sigma_v}\right),
\end{equation}
where $\phi(\cdot)$ is the PDF of the standard Normal distribution.

\paragraph{Binary:} In the binary classification
setting the outputs are categories that
we can be labeled as  $y_i = -1$ and $y_i = +1$, that is $y_i \in \{-1,1\}$.
The \textit{probit} likelihood model is: 
\begin{equation}
\label{eq:probitlike1}
 p(y_i|f(\bx_i))=\Phi(y_i f(\bx_i)),
\end{equation}
that is a Bernoulli distribution with probability
$\Phi(f(\bx_i))$, where $\Phi(\cdot)$ is the CDF of the standard Normal distribution.


\paragraph{Ordinal:} In ordinal regression, $y_i \in \{1,2,\dots,r\}$, where the integer $1,2,\dots,r$ are used to denote  order categories such as, for instance, movies' ratings.
In ordinal regression, we map these ordered categories into a partition of $\mathbb{R}$, that is $j \rightarrow (b_{j-1},b_{j}]$ with $\mathbb{R}=(-\infty,b_1]\cup(b_1,b_2]\cup\dots\cup(b_{r-1},\infty)$.
The likelihood can be modelled as an indicator function $p(y_i|f(\bx_i))=I_{[(b_{y_{i-1}},b_{y_{i}}]}(f(\bx_i))$.
 In case this observation model is contaminated by noise, we assume a Gaussian noise with zero mean and unknown variance $\sigma^2_v$, the likelihood model becomes \citep{chu2005gaussian}:
\begin{equation}
\label{eq:ordinalike1}
 p(y_i|f(\bx_i))=\int I_{(b_{y_{i-1}},b_{y_{i}}]}(f(\bx_i)+v) \phi\left(\tfrac{v}{\sigma_v}\right) dv=\Phi\left(\frac{b_{y_{i}}-f(\bx_i)}{\sigma_v}\right)-\Phi\left(\frac{b_{y_{i-1}}-f(\bx_i)}{\sigma_v}\right).
\end{equation}
The  $b_i$ defining the partition $(-\infty,b_1],(b_1,b_2],\dots,(b_{r-1},\infty)$
are unknown, that is they are hyperparameters of the model.
Note also that binary classification can be see as a special case of ordinal regression with $r=2$.

\paragraph{Preference:} In preference learning, 
$y_i$ is a preference relation over a set of predefined labels. For instance,  assume we label each 
input $\bx_i$ with the label $i$, then we can define a preference relation over the inputs. The label preference  $i\succ j$ means that the input $\bx_i$ is preferred to $\bx_j$.
This can be modelled by assuming that  there is an unobservable latent  function  $f$ associated  with each input $\bx_k$, and that the function values $\{f(\bx_k)\}_k$ preserve the preference relations observed in the dataset. 
Then the likelihood can be modelled as an indicator function $p(\bx_i\succ \bx_j|f(\bx_i),f(\bx_j))=I_{[f(\bx_j),\infty)}(f(\bx_i))$. This constrains the latent function values of the instances to be consistent with their preference relations. To allow some tolerance to noise in the inputs or the preference relations, one can assume the latent functions are contaminated with Gaussian noise \citep{ChuGhahramani_preference2005}:
\begin{equation}
\label{eq:preferenceike1}
\begin{aligned}
  p(\bx_i\succ \bx_j|f(\bx_i),f(\bx_j))&=\int I_{[f(\bx_j)+v_1,\infty)}(f(\bx_i)+v_2) \phi\left(\tfrac{v_1}{\sigma_v}\right) \phi\left(\tfrac{v_2}{\sigma_v}\right) dv_1dv_2\\
  &=\Phi\left(\frac{f(\bx_i)-f(\bx_j)}{\sqrt{2}\sigma_v}\right).
\end{aligned}
\end{equation}
More generally, the preferences of each input can be presented in the form of a  preference graph \citep{ChuGhahramani_preference2005}, where the labels are the graph vertices. Some examples are shown in \citep[Fig.1]{ChuGhahramani_preference2005}.
Therefore, in this more general case,
$y_i=\{(c_i^{j^+},c_i^{j^-})\}_{j=1}^{g_i}$, where $c_i^{j^-}$ is the initial label vertex of the $j$-th edge and $c_i^{j^+}$ is the terminal label, and $g_i$ is the number of edges.
This setting can be modelled by introducing a 
latent function $f_l$ for each predefined label \citep{ChuGhahramani_preference2005}.
The observed edge $(c_i^{j^+},c_i^{j^-})$ is modelled as the following constraint
$f_{c_i^{j^+}}(\bx_i) \geq f_{c_i^{j^-}}(\bx_i)$ and the likelihood is
\begin{equation}
\label{eq:preferenceike2}
\begin{aligned}
  p(\{(c_i^{j^+},c_i^{j^-})\}_{j=1}^{g_i}|{\bf f}(\bx_i))&=\prod\limits_{j=1}^{g_i}\Phi\left(\frac{f_{c_i^{j^+}}(\bx_i) - f_{c_i^{j^-}}(\bx_i)}{\sqrt{2}\sigma}\right),
\end{aligned}
\end{equation}
where ${\bf f}$ denotes the vector of latent function (one for each label).
For instance, note that standard Multiclass Classification \citep{williams1998bayesian}, Ordinal Regression, Hierarchical Multiclass classification, can be formulated in this way
\citep{ChuGhahramani_preference2005}.

\begin{figure}[htp!]
\centering
 \includegraphics[width=12cm]{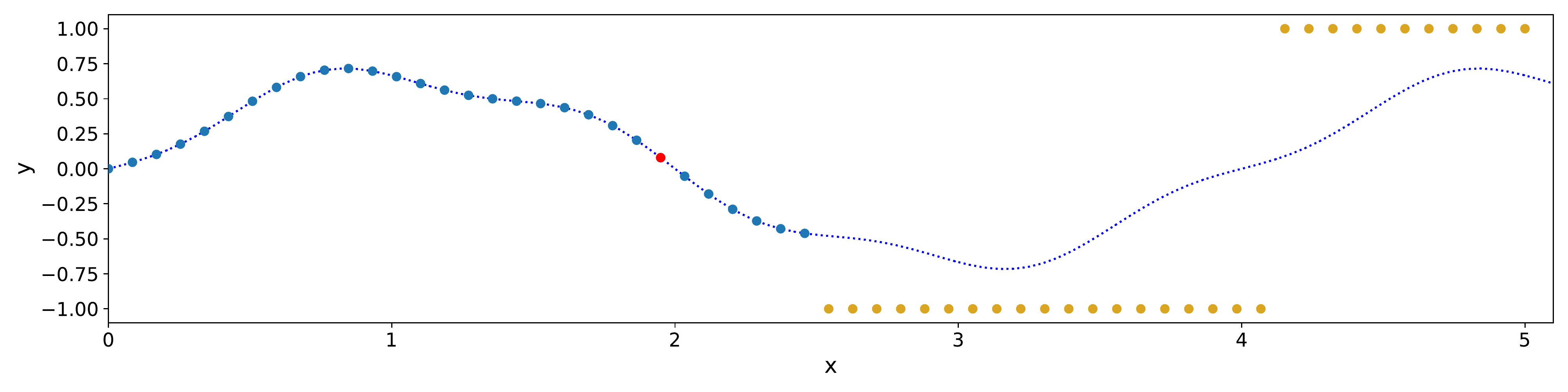}
 \caption{Mixed numeric and preference observations}
 \label{fig:1}
\end{figure}

\paragraph{Mixed:} In some applications, we may have scalar, binary and preference  observations at the same time.
Assuming independence, the likelihood model of $n$ mixed observations can in general be written as the product of  normal PDFs and normal CDFs. An example of mixed type data is shown in Figure \ref{fig:1}. The dotted line represents the function we used to generate the observations. The  left (blue) points are numeric (non-noisy) observations and the right points represent preferences. We used the colored points (red and gold) to visualise the $30$ preferential observations. The meaning of these points is as follows: (i)  the value of the functions computed at the $x$s corresponding to the bottom gold  points is less than the value of the function computed at the $x$ corresponding to the red point; (ii) the value of the function computed at the $x$s corresponding to the the  top gold  points is greater than the value of the function computed at the $x$ corresponding to the red point. These $30$ qualitative judgments is the only information we have on the function for $x \in [2.5,5]$. 
\\
 
~\\
\paragraph{State-of-the-art:} The state-of-the-art
(nonparametric) Bayesian approach to deal   with the above problems is to impose a Gaussian Process (GP) prior on the (latent) function(s) $f$.
For scalar observations, due to the conjugacy between normal likelihood and normal prior, the posterior model is still a GP and its mean and covariance functions can be computed analytically \citep{o1978curve}, \citep[Ch.2]{rasmussen2006gaussian}.

For binary data (classification), the posterior process is not a GP. Several algorithms for approximate inference have been proposed, which are based on approximating the non-Gaussian posterior with a tractable Gaussian distribution. There are three main types of approximation: (i) Laplace Approximation (LP) \citep{mackay1996bayesian,williams1998bayesian}; (ii) Expectation Propagation (EP) \citep{minka2001family}; (iii) Kullback-Leibler divergence (KL) minimization \citep{opper2009variational}, comprising Variational Bounding (VB) \citep{gibbs2000variational} as a particular case. An exhaustive theoretical and empirical analysis of the above approaches was performed by \cite{nickisch2008approximations}. They conclude that EP approximation is, in terms of accuracy,
always the method of choice, except when you cannot afford the slightly longer running time compared to the fastest LP approximation.

Ordinal regression with GPs was proposed by \cite{chu2005gaussian} using the likelihood  \eqref{eq:ordinalike1}. The posterior is not a GP
and so two approximations of the posterior were derived: LP and EP. The authors show that both LP and EP outperform the support vector approach (SVM), and that the EP approach is generally better than LP.

Preference learning based on GPs was proposed in \citep{ChuGhahramani_preference2005} using the likelihoods  \eqref{eq:preferenceike1} and \eqref{eq:preferenceike2}. Again, the posterior is not a GP and  the LP approximation is used to approximate the posterior with a GP; the approach outperforms SVMs. 

In a recent paper \citep{Benavoli_etal2020}, we showed that, although the probit likelihood 
\eqref{eq:probitlike1} and the GP are not conjugate, the posterior process can still be computed in closed form and it is a \text{Skew Gaussian Process} (SkewGP).
Moreover, by extending a result derived by 
\cite{durante2018conjugate} for the parametric case, we proved that SkewGP and probit likelihood are conjugate. Such a novel result allowed us  to  compute the exact  posterior for binary  classification and for preference learning \citep{benavoli2020preferential}.

In the next sections, we extend this result by showing that SkewGP is conjugate with both the normal and affine probit likelihood and, more in general, with their product. This shows that SkewGP encompasses GP for both regression and classification.

The rest of the paper is organised as follows. Section \ref{sec:skewnormal}
reviews the properties of the Skew Normal distribution and defines Skew Gaussian Processes. Section \ref{sec:main}, which includes the main results of the paper,  shows that SkewGPs provide closed-form solution to nonparametric regression, classification, preference and mixed problems.
Section \ref{sec:cdfs} provides algorithms to efficiently compute predictions for SkewGPs and to compute a fast approximation of the marginal likelihood.
Section \ref{sec:applications} discusses the application of SkewGPs to active learning and Bayesian optimisation. We show that SkewGPs outperform
the Laplace and Expectation Propagation approximation.
Finally, Section \ref{sec:conclusions} concludes the paper.

\section{Background on the Skew-Normal distribution and Skew Gaussian Processes}
\label{sec:skewnormal}

Skew-normal distributions  have long been seen \citep{o1976bayes} as generalizations of the normal distribution allowing for non-zero skewness. Here we follow \cite{o1976bayes} and we say that a real-valued continuous random variable has skew-normal distribution if it has the following probability density function (PDF)
$$
p(z)={\frac {2}{\sigma }}\phi \left({\frac {z-\xi }{\sigma }}\right)\Phi \left(\alpha \left({\frac {z-\xi }{\sigma }}\right)\right), \qquad z \in \mathbb{R},
$$
where $\phi$ and $\Phi$ are the PDF and Cumulative Distribution Function (CDF), respectively, of the standard univariate Normal distribution. The numbers $\xi \in \mathbb{R}, \sigma>0, \alpha \in \mathbb{R}$ are the location, scale and skewness parameters respectively.


The generalization of a univariate skew-normal to the multivariate case is not completely straightforward and over the years many generalisations of this distribution were proposed. \cite{arellano2006unification} provided a unification of those generalizations in a single and tractable multivariate \textit{Unified  Skew-Normal} distribution. This distribution satisfies closure properties for marginals and conditionals
and allows more flexibility due the introduction of additional
parameters.

\subsection{Unified Skew-Normal distribution}
\label{sec:unified}
The Unified Skew-Normal is a very general family of multivariate distributions that allows for skewness on different directions through latent variables. We say that a $p$-dimensional vector $\bz \in \mathbb{R}^p$ is distributed as a Unified Skew-Normal distribution with latent skewness dimension $s$, $ \bz \sim \text{SUN}_{p,s}(\bxi,\Omega,\Delta,\bgamma,\Gamma)$, if its probability density function \citep[Ch.7]{azzalini2013skew} is:
\begin{equation}
\label{eq:sun}
p(\bz) = \phi_p(\bz-\bxi;\Omega)\frac{\Phi_s\left(\bgamma+\Delta^T\bar{\Omega}^{-1}\Domega^{-1}(\bz-\bxi);\Gamma-\Delta^T\bar{\Omega}^{-1}\Delta\right)}{\Phi_s\left(\bgamma;\Gamma\right)},
\end{equation}
where $\phi_p(\bz-\bxi;\Omega)$ is the PDF of a multivariate Normal distribution with mean $\bxi \in \mathbb{R}^p$ and covariance $\Omega=\Domega\bar{\Omega} \Domega\in \mathbb{R}^{p\times p}$, $\bar{\Omega}$ is a correlation matrix and $\Domega$ a diagonal matrix  containing the square root of the diagonal elements in $\Omega$. The notation $\Phi_s(\bl[a];M)$ denotes the CDF of $\phi_s(0;M)$ evaluated at $\bl[a]\in \mathbb{R}^s$. 
The distribution is  parametrized by a location vector $\bxi$, a covariance matrix $\Omega$ and the latent variable parameters $\bgamma \in \mathbb{R}^s, \Gamma \in \mathbb{R}^{s\times s},\Delta^{p \times s}$. In particular $\Delta$ is the skewness matrix.

	\begin{figure}[htp]
	\centering
	\begin{tabular}{c @{\quad} c }
		\includegraphics[width=.48\linewidth]{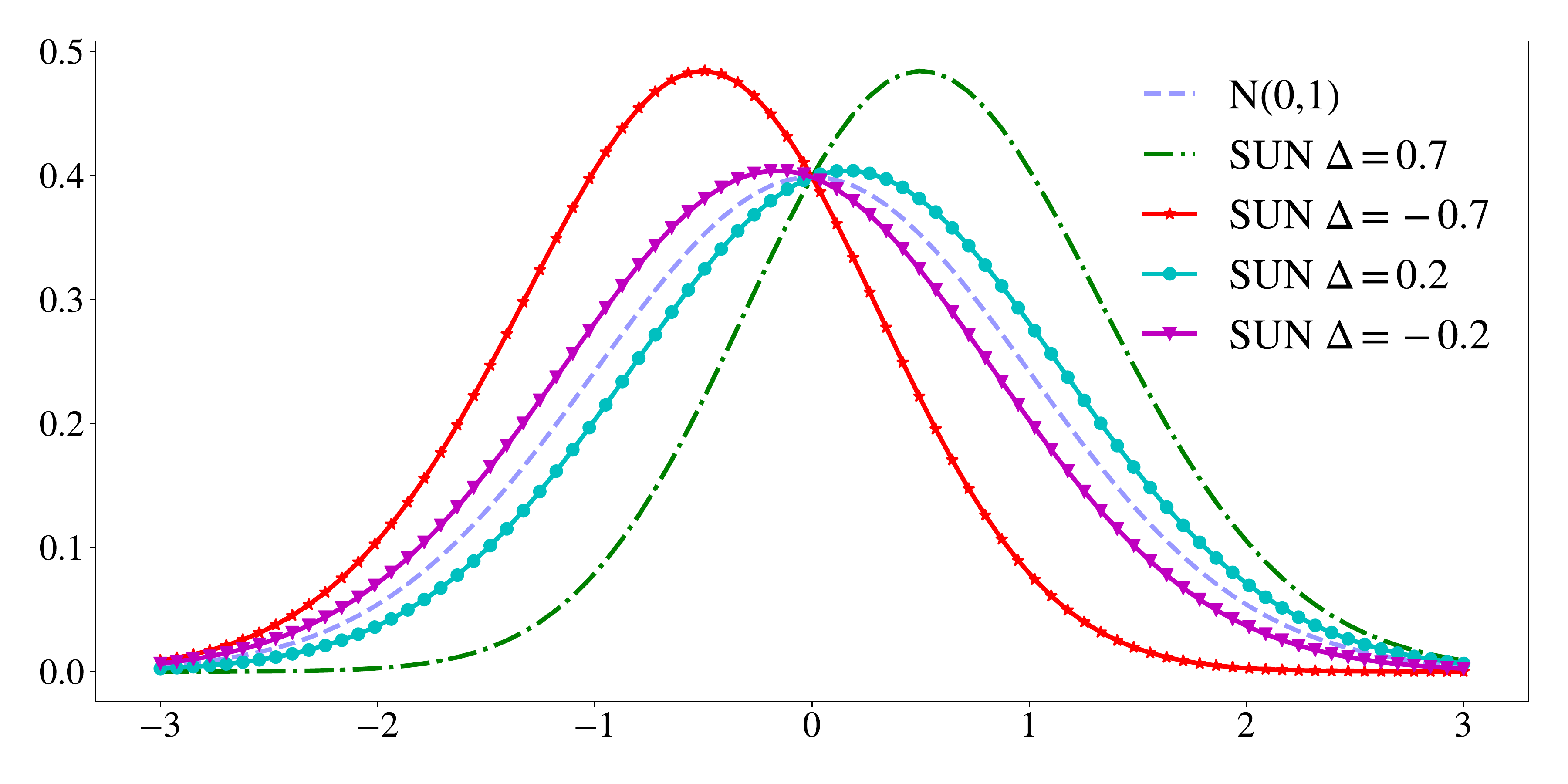} &
		\includegraphics[width=.48\linewidth]{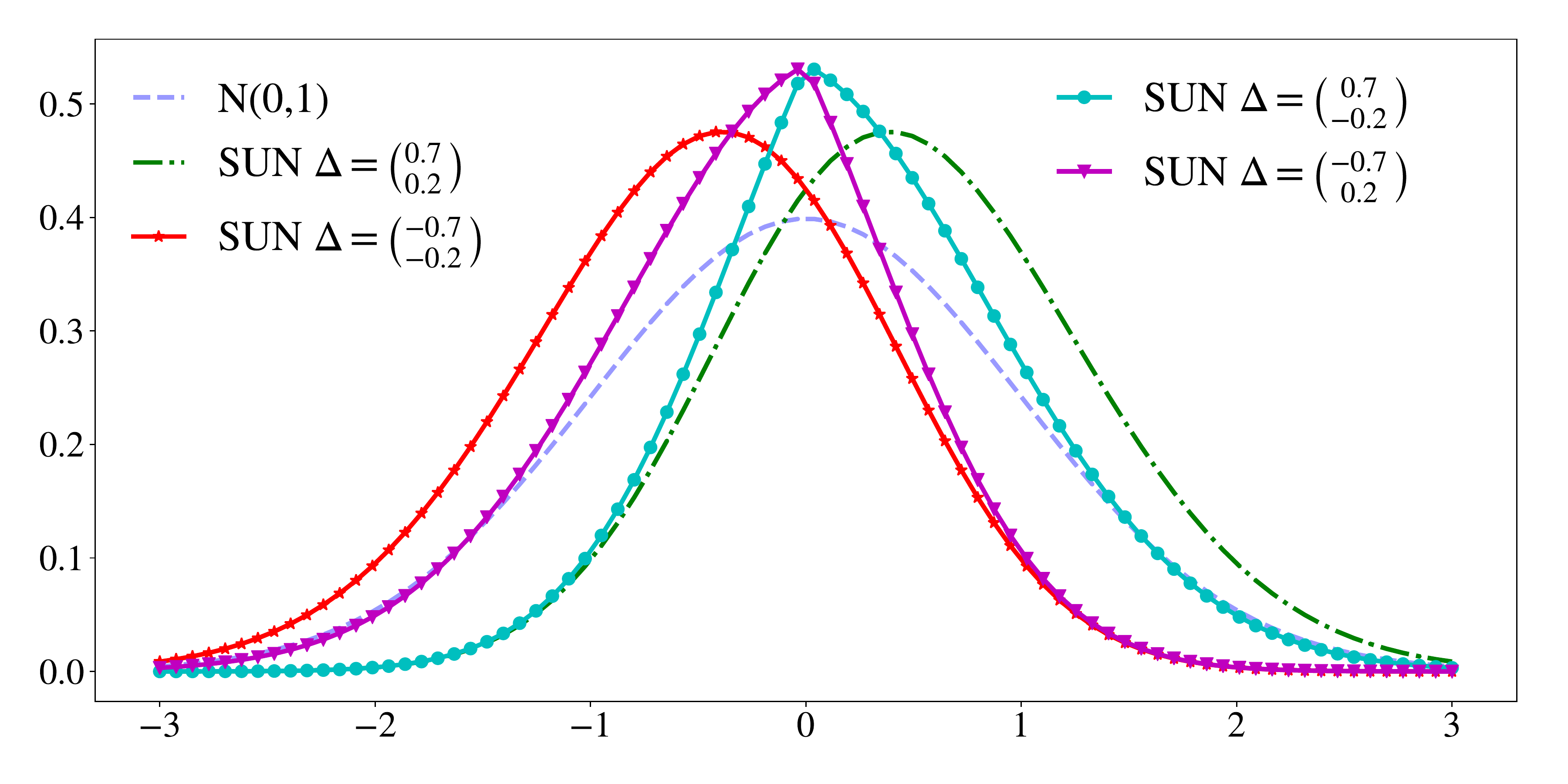} \\
		\small (a1) $s=1$, $\Gamma=1$  & \small (a2) $s=2$, $\Gamma_{1,2}=0.7$
	\end{tabular}
	\caption{Density plots for $\text{SUN}_{1,s}(0,1,\Delta,\gamma,\Gamma)$. For all plots $\Gamma$ is a correlation matrix, $\gamma = 0$, dashed lines are the contour plots of $y \sim N_1(0,1)$.}
	\label{fig:SUN1d}
\end{figure}

\begin{figure}
	\centering
	\begin{tabular}{c @{\quad} c @{\quad} c @{\quad} c}
		\includegraphics[width=.24\linewidth]{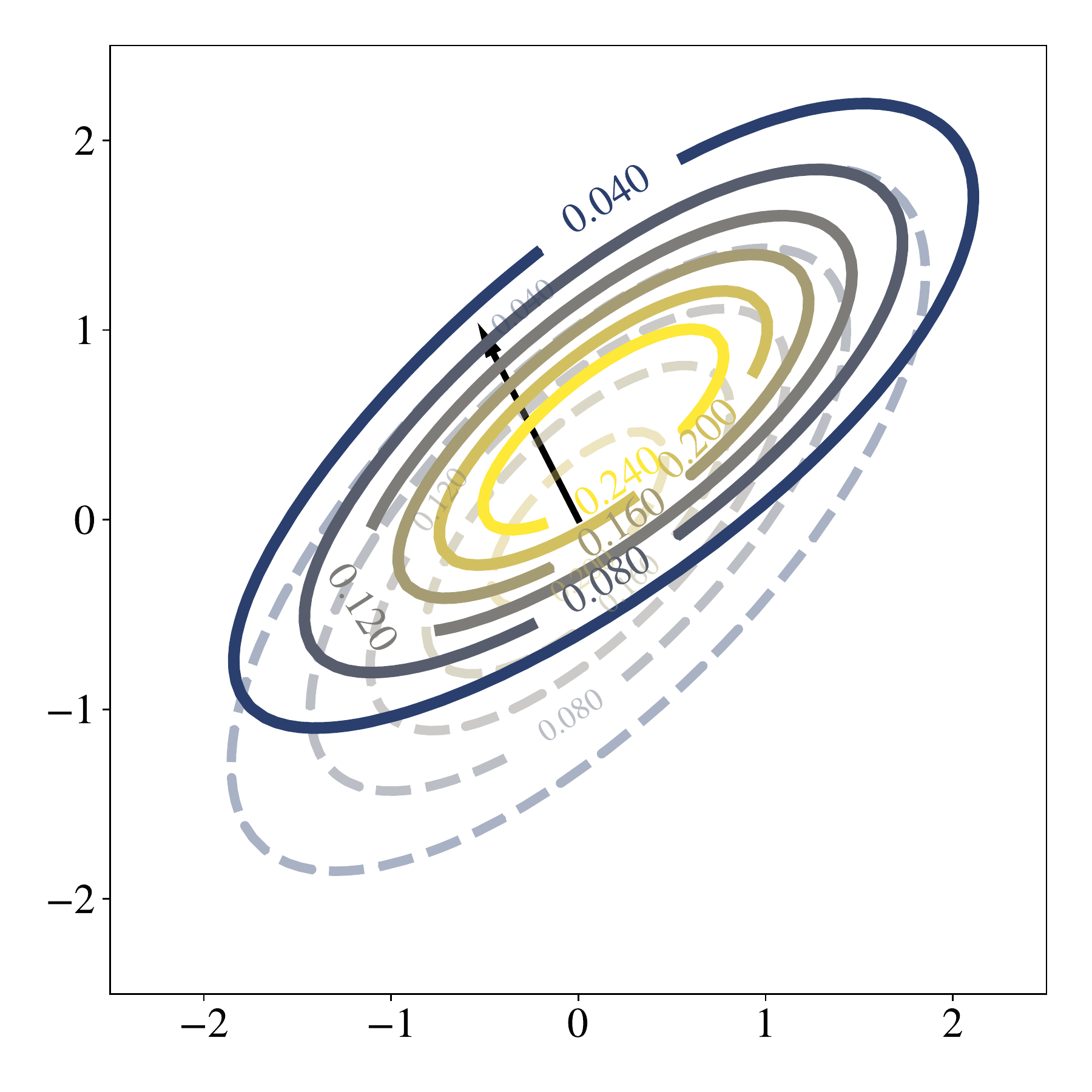} &
		\includegraphics[width=.24\linewidth]{{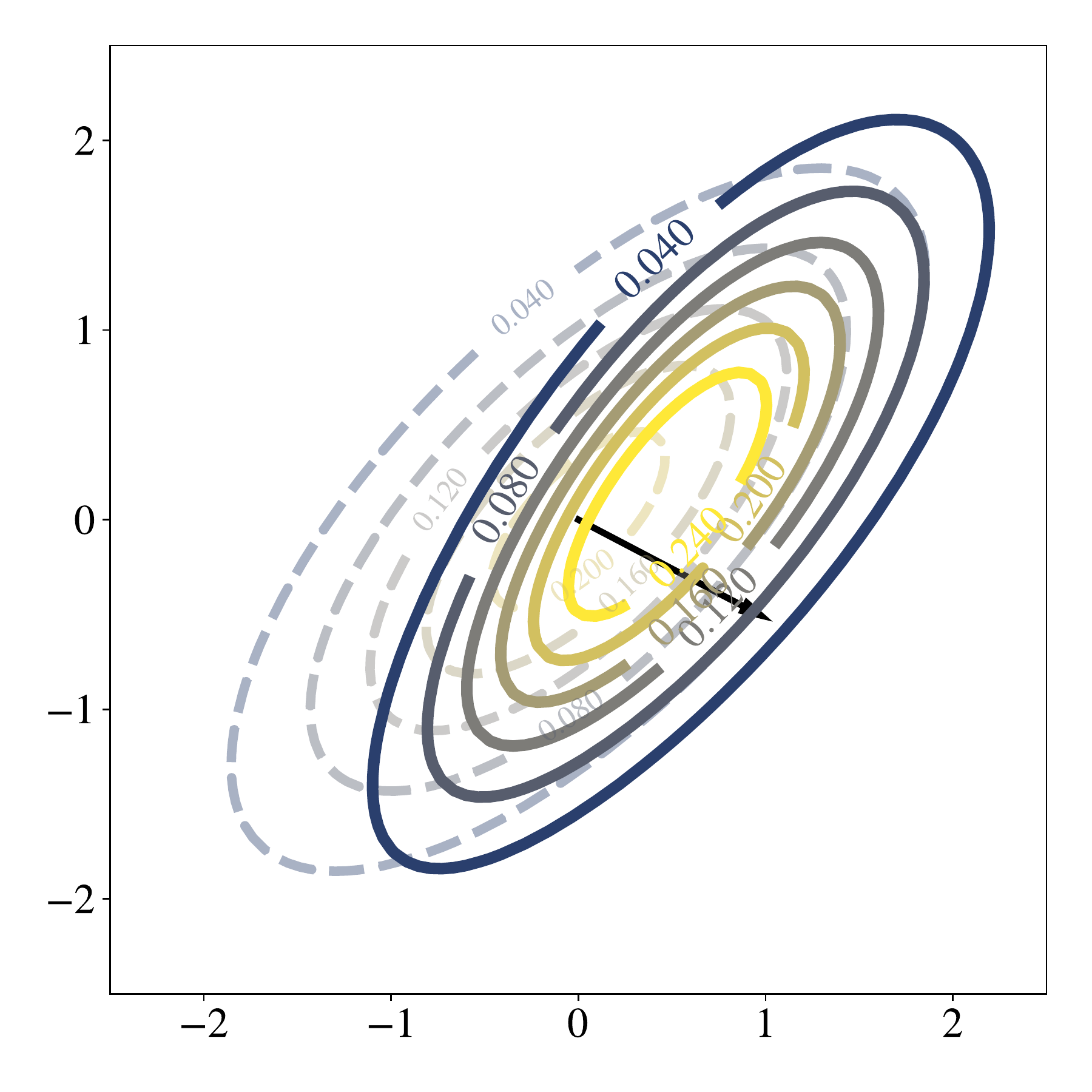}} &
		\includegraphics[width=.24\linewidth]{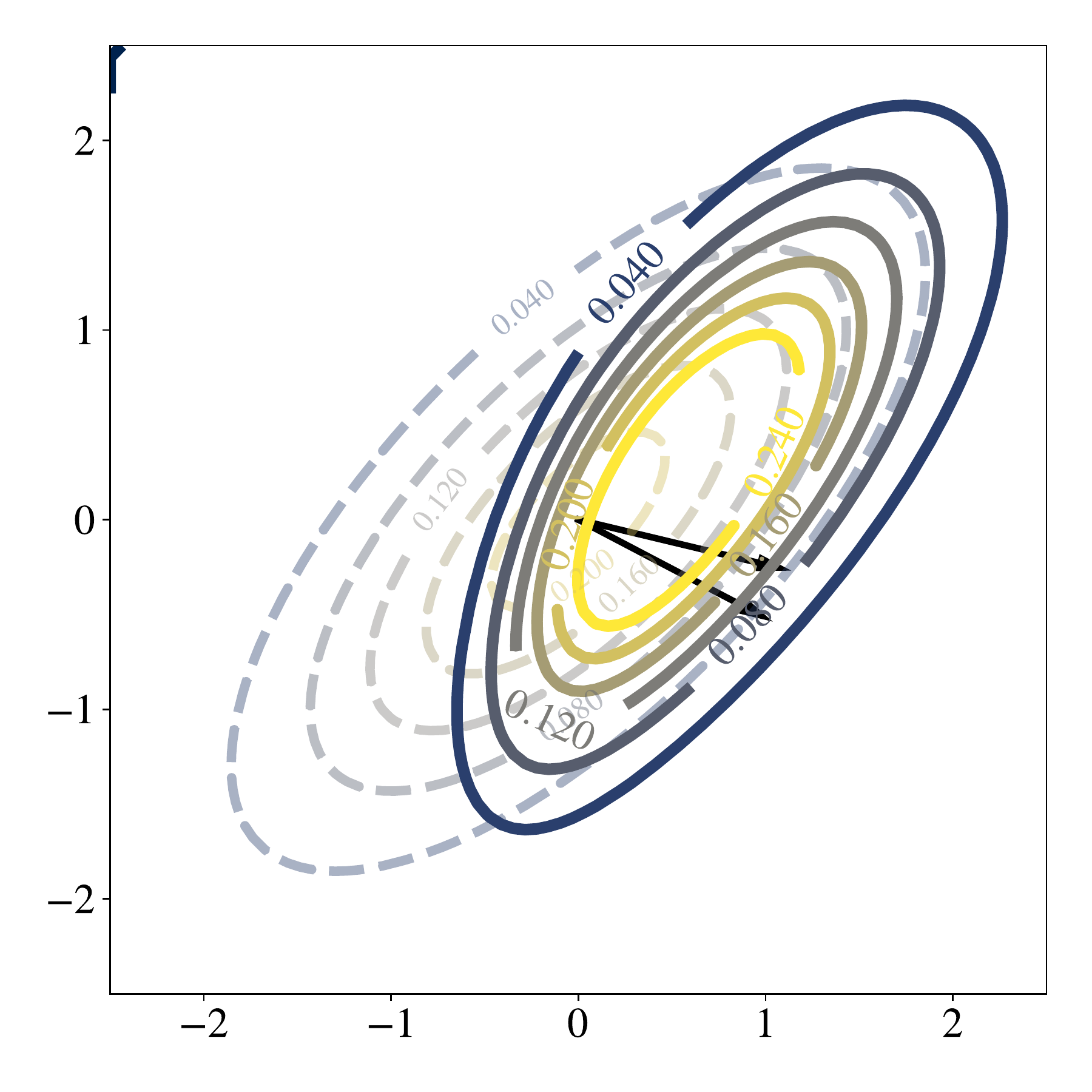} &
		\includegraphics[width=.24\linewidth]{{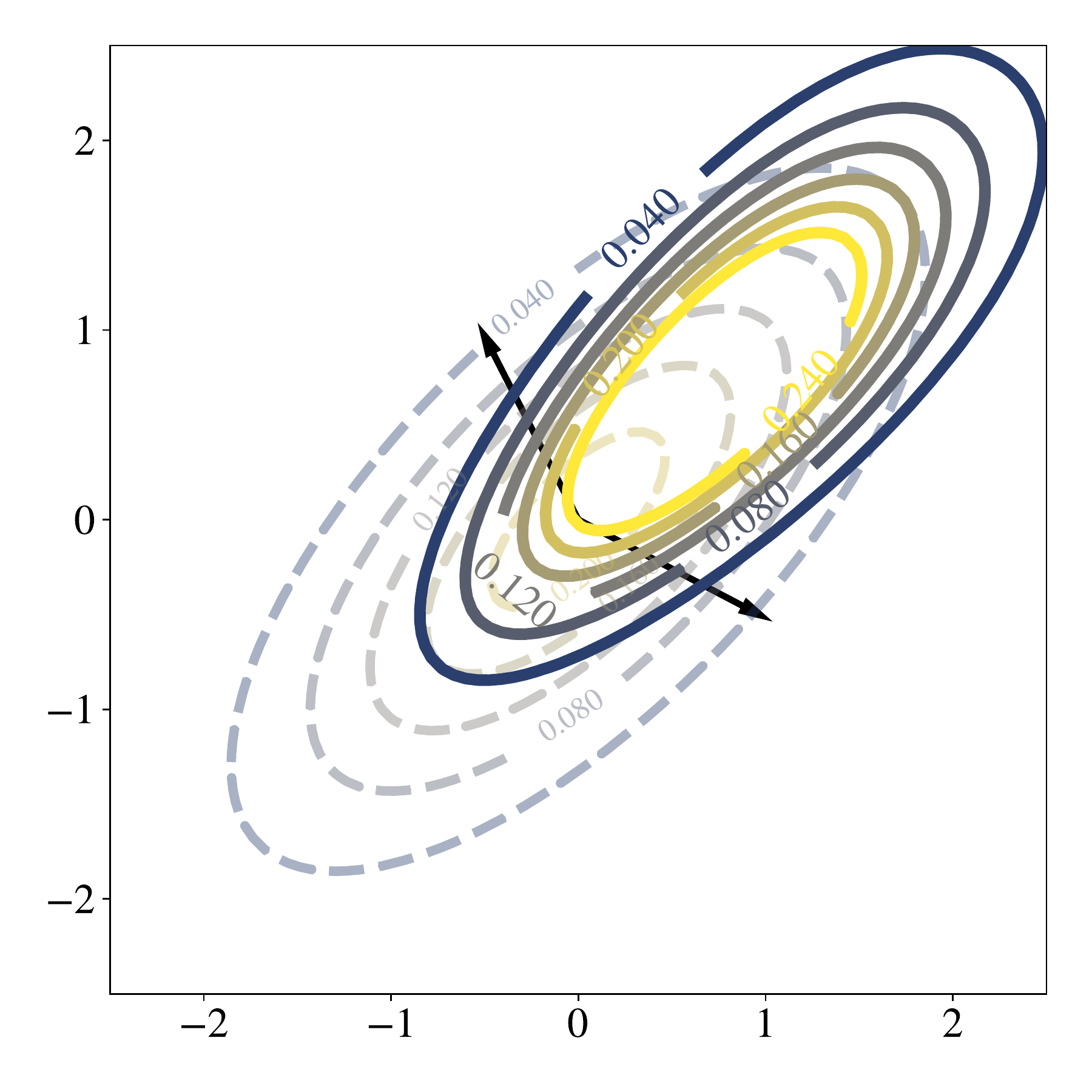}} \\
		\small (a1) $s=1$, $\Gamma=1$  & \small (a2) $s=1$, $\Gamma=1$ & \small (a3) $s=2$, $\Gamma_{1,2}=0.9$ & \small (a4) $s=2$, $\Gamma_{1,2}=-0.2$ \\
		\small $\Delta = [0.7,0.2]^T$,  & \small  $\Delta = [0.2,0.7]^T$ & \small $\Delta = \begin{bmatrix}
		0.2 & 0.5 \\
		0.7 & 0.9
		\end{bmatrix}$ & $\Delta = \begin{bmatrix}
		0.2 & 0.7 \\
		0.7 & 0.2
		\end{bmatrix}$ 
	\end{tabular}
	\caption{Contour density plots for four unified skew-normal. For all plots $p=2$, $\bxi=[0,0]^T$, $\Omega$ and $\Gamma$ are correlation matrices with $\Omega_{1,2}= 0.7$, $\gamma = 0$, dashed lines are the contour plots of $y \sim N_2(\bxi,\Omega)$.}
	\label{fig:SUN}
\end{figure}

The PDF  \eqref{eq:sun} is well-defined provided that the matrix $M$ below is positive definite, i.e.\footnote{In \citep{azzalini2013skew}, it is assumed that $M$ is a correlation matrix. This means that $\Gamma$ must be a correlation matrix.
However, from  \eqref{eq:sun} note that, $\tfrac{\Phi_s\left(\bgamma+\Delta^T\bar{\Omega}^{-1}\Domega^{-1}(\bz-\bxi);\Gamma-\Delta^T\bar{\Omega}^{-1}\Delta\right)}{\Phi_s\left(\bgamma;\Gamma\right)}$ is equal to  $\tfrac{\Phi_s\left({\bf v}^{-1}\bgamma+{\bf v}^{-1}\Delta^T\bar{\Omega}^{-1}\Domega^{-1}(\bz-\bxi);{\bf v}^{-1}(\Gamma-\Delta^T\bar{\Omega}^{-1}\Delta){\bf v}^{-1}\right)}{\Phi_s\left({\bf v}^{-1}\bgamma;{\bf v}^{-1}\Gamma{\bf v}^{-1}\right)}$  where ${\bf v}$ is the diagonal matrix that makes ${\bf v}^{-1}\Gamma{\bf v}^{-1}$
a correlation matrix. Therefore, in this paper, for simplicity we do not restrict  $\Gamma$ to be a correlation matrix. In any case, this does not have any impact on
the properties of the SUN distribution we discuss in this section.}
\begin{equation}
\label{eq:positivity}
M:=\begin{bmatrix}
\Gamma & \Delta^T\\
\Delta & \bar{\Omega}
\end{bmatrix} \in \mathbb{R}^{(s+p)\times(s+p)}>0.
\end{equation}
Note that when $\Delta=0$, \eqref{eq:sun}  reduces  to  $\phi_p(\bz-\bxi;\Omega)$, i.e. a skew-normal with zero skewness matrix is a normal distribution. Moreover we assume that $\Phi_0(\cdot)=1$, so that, for $s=0$, \eqref{eq:sun} becomes a multivariate Normal distribution.

Figure~\ref{fig:SUN1d} shows the density of a univariate SUN distribution with latent dimensions $s=1$ (a1) and $s=2$ (a2). The effect of a higher latent dimension can be better observed in bivariate SUN densities as shown in  Figure~\ref{fig:SUN}. 
The contours of the corresponding bivariate normal are dashed. We also plot the skewness directions given by $\bar{\Omega}^{-1}\Delta$. Note that a SUN with two latent dimensions (Fig.~\ref{fig:SUN}, (a3), (a4)) has two direction of skewness. 

\subsection{Additive representations}	
\label{sec:Additive}
The role of the latent dimension $s$ can be briefly explained as follows. 
Consider a random vector $\begin{bmatrix}
	\bx_0 \\
	\bx_1 
	\end{bmatrix} \sim \phi_{s+p}(0;M)$ with $M$ as in \eqref{eq:positivity} and define $\mathbf{y}$ as the vector with distribution $(\bx_1 \mid \bx_0+\bgamma>0)$. The density of $y$ can be written as 
	\begin{align*}
	f(\mathbf{y}) &= \frac{\int_{\bx_0+\bgamma>0}\phi_{s+p}((\mathbf{t}_0,\mathbf{y});M)d\mathbf{\mathbf{t}_0}}{\int_{\bx_0+\bgamma>0}\phi_s(\mathbf{t};\Gamma)d\mathbf{t}} = \phi_p(\mathbf{y};\bar{\Omega})\frac{P(\bx_0+\bgamma>0 \mid \bx_1 = \mathbf{y})}{\Phi_s(\bgamma;\Gamma)} \\
	&= \phi_p(\mathbf{y};\bar{\Omega})\frac{\Phi_s\left(\bgamma+\Delta^T\bar{\Omega}^{-1}\mathbf{y}; \Gamma-\Delta^T\bar{\Omega}^{-1}\Delta\right)}{\Phi_s(\bgamma;\Gamma)},
	\end{align*}
	where the first equality comes from a basic property of conditional distributions, see, e.g.\citep[Ch. 1.3]{azzalini2013skew}, and the second equality is a consequence of the multivariate normal conditioning properties. Then we have that 
	$$\bz = \bxi + \Domega\mathbf{y}\sim\text{SUN}_{p,s}(\bxi,\Omega,\Delta,\bgamma,\Gamma).$$
	Note that the skewness of $\bz$ is determined by the correlation $\Delta$ of $\bx_1$ with the latent $s$-dimensional vector $\bx_0$.
	
	The previous representation is useful for understanding the role of the latent dimension $s$ in a skew-Gaussian vector. We report below another representation which is more practical for sampling. Consider the independent random vectors 	
	$\mathbf{r}_0 \sim \phi_p(0; \bar{\Omega} - \Delta \Gamma^{-1}\Delta^T)$ and $\mathbf{r}_{1,-\bgamma}$, the truncation below $\bgamma$ of $\mathbf{r}_1\sim \phi_s(0,\Gamma)$. Then the random variable 
	\begin{equation}
	\mathbf{z}_u = \bxi + \Domega(\mathbf{r}_0 + \Delta \Gamma^{-1}\mathbf{r}_{1,-\bgamma}),
	\label{eq:additiveSUN}
	\end{equation}
	is distributed as~\eqref{eq:sun}, \citep[Ch.7]{azzalini2013skew} and \citep[Sec.A.1]{benavoli2020preferential}. 
	 The additive representation introduced above can be used to draw samples from the distribution as it will be discussed in Section \ref{sec:cdfs}. 

\subsection{Closure properties}
\label{sec:closure}
Among the many interesting properties of the Skew-Normal family (see \citet[Ch.7]{azzalini2013skew} for details), here we are particularly interested in its closure under marginalization and affine transformations. Consider $\bz \sim\text{SUN}_{p,s}(\bxi,\Omega,\Delta,\bgamma,\Gamma)$ and partition $\bz = [\bz_1 , \bz_2]^T$,
where $\bz_1 \in \mathbb{R}^{p_1}$ and $\bz_2 \in \mathbb{R}^{p_2}$
with $p_1+p_2=p$, then
\begin{equation}
\label{eq:marginalFinDim}
\begin{array}{c}
\bz_1  \sim SUN_{p_1,s}(\bxi_1,\Omega_{11},\Delta_1,\bgamma,\Gamma), \vspace{0.2cm}\\
\text{with }~~
\bxi=\begin{bmatrix}
\bxi_1\\\bxi_2
\end{bmatrix},~~~
\Delta=\begin{bmatrix}
\Delta_1\\\Delta_2
\end{bmatrix},~~~
\Omega=\begin{bmatrix}
\Omega_{11} & \Omega_{12}\\
\Omega_{21} & \Omega_{22}
\end{bmatrix}.
\end{array}
\end{equation}
Moreover, \citep[Ch.7]{azzalini2013skew} the conditional distribution is a unified skew-Normal, i.e.,  
%
$(\mathbf{Z}_2|\mathbf{Z}_1=\bz_1) \sim SUN_{p_2,s}(\bxi_{2|1},\Omega_{2|1},\Delta_{2|1},\bgamma_{2|1},\Gamma_{2|1})$, where 
\begin{align*}
\bxi_{2|1} & :=\bxi_{2}+\Omega_{21}\Omega_{11}^{-1}(z_1-\bxi_1), \quad
\Omega_{2|1} := \Omega_{22}-\Omega_{21}\Omega_{11}^{-1}\Omega_{12},\\
\Delta_{2|1} &:=\Delta_2 -\bar{\Omega}_{21}\bar{\Omega}_{11}^{-1}\Delta_1,\\
\bgamma_{2|1}& :=\bgamma+\Delta_1^T \Omega_{11}^{-1}(z_1-\bxi_1), \quad
\Gamma_{2|1}:=\Gamma-\Delta_1^T\bar{\Omega}_{11}^{-1}\Delta_1,
\end{align*}
and $\bar{\Omega}_{11}^{-1}:=(\bar{\Omega}_{11})^{-1}$.

In Section \ref{sec:main}, we exploit this property to obtain samples from the predictive posterior distribution at a new input $\bx^*$ given samples of the posterior at the training inputs.
%

\subsection{SkewGP}

The unified skew-normal distribution can be generalized \citep{Benavoli_etal2020} to a stochastic process.  We briefly recall here its construction.

Consider a location function $\xi: \mathbb{R}^d \rightarrow \mathbb{R}$, a scale (kernel) function $\Omega: \mathbb{R}^d \times \mathbb{R}^d \rightarrow \mathbb{R}$, a skewness vector function $\Delta: \mathbb{R}^d \rightarrow \mathbb{R}^s$ and the parameters $\bgamma \in \mathbb{R}^s, \Gamma \in \mathbb{R}^{s \times s}$.  We say $f: \mathbb{R}^d \rightarrow \mathbb{R}$ is a SkewGP with latent dimension $s$, if for any sequence of $n$ points $\bx_1, \ldots, \bx_n \in \mathbb{R}^d$, the vector $[f(\bx_1), \ldots, f(\bx_n)] \in \mathbb{R}^n$ is skew-normal distributed with parameters $\bgamma, \Gamma$ and location, scale and skewness matrices,
respectively, given by
\begin{equation}
\begin{array}{rl}
\xi(X):=\begin{bmatrix}
\xi(\bx_1)\\
\xi(\bx_2)\\
\vdots\\
\xi(\bx_n)\\
\end{bmatrix},~~
\Omega(X,X)&:=
\begin{bmatrix}
\Omega(\bx_1,\bx_1) & \Omega(\bx_1,\bx_2) &\dots & \Omega(\bx_1,\bx_n)\\
\Omega(\bx_2,\bx_1) & \Omega(\bx_2,\bx_2) &\dots & \Omega(\bx_2,\bx_n)\\
\vdots & \vdots &\dots & \vdots\\
\Omega(\bx_n,\bx_1) & \Omega(\bx_n,\bx_2) &\dots & \Omega(\bx_n,\bx_n)\\
\end{bmatrix},\vspace{0.2cm}\\
\Delta(X)&:=\begin{bmatrix}
~~\Delta(\bx_1) & \quad\quad\Delta(\bx_2) &~~\dots & ~\quad\Delta(\bx_n)\\
\end{bmatrix}.
\end{array}
\end{equation}

The skew-normal distribution is well defined if the matrix 
$M=\left[\begin{smallmatrix}
\Gamma & \Delta(X) \\
\Delta(X)^T & \bar{\Omega}(X,X)
\end{smallmatrix}\right]
$
is positive definite for all $X = \{\bx_1, \ldots, \bx_n \} \subset \mathbb{R}^d$ and for any $n$. \citet{Benavoli_etal2020} shows that SkewGp is a well defined stochastic process. In that case we write $f \sim \text{SkewGP}_s(\xi,\Omega,\Delta,\gamma,\Gamma)$.

We briefly review here a possible choice for the functions $\Omega, \Delta$ and the matrix $\Gamma$ that guarantees that $M$ is always positive definite. We follow \citet{Benavoli_etal2020} and we choose a positive definite kernel stationary function\footnote{This construction can easily be generalised to non-stationary kernels.} $K: \mathbb{R}^d\times \mathbb{R}^d \rightarrow \mathbb{R}$ which we will use to generate both $\Omega$ and $\Delta$. Given $n$ points and $X = \{\mathbf{x}_1, \ldots, \mathbf{x}_n\} \subset \mathbb{R}^d$, $s$ pseudo-points $U= \{\mathbf{u}_1, \ldots, \mathbf{u}_s \} \subset \mathbb{R}^d$ and a phase diagonal matrix $L \in \mathbb{R}^{s \times s}$ with elements $L_{ii} \in \{-1,1\}$, we build the matrix $M$ as 
\begin{equation}
M = \begin{bmatrix}
\Gamma & \Delta(X,U) \\
\Delta(U,X) & \Omega(X,X)
\end{bmatrix} = \begin{bmatrix}
L\bar{K}(U,U) L & L \bar{K}(U,X) \\
\bar{K}(X,U) L & \bar{K}(X,X)
\end{bmatrix},
\end{equation} 
where $\bar{K}(\mathbf{x},\mathbf{x}^\prime) = \frac{1}{\sigma^2}K(\mathbf{x},\mathbf{x}^\prime)$. This structure guarantees that $M$ is a positive definite matrix for any $X$. Pseudo-points allow for a flexible handling of the skewness determined by $\Gamma$ and $\Delta$. See \citet{Benavoli_etal2020} for several examples of the effect of the pseudo-points positions.

SkewGPs can then be used as prior for $f$ in a Bayesian model. Note that, with $s=0$, this construction recovers a GP with covariance kernel $\Omega=K$.  Below we show that this larger family of prior distributions for $f$ present remarkable conjugacy properties with many common likelihoods. 



\section{Conjugacy of SkewGP} 
\label{sec:main}
This section includes the main results of the paper: we will prove that SkewGP is conjugate with both the normal and probit affine likelihood and, more in general, with their product.

\subsection{Normal likelihood}
Consider $n$ input points $X = \{\bx_i : i=1, \ldots, n\}$, with   $\bx_i \in \mathbb{R}^d$, and $m_r$ output 
points  $Y = \{y_i : i=1, \ldots, m_r\}$, with   $y_i \in \mathbb{R}$ and the likelihood 
\begin{equation}
p(Y \mid f(X)) = \prod\limits_{i=1}^{m_r} \phi\left(\frac{y_i-{\bf c}_i^T f(X)}{\sigma_v}\right)=\phi_{m_r}(Y-Cf(X);R),
\label{eq:Normallike}
\end{equation}
with $\phi_{m_r}(Y-Cf(X);R)$ denotes a multivariate normal PDF with zero mean and covariance $R$ computed at $Y-Cf(X)$, where  $ \mathbb{R}^{m_r \times n} \ni C=[{\bf c}^T_i]_{i=1}^{m_r}$ and ${\bf c}_i \in \reals^n$ is a data dependent vector and $R=\sigma_v^2 I_{m_r}$ is a covariance matrix, with $I_{m_r}$ being the identity matrix of dimension $m_r$.

\begin{lemma}
	\label{lemma:Normal}
	Let us assume that
	$f({\bf x}) \sim \text{SkewGP}_s(\xi({\bf x}), \Omega({\bf x},{\bf x}'),\Delta({\bf x}),\bgamma, \Gamma)$ and consider the Normal likelihood $p(Y \mid f(X))=\phi_{m_r}(Y-Cf(X);R)$ where $C\in \mathbb{R}^{m_r \times n}$ and $R=\mathbb{R}^{m_r \times m_r}$.
		The posterior distribution of $f(X)$ is a SUN:
	\begin{align}
	\nonumber
	p(f(X)|Y)&= \text{SUN}_{m_r,s}(\bxi_p,\Omega_p,\Delta_p,\gamma_p,\Gamma_p)~~~\text{ with }\\
		\label{eq:posteriorclass}
	&\bxi_p  =\bxi+\Omega C^T(C\Omega C^T+R)^{-1}(Y-C\xi),\\
	&\Omega_p = \Omega-\Omega C^T(C\Omega C^T+R)^{-1}C\Omega,\\
	\label{eq:tt0}
	&\Delta_p =\bar{\Omega}_pD_{\Omega_p}\Domega^{-1} \bar{\Omega}^{-1}\Delta,\\
	\label{eq:tt1}
	&\bgamma_p =\bgamma+\Delta^T\bar{\Omega}^{-1}\Domega^{-1}(\bxi_p-\bxi),\\
	\label{eq:tt2}
	&\Gamma_p=\Gamma-\Delta^T\bar{\Omega}^{-1}\Delta +\Delta_p^T\bar{\Omega}_p^{-1}\Delta_p,
	\end{align}
	where, for simplicity of notation, we denoted $\xi(X),\Omega(X,X),\Delta(X)$ as $\bxi,\Omega,\Delta$, and
	$\Omega = \Domega \bar{\Omega} \Domega$ and $\Omega_p = D_{\Omega_p} \bar{\Omega}_p D_{\Omega_p}$.
\end{lemma}
All the proofs are in the Appendix. For either  $s=0$ or $\Delta=0$, the SkewGP prior becomes a GP and it can noted that, in this case, the posterior is  Gaussian with posterior mean $\bxi_p$ and posterior covariance $\Omega_p$ (that is the terms 
in \eqref{eq:tt0}--\eqref{eq:tt2} disappear).

In  practical applications of SkewGP, the hyperparameters of the scale function $\Omega(\bx,\bx')$, of the skewness vector function $\Delta(\bx) \in \mathbb{R}^s$  and the hyperparameters $\bgamma \in \mathbb{R}^s,\Gamma \in \mathbb{R}^{s \times s}$  must be selected. As for GPs, we use Bayesian model selection to  set such hyperparameters and this requires the maximization of the marginal likelihood with respect to these hyperparameters.
\begin{corollary}
\label{co:Normal}
 Consider the probabilistic model in Lemma \ref{lemma:Normal},  the marginal likelihood of the observations $Y$ is
 \begin{equation}
  \label{eq:ml_normal}
  p(Y)=\phi_{m_r}(Y-C\xi(X);C\Omega(X,X)C^T+R)\frac{ \Phi_{s}(\bgamma_p;~\Gamma_p)}{\Phi_{s}(\bgamma;~\Gamma)}.
 \end{equation}
\end{corollary}
Observe again that for either  $s=0$ or $\Delta=0$, the marginal likelihood coincides with that of a GP because the ratio $\Phi_{s}(\bgamma_p;~\Gamma_p)/\Phi_{s}(\bgamma;~\Gamma)$ disappears.
The computation of the marginal likelihood \eqref{eq:ml_normal} requires the calculation of two multivariate CDFs. We will address this point in Section \ref{sec:cdfs}.

We now prove that,  a-posteriori, for a new test point $\bf x$, the function  $f(\bf x)$ is SkewGP distributed under the Normal likelihood in \eqref{eq:Normallike}.

\begin{theorem}
	\label{th:1}
	Let us assume a SkewGP prior 	$f({\bf x}) \sim \text{SkewGP}_s(\xi({\bf x}), \Omega({\bf x},{\bf x}'),\Delta({\bf x}),\bgamma, \Gamma)$, the likelihood $p(Y \mid f(X)) = \phi_{m_r}(Y-Cf(X);R) $, then a-posteriori $f$ is SkewGP with mean, scale, and skew functions:
	\begin{align}
	\label{eq:ss1}
	\tilde{\bxi}({\bf x})  &=\bxi({\bf x})+\Omega({\bf x},X) C^T(C\Omega(X,X) C^T+R)^{-1}(Y-C\xi(X)),\\
	\label{eq:ss2}
	\tilde{\Omega}({\bf x},{\bf x}) &= \Omega({\bf x},{\bf x})-\Omega({\bf x},X) C^T(C\Omega(X,X) C^T+R)^{-1}C\Omega(X,{\bf x}),\\
	\nonumber
	\tilde{\Delta}({\bf x}) &=D_{\tilde{\Omega}(\bx,\bx)}^{-1}D_{\Omega(\bx,\bx)}\Delta({\bf x})\\
	\label{eq:ss3}
	&-D_{\tilde{\Omega}(\bx,\bx)}^{-1}\Omega({\bf x},X)C^T (C\Omega(X,X) C^T+R)^{-1} C D_{\Omega(X,X)}\Delta(X),
	\end{align}
	and parameters $\bgamma_p,\Gamma_p$ as in Lemma \ref{lemma:Normal}.
\end{theorem}
The posterior mean \eqref{eq:ss1} and
posterior covariance \eqref{eq:ss2} coincide with those of the posterior GP. The term \eqref{eq:ss3}
is the posterior skewness. This results proves that SkewGP process and the Normal likelihood are conjugate.

\subsection{Probit affine likelihood}
Consider $n$ input points $X = \{\bx_i : i=1, \ldots, n\}$, with   $\bx_i \in \mathbb{R}^d$, and a data-dependent matrix $W \in \mathbb{R}^{m_a\times n}$ and vector $Z \in \mathbb{R}^{m_a}$. We define an affine probit likelihood as 
\begin{equation}
p(W,Z \mid f(X)) = \Phi_{m_a}(Z+Wf(X); \Sigma),
\label{eq:affineProbit}
\end{equation}
where $\Phi_{m_a}(\bx; \Sigma)$ is the $m_a$-variate Gaussian CDF evaluated at $\bx \in\mathbb{R}^m$  with covariance $\Sigma \in \mathbb{R}^{m_a\times m_a}$. 
Note that this likelihood model includes the classic GP probit classification model \citep{rasmussen2006gaussian} with binary observations $y_1, \ldots, y_n \in \{0,1\}$ encoded in the matrix
$
W=\text{diag}(2y_1-1, \ldots, 2y_n-1)
$, 
where $m=n$, $Z=0$ and $\Sigma=I_{m_a}$ (the identity matrix of dimension $m_a$), that is
\begin{equation}
\label{eq:probWclass}
\Phi_{m_a}\left(   \begin{bmatrix}
    2y_1-1 & & \\
    & \ddots & \\
    & & 2y_n-1
  \end{bmatrix}f(X); I_{m_a} \right).
\end{equation}
Moreover,  the likelihood in \eqref{eq:affineProbit} is equal to the preference likelihood \eqref{eq:preferenceike1} for $Z=0$ and a particular choice of $W$. In fact, consider the likelihood 
\begin{equation}
\label{eq:affinprefW}
\Phi_{m_a}\left( \begin{bmatrix}
	\tfrac{f(\bv_1) - f(\bu_1)}{\sigma_v} \\
	\vdots \\
	\tfrac{f(\bv_m) - f(\bu_m)}{\sigma_v}
	\end{bmatrix}; I_{m_a} \right),
\end{equation}
 where
	$\bv_i,\bu_j \in X$. If we denote by $W \in \mathbb{R}^{m_a \times n}$ the matrix defined as $W_{i,j} = V_{i,j} - U_{i,j}$ where $V_{i,j}=\tfrac{1}{\sigma_v}$ if $\bv_i={\bf x}_j$ and $0$ otherwise and $U_{i,j}=\tfrac{1}{\sigma_v}$ if $\bu_i={\bf x}_j$ and $0$ otherwise. Then we can write the likelihood \eqref{eq:affinprefW} as in \eqref{eq:affineProbit}.\footnote{We omitted the constant $\sqrt{2}$ for simplicity.}
	The conjugacy between SkewGP and the likelihood \eqref{eq:probWclass} was proved in \citep{Benavoli_etal2020}  extending to the nonparametric setting a result 
	proved in \citep[Th.1 and Co.4]{durante2018conjugate} for the parametric setting. The conjugacy between SkewGP and the likelihood  \eqref{eq:affinprefW} was derived in \citep{benavoli2020preferential} extending the results for classification.
	In this paper, we further expand those results by including the shifting term $Z$ (which, for instance, allows us to deal with the likelihood \eqref{eq:ordinalike1}) and a general covariance matrix $\Sigma$, as in \eqref{eq:affineProbit}.


\begin{lemma}
	\label{lemma:Affine}
	Let us assume that
	$f({\bf x}) \sim \text{SkewGP}_s(\xi({\bf x}), \Omega({\bf x},{\bf x}'),\Delta({\bf x}),\bgamma, \Gamma)$ and consider the likelihood $ \Phi_{m_a}(Z+W f(X); \Sigma) $ where $\Sigma \in \mathbb{R}^{m_a \times m_a}$ is symmetric positive definite, $W\in \mathbb{R}^{m_a \times n}$ and $Z\in \mathbb{R}^{m_a}$  are an arbitrary data dependent matrix
	and, respectively, vector.
	The posterior of $f(X)$ is a SUN:
	\begin{align}
	\label{eq:posteriorclass}
	p(f(X)|W,Z)&= \text{SUN}_{n,s+m_a}(\bxi_p,\Omega_p,\Delta_p,\bgamma_p,\Gamma_p)\\
	\bxi_p & =\bxi, \qquad \Omega_p = \Omega, \\
	\Delta_p &=[\Delta,~~\bar{\Omega}\Domega W^T],\\
	\bgamma_p& =[\bgamma,~~Z+W\xi]^T, \\
	\Gamma_p&=\begin{bmatrix}
	\Gamma & ~~\Delta^T \Domega W^T \\
	W \Domega \Delta & ~~(W \Omega W^T + \Sigma) \end{bmatrix},
	\end{align}
	where, for simplicity of notation, we denoted $\xi(X),\Omega(X,X),\Delta(X)$ as $\bxi,\Omega,\Delta $ respectively and
	$\Omega = \Domega \bar{\Omega} \Domega$.
\end{lemma}
In this case, the posterior is a skew normal distribution even  for either $s=0$ or
$\Delta(\bx)=0$. This shows that, for GP classification and GP preference learning, the true posterior is a skew normal distribution.\footnote{Since here we are basically considering a parametric setting (the posterior of $f$ computed at the training points), this lemma extends the results derived in \citep[Th.1 and Co.4]{durante2018conjugate} for the parametric case to general probit affine likelihoods  $\Phi_{m_a}(Z+W\bl[f];\Sigma)$. In the proofs of Theorem 1 and Corollary 4 in \citep{durante2018conjugate},  $\Gamma_p$ (posterior $\Gamma$) is rescaled to make it a correlation matrix. As explained in Section \ref{sec:unified}, we do not do it for simplicity of notation.}

\begin{corollary}
	\label{co:ml}
	Consider the probabilistic model 
	in Lemma \ref{lemma:Affine}, the marginal likelihood of the observations $Z,W$ is
	\begin{equation}
	p(Z,W) =  \frac{\Phi_{s+m_a}(\bgamma_p;~\Gamma_p) }{\Phi_{s}(\bgamma;~\Gamma)}.
	\label{eq:marginalLikelihoodAffine}
	\end{equation}
\end{corollary}

  We now  prove that,  a-posteriori, for a new test point $\bf x$, the function  $f(\bf x)$ is SkewGP distributed under the affine probit likelihood in \eqref{eq:affineProbit}.

\begin{theorem}
	\label{th:2}	
	Let us assume a SkewGP prior 	$f({\bf x}) \sim \text{SkewGP}_s(\xi({\bf x}), \Omega({\bf x},{\bf x}'),\Delta({\bf x}),\bgamma, \Gamma)$, the likelihood $ \Phi_{m_a}(Z+W f(X)) $ with $W\in \mathbb{R}^{m_a \times n}$ and $Z\in \mathbb{R}^{m_a}$, then a-posteriori $f$ is SkewGP with mean, covariance and skewness functions:
	\begin{align}
	\label{eq:predPostSample}
	\tilde{\bxi}({\bf x}) &=\xi(\bx)\\
	\nonumber
	\tilde{\Omega}({\bf x},{\bf x}) &= \Omega(\bx,\bx),\\
	\nonumber
	{\tilde{\Delta}({\bf x})} &=\begin{bmatrix}
\Delta(\bx)~ & D_{\Omega(\bx,\bx)}^{-1}\Omega(\bx,X)W^T
\end{bmatrix},
	\end{align} 
	 and parameters $\bgamma_p,\Gamma_p$ as in Lemma \ref{lemma:Affine}.
\end{theorem}
This results proves that SkewGP process and the probit affine likelihood are conjugate, which provides closed-form expression for the posterior for both classification and preference learning.



\begin{figure}[H]
\centering
 \includegraphics[width=6cm]{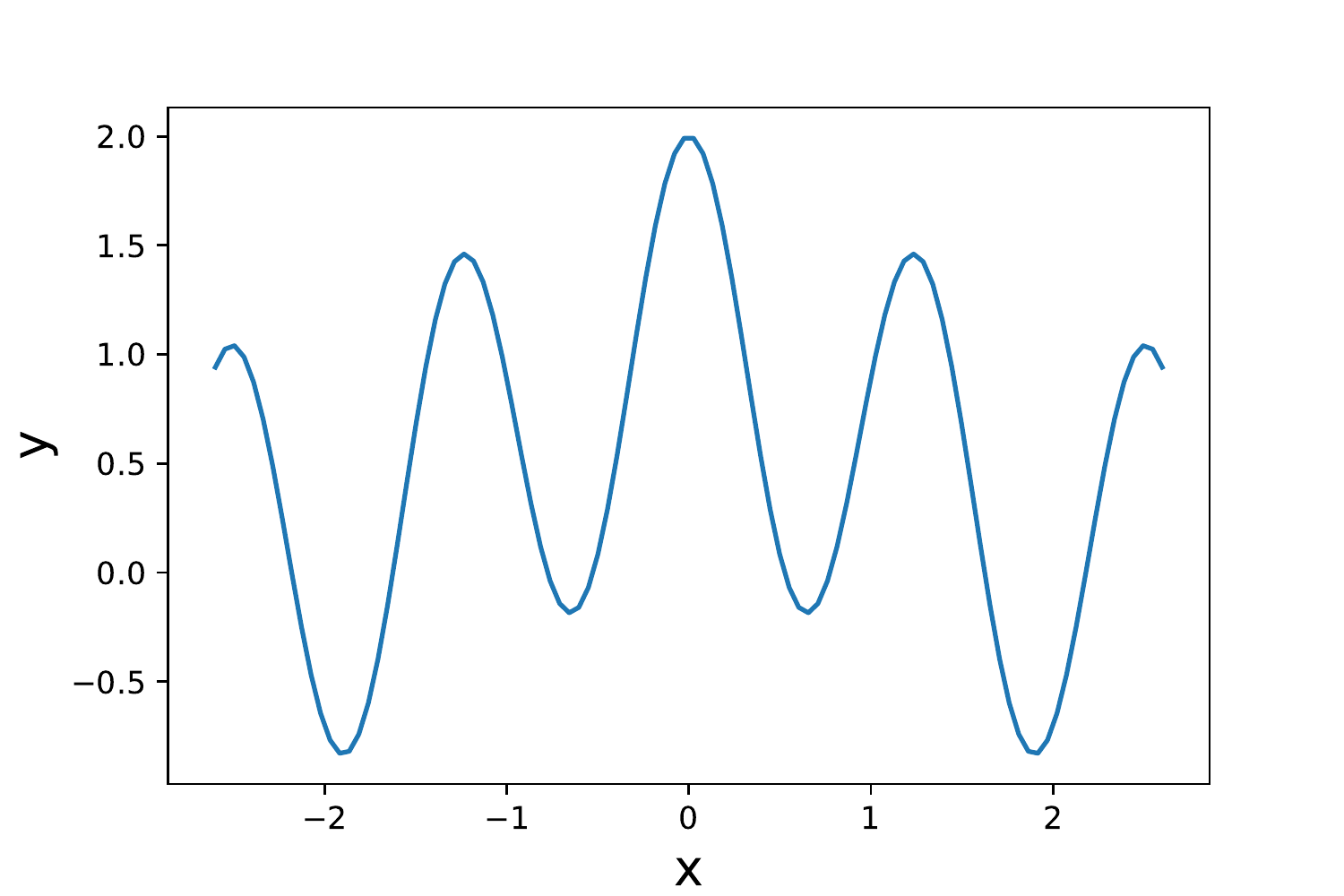}
 \caption{1D function $g(x)=cos(5x)+e^{-\frac{x^2}{2}}$}
  \label{fig:ex0}
\end{figure}

\paragraph{Example 1D preference learning.} In this paragraph, we  provide a  one-dimensional example of how we can use the above derivations  to compute the exact posterior for preference learning. Consider the  non-linear function $g(x)=cos(5x)+e^{-\frac{x^2}{2}}$ which is  in Figure \ref{fig:ex0}.
We use this function to generate 45 random pairwise preferences
between 25 random points $x_i \in [-2.6,2.6]$.
Our aim is to infer $f(x)$, that is the latent function that models the preference relations observed in the dataset. We will then use the learned model to compute $f(x)-f(0.05)$, which can tell us the points $x$ which are preferred to $x_r=0.05$ ($0.05$  is the point corresponding to the maximum value of $g$  in the dataset).

In all cases we will consider a GP prior with zero mean and radial basis function (RBF) kernel over the unknown function $f$. We will compare the exact posterior computed via SkewGP with two approximations: Laplace (LP) and Expectation Propagation (EP). We will discuss how to compute the predictions 
for SkewGP in Section \ref{sec:cdfs}.

\begin{figure}[htp!]
\centering
 \includegraphics[width=11.0cm]{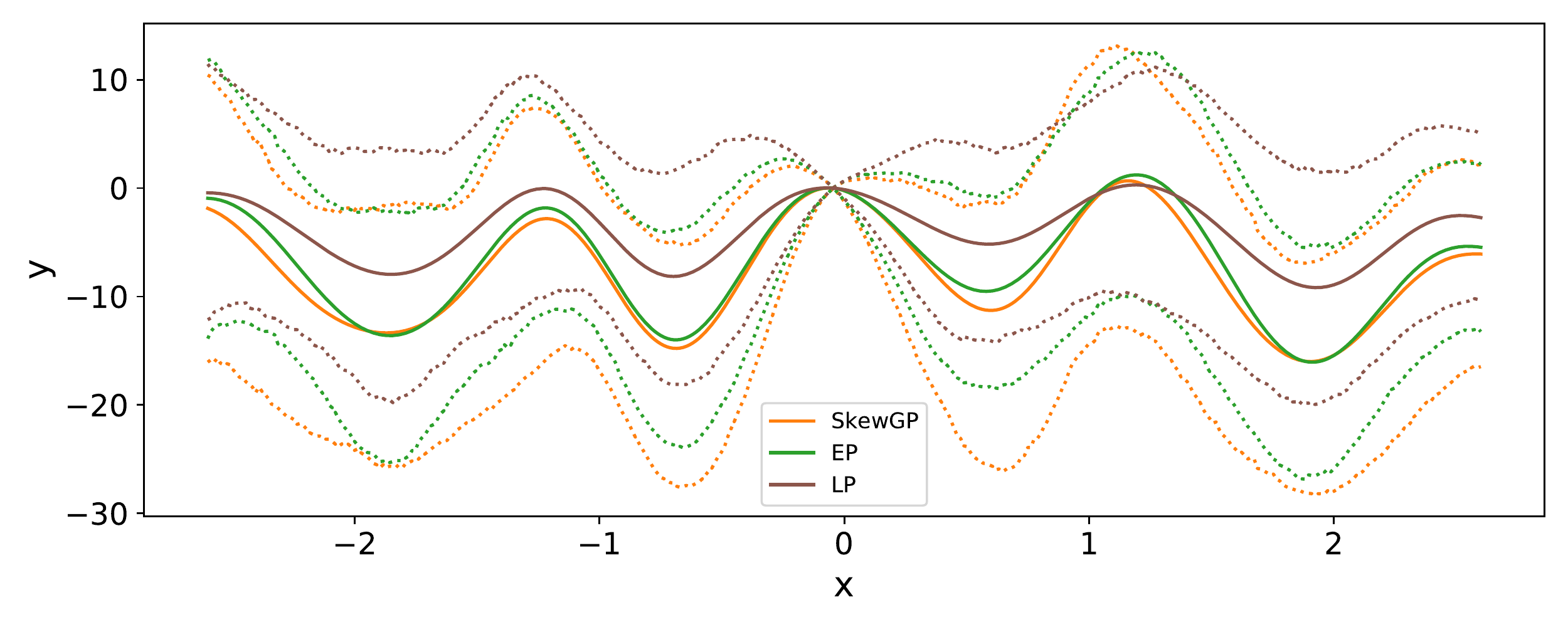}
  \includegraphics[width=5.5cm]{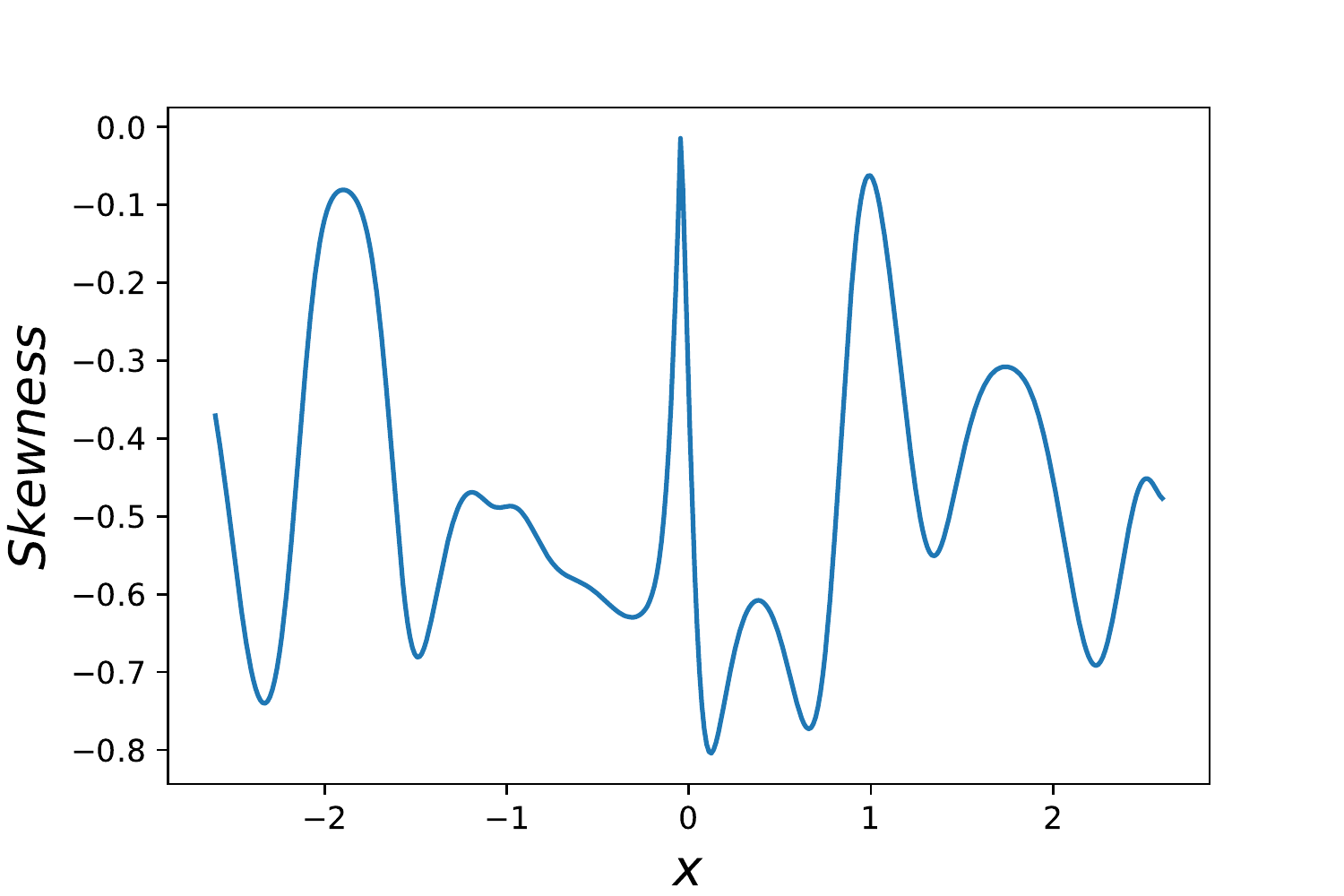}
   \includegraphics[width=5.5cm]{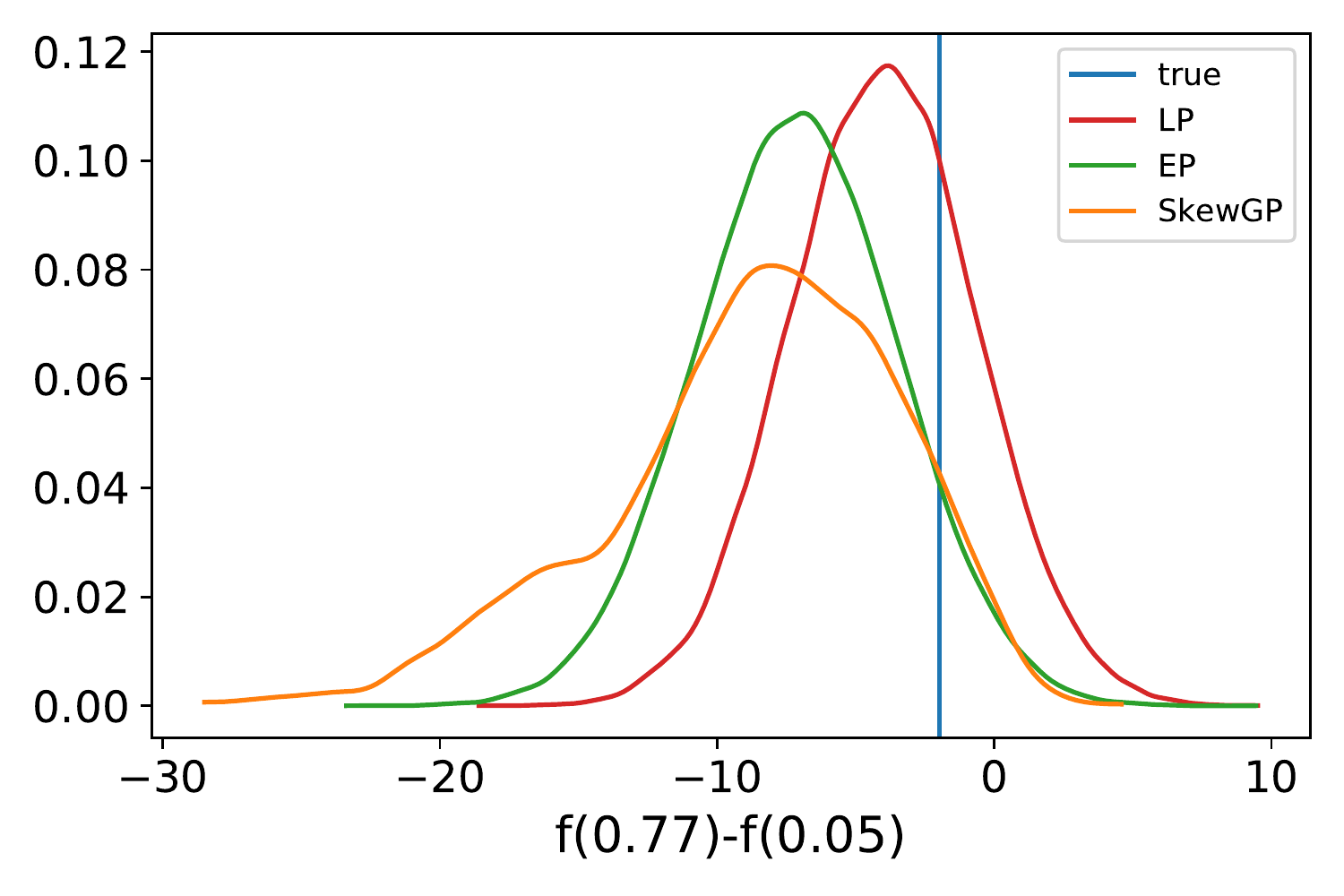}
   \caption{Preference learning. The top plot  shows the posterior means and credible intervals for LP, EP and SkewGP. 
   The bottom-left plot shows the skewness statistics for the SkewGP posterior as a function of $x$. The right plot the predictive distribution of $f(0.77)-f(0.05)$ for the three models.}
   \label{fig:ex1}
\end{figure}

Figure  \ref{fig:ex1}(top) shows the  predicted posterior distribution $f(x)-f(0.05)$ (and relative 95\% credible region) computed according to LP, EP and SkewGP. All the methods use the same prior: a GP with zero mean and RBF covariance function (the hyperparameters are the same: lengthscale $0.33$ and variance $\sigma^2=50$).
Therefore, the only difference between the  exact posterior (SkewGP) and the posteriors of EP and LP
is due to the different approximations. The true posterior (SkewGP) of the preference function is skewed as it can be seen in Figure \ref{fig:ex1}(bottom-left), which reports the \textit{skewness statistics} for  $h(x)=f(x)-f(x_r)$ as a function of $x$, defined as:
$$
SS(h(x)):={\tfrac {\operatorname {E} \left[(h(x)-\mu )^{3}\right]}{(\operatorname {E} \left[(h(x)-\mu )^{2}\right])^{3/2}}},
$$
with $\mu:=\operatorname {E} \left[h(x)\right]$, and $\operatorname {E}$ is computed  via Monte Carlo sampling  from the posterior.

The strong skewness of the posterior is the reason why both the LP and EP approximations are not able to correctly approximate the posterior as shown in Figure \ref{fig:ex1}(top).  Figure \ref{fig:ex1}(bottom-right) shows the predictive posterior distribution for $f(0.77)$ for the three models. It can be noticed that the true posterior (SkewGP) is  skewed, which explains why the LP and EP approximations are not accurate.

\subsection{Mixed likelihood}
Consider $n$ input points $X = \{\bx_i : i=1, \ldots, n\}$, with   $\bx_i \in \mathbb{R}^d$, and the product likelihood:  
\begin{equation}
p(Y,Z,W \mid f(X)) = \Phi_{m_a}(Z+Wf(X); \Sigma)\phi_{m_r}(Y-Cf(X);R).
\label{eq:mixedlike}
\end{equation}
where $C,R,Y,W,Z,\Sigma$ have the same dimensions as before. To obtain the posterior we  apply first Theorem \ref{th:2}
and then Theorem \ref{th:1}. 
\begin{theorem}
	\label{th:3}	
	Let us assume a SkewGP prior 	$f({\bf x}) \sim \text{SkewGP}_s(\xi({\bf x}), \Omega({\bf x},{\bf x}'),\Delta({\bf x}),\bgamma, \Gamma)$, the likelihood \eqref{eq:mixedlike}, then a-posteriori $f$ is SkewGP with mean, covariance and skewness functions:
		\begin{align}
	\tilde{\bxi}({\bf x})  &=\bxi({\bf x})+\Omega({\bf x},X) C^T(C\Omega(X,X) C^T+R)^{-1}(Y-C\xi(X)),\\
	\tilde{\Omega}({\bf x},{\bf x}) &= \Omega({\bf x},{\bf x})-\Omega({\bf x},X) C^T(C\Omega(X,X) C^T+R)^{-1}C\Omega(X,{\bf x}),\\
	\nonumber
	\nonumber
	\tilde{\Delta}({\bf x}) &=D_{\tilde{\Omega}(\bx,\bx)}^{-1}D_{\Omega(\bx,\bx)} \begin{bmatrix}
\Delta(\bx)~ & D_{\Omega(\bx,\bx)}^{-1}\Omega(\bx,X)W^T
\end{bmatrix}\\
	&-D_{\tilde{\Omega}(\bx,\bx)}^{-1}\Omega({\bf x},X)C^T (C\Omega(X,X) C^T+R)^{-1} C D_{\Omega(X,X)} \begin{bmatrix}
\Delta(X)~ & D_{\Omega(X,X)}^{-1}\Omega(X,X)W^T
\end{bmatrix},\\
\tilde{\bgamma} &= \bgamma_p+\begin{bmatrix}
\Delta(X)~ & D_{\Omega(X,X)}^{-1}\Omega(X,X)W^T
\end{bmatrix}^T\bar{\Omega}(X,X)^{-1}D_{\Omega(X,X)}^{-1}(\tilde{\bxi}(X)-\bxi(X))\\
\nonumber
\tilde{\Gamma}& =  \Gamma_p- \begin{bmatrix}
\Delta(X)~ & D_{\Omega(X,X)}^{-1}\Omega(X,X)W^T
\end{bmatrix}^T\bar{\Omega}^{-1}(X,X)\begin{bmatrix}
\Delta(X)~ & D_{\Omega(X,X)}^{-1}\Omega(X,X)W^T
\end{bmatrix}\\
&+\Delta_p^T\tilde{\Omega}(X,X)^{-1}\Delta_p,
	\end{align}
with 
\begin{align}
\Delta_p&=\tilde{\Omega}(X,X)D_{\tilde{\Omega}(X,X)}D_{\Omega(X,X)}^{-1} \bar{\Omega}^{-1}(X,X)\begin{bmatrix}
\Delta(X)~ & D_{\Omega(X,X)}^{-1}\Omega(X,X)W^T,
\end{bmatrix}
\end{align}
and  $\bgamma_p,\Gamma_p$ as in Lemma \ref{lemma:Affine}.
\end{theorem}

\begin{corollary}
\label{co:Mixed}
The marginal likelihood of the observations $Y$ is
 \begin{equation}
  \label{eq:ml_normal_mix}
  p(Y)=\phi_{m_r}(Y-C\xi(X);C\Omega(X,X)C^T+R)\frac{ \Phi_{s+m_a}(\tilde{\bgamma};~\tilde{\Gamma})}{\Phi_{s}(\bgamma;~\Gamma)}.
 \end{equation}
 with $\tilde{\bgamma},\tilde{\Gamma}$ in
 Theorem \ref{th:3}.
\end{corollary}

This general setting shows that SkewGPs are conjugate with a larger class of likelihoods and, therefore, encompass GPs.

In the next paragraphs we provide two one-dimensional examples of (i) mixed numeric and preference regression; (ii) mixed numeric and binary output regression.

\paragraph{Example 1D mixed numeric and preference regression}
 Consider again the regression problem discussed in Section \ref{sec:intro} and the dataset shown in Figure \ref{fig:1}.

\begin{figure}[htp!]
\centering
 \includegraphics[width=11cm]{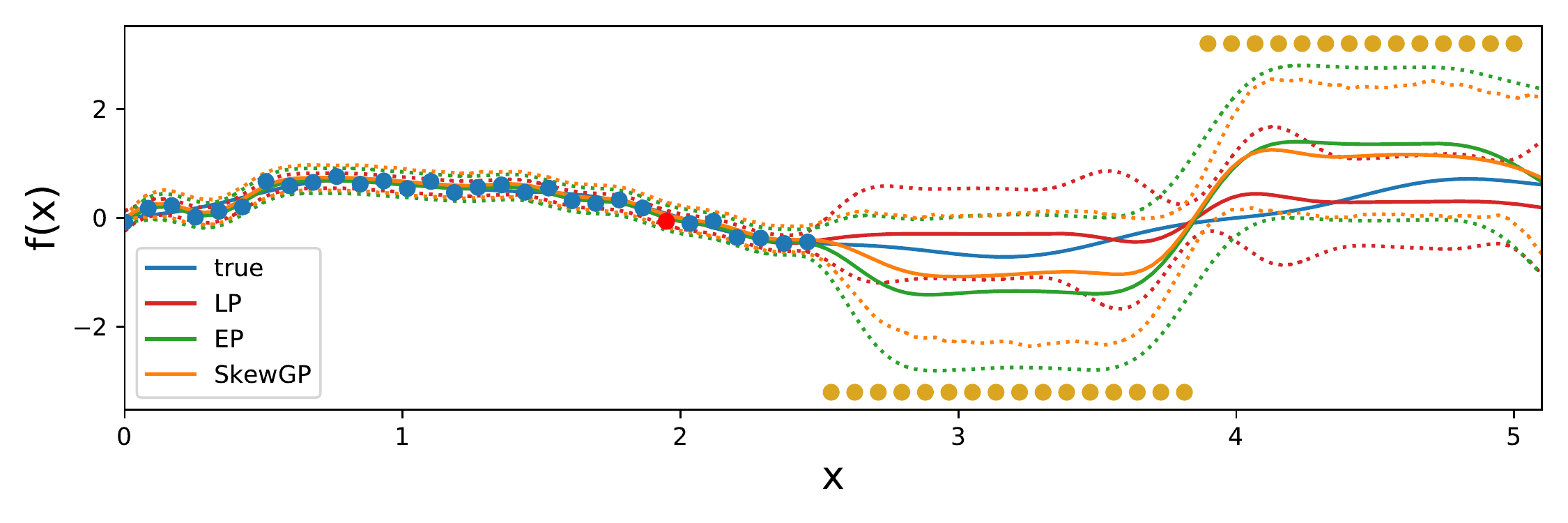}
    \includegraphics[width=5.5cm]{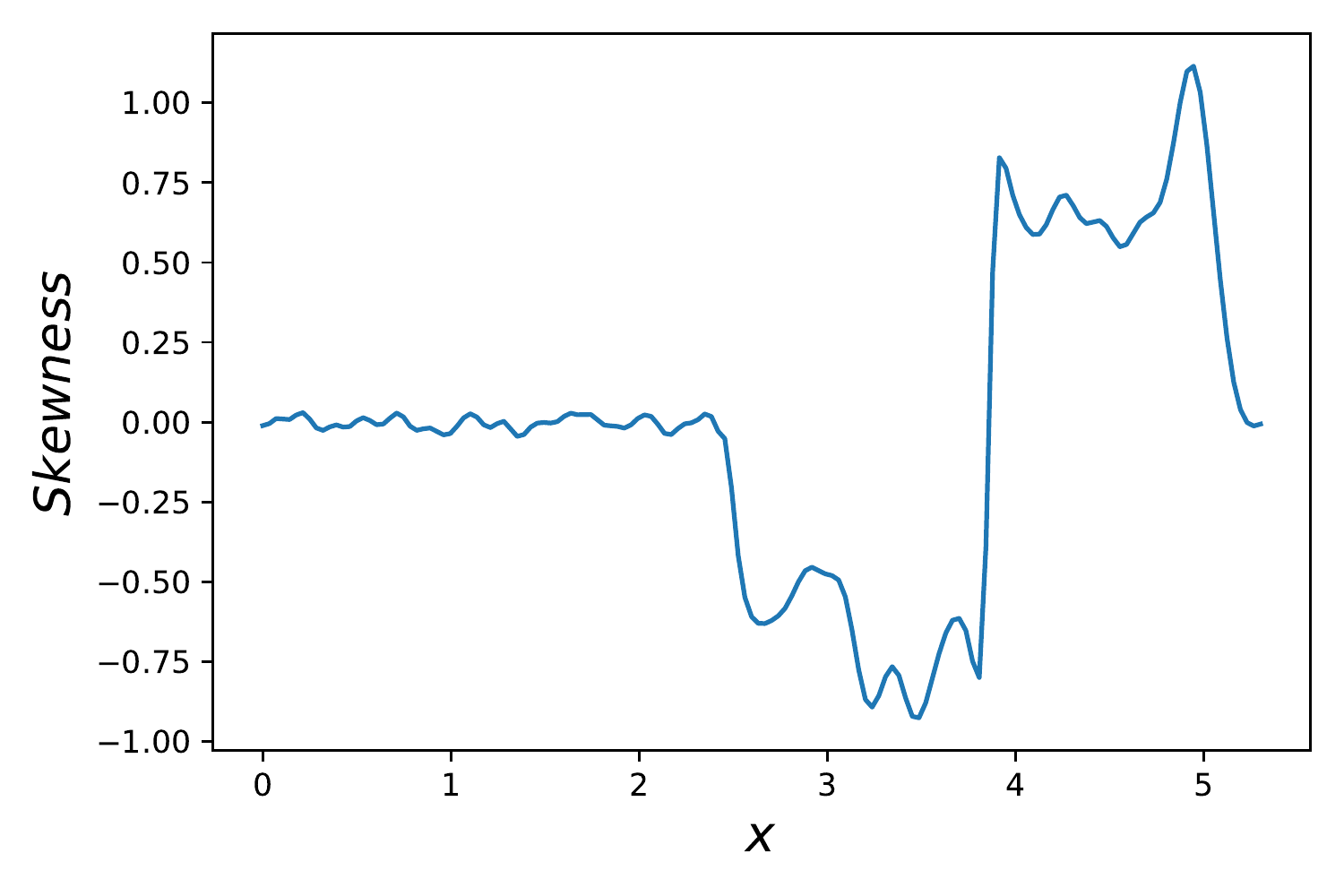}
   \includegraphics[width=5.5cm]{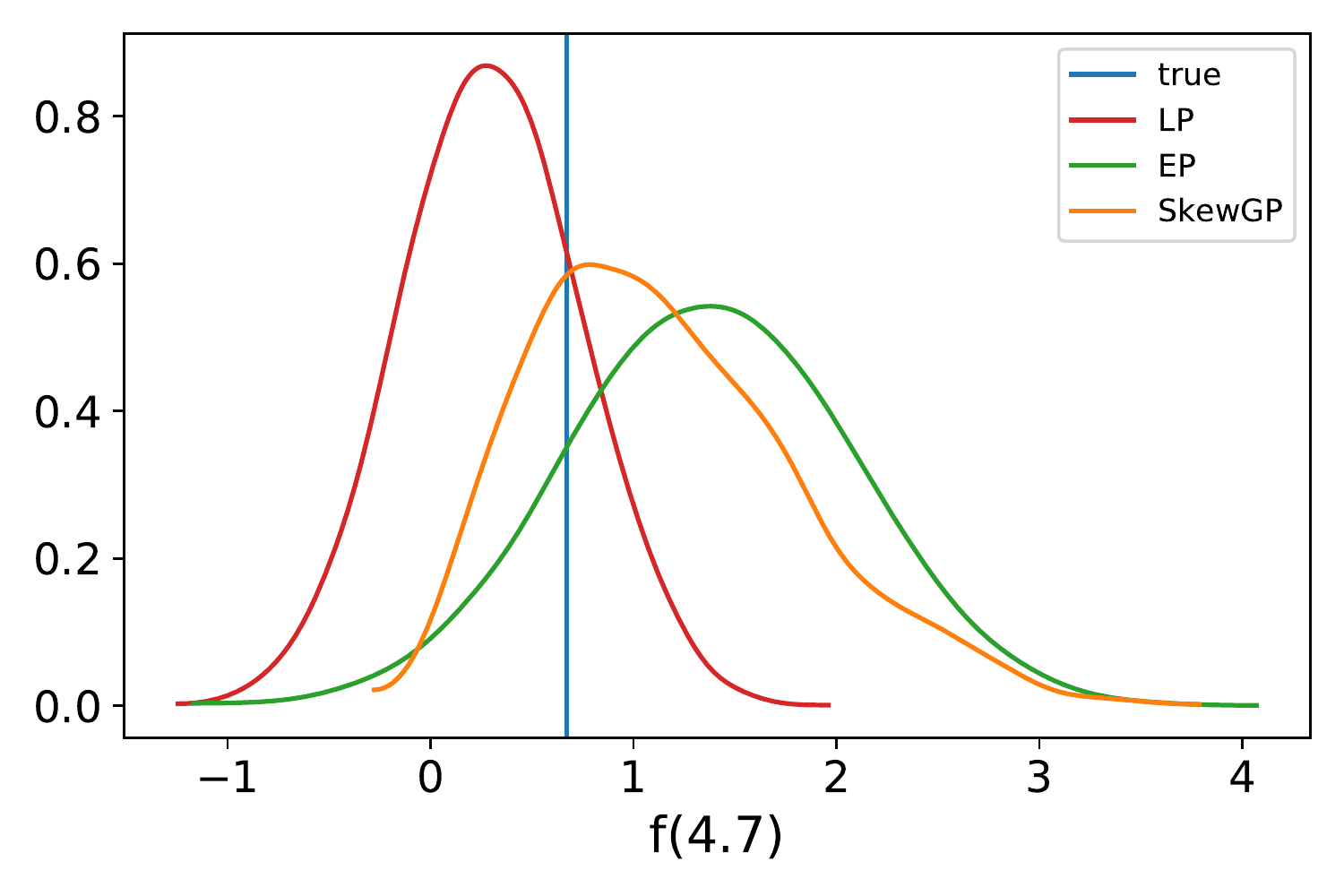}
   \caption{Mixed numeric and preference learning. Top: dataset, true function (blue), mean and credible intervals for LP, EP, SkewGP. Bottom-left: skewness statistics for the SkewGP posterior as a function of $x$. Bottom-right: predictive distribution of $f(4.7)$ for the three models.}
   \label{fig:ex2}
\end{figure}

Figure \ref{fig:ex2}(top) shows the posterior mean and relative credible region of the regression function $f$ computed according to  SkewGP, LP and EP. All the methods use the same prior: a GP with zero mean and RBF covariance function (the hyperparameters are the same: lengthscale $0.19$, variance $\sigma^2=1$ and noise variance $\sigma^2=0.001$).

Figure \ref{fig:ex2}(bottom-left) reports the skewness statistics for SkewGP and Figure \ref{fig:ex2}(bottom-right) the predictive posterior distribution for $f(4.7)$ for the three models. It can be noticed that the true posterior (SkewGP) is very skewed for $x\geq 2.5$, which explains why the LP and EP approximations are not accurate.
The LP approximation heavily
 underestimates the mean and the support of the true posterior (SkewGP), as also evident from Figure \ref{fig:ex2}(bottom-right). The EP approximation estimates a large variance to ``fit'' the skewed posterior. This confirms, also for the mixed case, what was observed by \cite{kuss2005assessing,nickisch2008approximations} for classification.
 The symmetry assumption for the posterior for LP and EP affects the coverage of their credible intervals (regions). 
 
 It can be noticed that the three models coincide for $x< 2.5$, that is in the region including the numeric observations. The SkewGP posterior is indeed symmetric in this region, as it can be seen from Figure \ref{fig:ex2}(bottom-left).

\begin{figure}[htp!]
\centering
 \includegraphics[width=12cm]{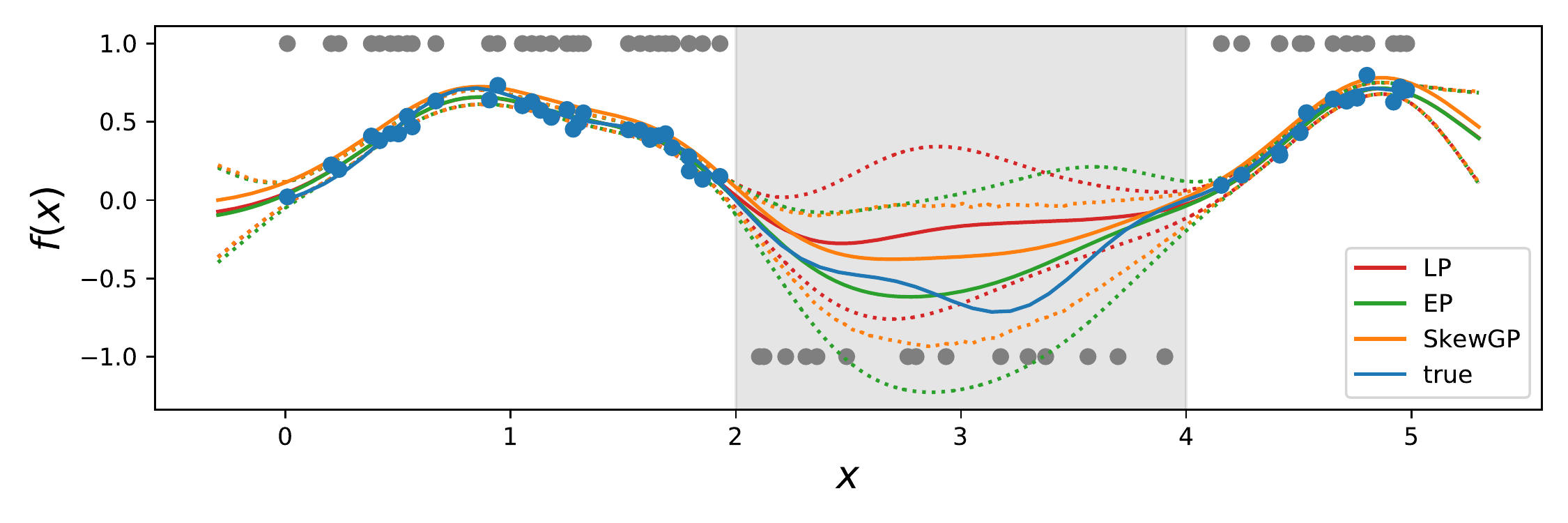}
    \includegraphics[width=5.5cm]{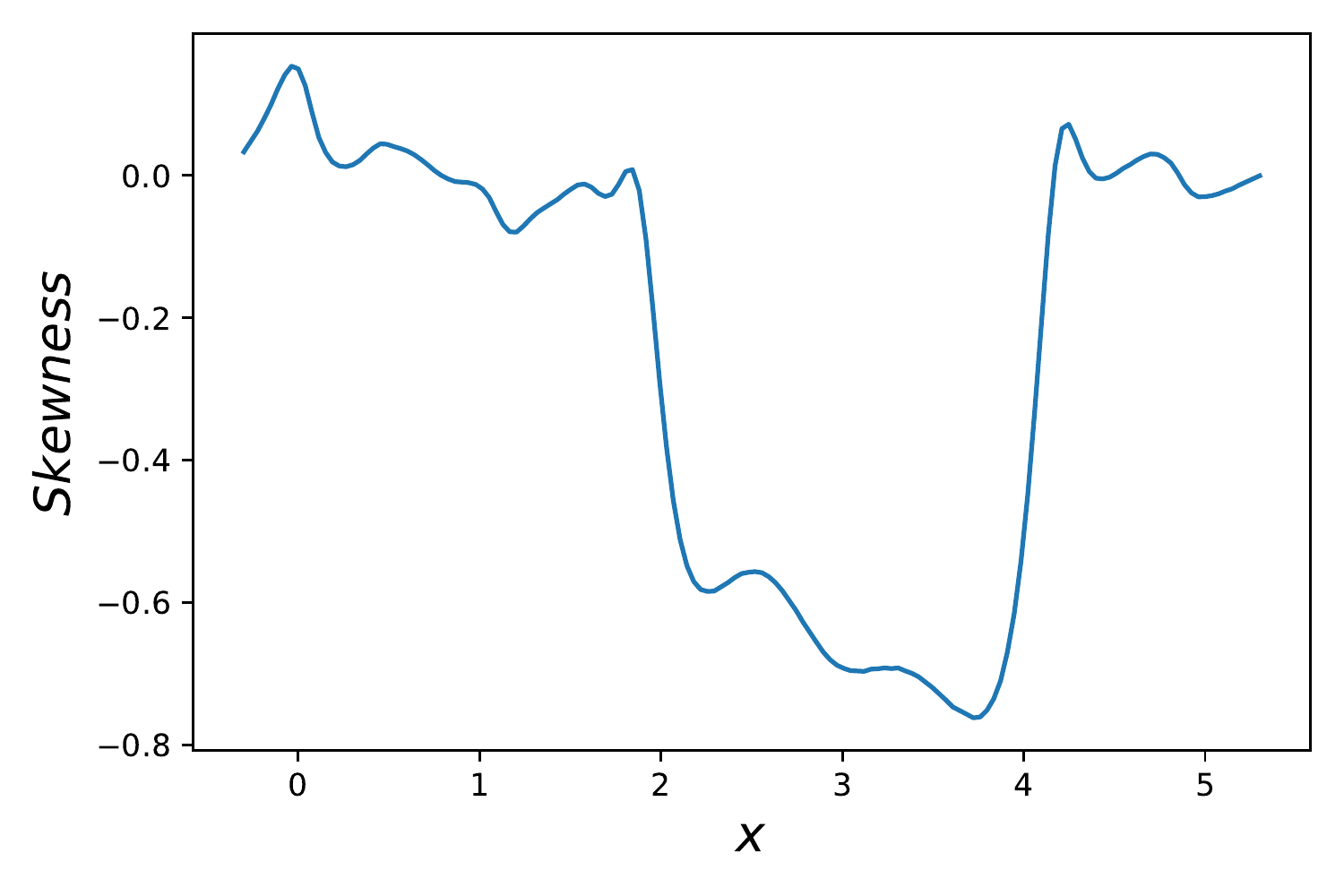}
   \includegraphics[width=5.5cm]{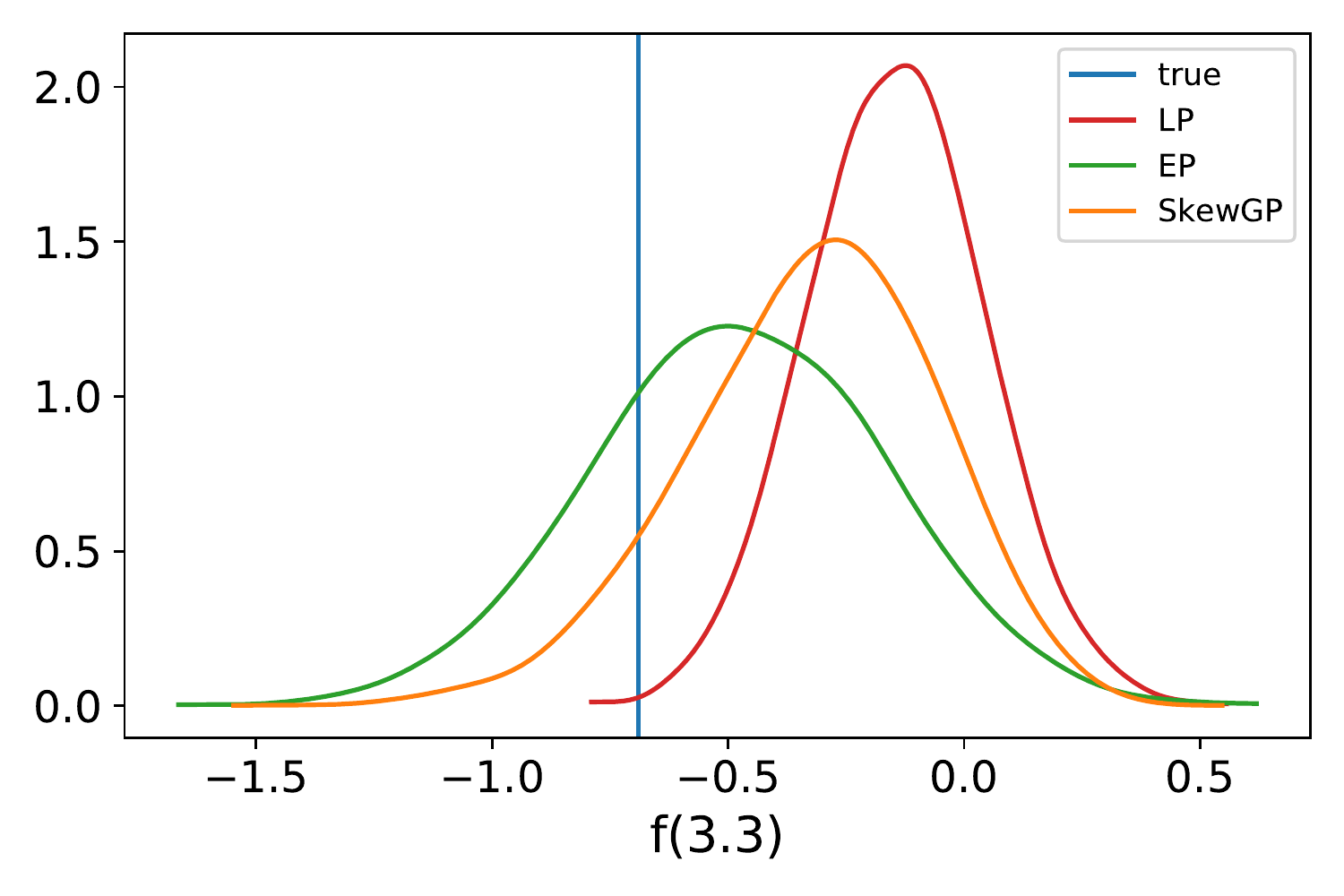}
   \caption{Mixed numeric and binary regression. Top: dataset, true function (blue), mean and credible intervals for LP, EP, SkewGP. Bottom-left: skewness statistics for the SkewGP posterior as a function of $x$. Bottom-right: predictive distribution of $f(3.3)$ for the three models.}
   \label{fig:ex3}
\end{figure}

\paragraph{Example 1D mixed numeric and binary regression}
   This mixed problem arises when binary judgments (valid or non-valid) together with scalar observations $f(x)$  are available.
Such situation often comes up in industrial applications. For example imagine a process $f$ that produces a certain noisy output $f(x)+v$ which depends on input parameters $x$. Given a certain input $x_k$, assume now that when $f(x_k) +v \leq h$   no output is produced. In this case,  observations are   pairs and the space of possibility is $\{(\text{valid},y),~~(\text{non-valid},None)\}.
$ with $y=f(x)+v$.
The  above setting can be modelled by the likelihood:
\begin{equation}
 \label{eq:likevalid}
\left\{\begin{array}{ll}
\phi\left(\frac{y-f(x)}{\sigma_v}\right)\Phi\left(\frac{f({x})-h}{\sigma_v}\right), &    (\text{valid},y)\\                    
\Phi\left(\frac{h-f({x})}{\sigma_v}\right), &    (\text{non-valid},None),\\
\end{array}\right.
\end{equation}
and, therefore, formulated as a mixed regression problem.\footnote{In \citep{benavoli2020preferential}, we considered mixed preferential-binary observations,  where binary judgments (valid or non-valid) together with preference judgments are available.} Figure \ref{fig:ex3}(top) shows the true function we used to generate the data (in blue), the numeric data (blue points) and  binary data (gray points where $y_1=1$ means valid and  $y_i=-1$ means non-valid). The threshold $h$ has been set to $h=0$ and so the non-valid zone is the region in gray.
We report the posterior means and credible regions of LP, EP and SkewGP. All the methods use the same prior: a GP with zero mean and Radial Basis Function (RBF) covariance function (the hyperparameters were estimated using EP: lengthscale $0.497$, variance $\sigma^2=0.147$ and noise variance $\sigma^2=0.0021$).
Figure \ref{fig:ex3}(bottom-left) reports the skewness statistics for SkewGP and Figure \ref{fig:ex3}(bottom-right) the predictive posterior distribution for $f(3.3)$ for the three models. It can be noticed that the true posterior (SkewGP) is very skewed in the non-valid zone ($2 \leq x\leq 4$), which explains why the LP and EP approximations are not accurate. Again the LP approximation heavily  underestimates the mean and the support of the true posterior (SkewGP) and the EP approximation estimates a large variance to ``fit'' the skewed posterior.

\section{Sampling from the posterior and hyperparameters selections}
\label{sec:cdfs}

The computation of predictive inference (mean, credible intervals etc.) is achieved by sampling the posterior SkewGP.
This can be obtained using the additive representation discussed in Section \ref{sec:Additive}. Recall that $\mathbf{z} \sim \operatorname{SUN}_{p,s}(\bxi,\Omega, \Delta, \bgamma, \Gamma)$ can be written as $\mathbf{z} = \bxi + \Domega(\mathbf{r}_0 + \Delta \Gamma^{-1}\mathbf{r}_{1,-\gamma})$ with $\mathbf{r}_0\sim \phi_p(0; \bar{\Omega}-\Delta \Gamma^{-1}\Delta^T)$ and $\mathbf{r}_{1,-\gamma}$ is the truncation below $\gamma$ of $\mathbf{r}_{1} \sim \phi_s(0;\Gamma)$. Note that sampling $\mathbf{r}_0$ can be achieved efficiently with standard methods, however using standard rejection sampling for the variable $\mathbf{r}_{1,-\gamma}$ would incur in exponentially growing sampling time as the dimension $m_a$ increases.
In \citep{Benavoli_etal2020,benavoli2020preferential} we used the recently introduced sampling technique \emph{linear elliptical slice sampling} (\emph{lin-ess},\citet{gessner2019integrals}) which improves Elliptical Slice Sampling (\emph{ess}, \citet{pmlrv9murray10a}) for multivariate Gaussian distributions truncated on a region defined by linear constraints. In particular this approach derives analytically the acceptable regions on the elliptical slices used in \emph{ess} and guarantees rejection-free sampling. We report the pseudo-code for \emph{lin-ess} in Algorithm~\ref{alg:LinEss}.\footnote{
The pseudo-code reports the main steps of the algorithm. We omitted some implementation details for simplicity, for instance how to select ${\bf x}_0$ or how to deal with empty  intersections. The full code of the algorithm is available at \url{https://github.com/benavoli/SkewGP}.}
Since \emph{lin-ess} is rejection-free,\footnote{Its computational bottleneck is 
 the Cholesky factorization of the covariance matrix $\tilde{\Gamma}$, same as for sampling from a multivariate Gaussian.} we can compute exactly (deterministically)  the computation complexity of \eqref{eq:additiveSUN}: 
$O(n^3)$ with storage demands of $O(n^2)$. SkewGPs have similar bottleneck computational complexity of full GPs.
Finally, it is \textbf{important} to observe that  $\mathbf{r}_{1,-\gamma}$ does not depend on test point ${\bf x}$
and, therefore, we do not need to re-sample $\mathbf{r}_{1,-\gamma}$ to sample
$f$ at another test point ${\bf x}'$. 
This is fundamental in active learning and Bayesian optimisation because   acquisition functions are functions of ${\bf x}$ and we need to optimize them with respect to ${\bf x}$.

\begin{algorithm}[H]
	\caption{Lin-Ess to sample from $\phi_s(\mu;\Gamma)$ truncated below $\gamma$}
	\SetKwInOut{Input}{input}\SetKwInOut{Output}{output}
	
	\Input{$\Gamma \in \mathbb{R}^{s\times s}$ symmetric p.d., $\mu, \gamma \in \mathbb{R}^s$, $m>0$}
	\Output{$\mathbf{x}_{sample} \in \mathbb{R}^{m \times s}$, $m$ samples from the $\phi_s(0;\Gamma)$ truncated below $\gamma$.}
	
	$L = \operatorname{Cholesky}(\Gamma)$\;
	select $\mathbf{x}_0$ s.t. $L\mathbf{x}_0-\gamma>0$\;
	\While{$i < m$}{
		sample $\nu \sim \phi_s(0;\Gamma)$\;
		compute $r^2 =  \mathbf{x}_0^2 + \nu^2$\;
		compute the angles ${\bm \theta}_1 = 2\arctan(\frac{\nu-\sqrt{r^2-\gamma^2}}{\mathbf{x}_0+\gamma})$ and ${\bm \theta}_2 = 2\arctan(\frac{\nu+\sqrt{r^2-\gamma^2}}{\mathbf{x}_0+\gamma})$\;
		find the largest intesection angles in the domain boundary $[\theta_{min},\theta_{max}]$\;
		sample $\theta \sim U(\theta_{min},\theta_{max})$\;
		compute $\mathbf{x}_{i, sample} = \mathbf{x}_0\cos(\theta)+ \nu\sin(\theta)$\;
        set $\mathbf{x}_0=\mathbf{x}_{i, sample}$\;
		i = i+1\;
	}
	\label{alg:LinEss}
\end{algorithm}

	We demonstrate this computational savings in a 1D classification task -- probit likelihood and GP prior on the latent function. We generated 8 datasets of size $$
	n\in \{1000,1500,2000,2500,3000,3500,4000, 4500\}
	$$ 
	as follows ${\bf x}_i \sim N(0,I_3)$ for $i=1,\dots,n$, $\sigma^2 \in \mathcal{U}(1,10)$, $\ell_j \in \mathcal{U}(0.1,1.1)$ for $j=1,\dots,3$, the latent function $f$ is sampled from a GP with zero mean and Radial Basis Function (RBF) kernel:
	\begin{equation}
	 \label{eq:RBF}
	\Omega({\bf x},{\bf x}') := \sigma^2 \exp \left(-\frac {1}{2}({\bx} -{\bx'})^T\begin{bmatrix}
	\tfrac{1}{\ell^2_1} &0 &0\\
	0 & \tfrac{1}{\ell^2_2} &0\\
	0 & 0 & \tfrac{1}{\ell^2_3}	                                                                              \end{bmatrix}                                                                               
({\bx} -{\bx'})\right),
	\end{equation}
and $y_i \sim \text{Bernoulli}(\Phi(f(x_i)))$ for $i=1,\dots,n$.

We  evaluated the computational load to approximate the posterior using Laplace's method (GP-LP), Expectation Propagation (GP-EP), Hamiltonian Monte Carlo (GP-HMC) and Elliptical Slice Sampling (GP-ess).\footnote{For GP-LP and GP-EP, we use the implementations in GPy \citep{gpy2014}, while for GP-HMC and GP-Ess we implemented the probabilistic model in PyMC3 \citep{salvatier2016probabilistic} using the default values for  acceptance rate. Note that,
for GP-HMC and GP-Ess, we apply the Monte Carlo methods directly to the probabilistic
model comprising the probit likelihood and the GP prior on the latent function $f$. In other words we do not exploit the fact that the posterior process is SkewGP.}
We  compared these four approaches against our SkewGP  formulation of the posterior (denoted as SkewGP-LinEss), which allows us exploit the decomposition discussed above to speed-up sampling time.
For each value of $n$, we generated 3 datasets and averaged the running time to: (i) compute $3100$ samples (for  SkewGP-liness, GP-HMC and GP-ess) using the first $100$ samples for tuning; (ii) analytically approximate the posterior for GP-LP and GP-EP.\footnote{Note that, all methods use the true kernel hyperparameters which are kept fixed.}

\begin{figure}[htp]
	\centering
	\includegraphics[height=3.5cm]{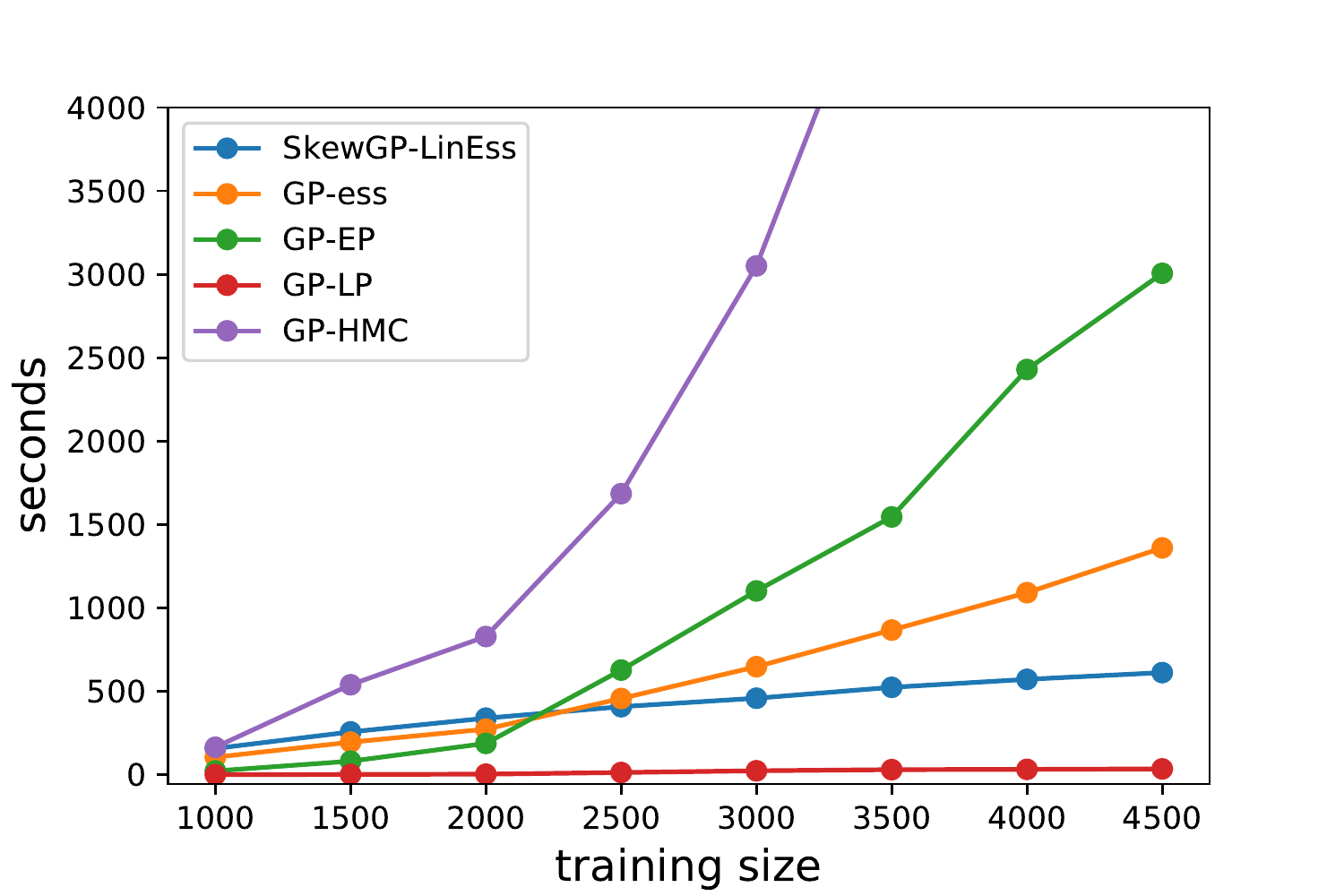}
		\includegraphics[height=3.5cm]{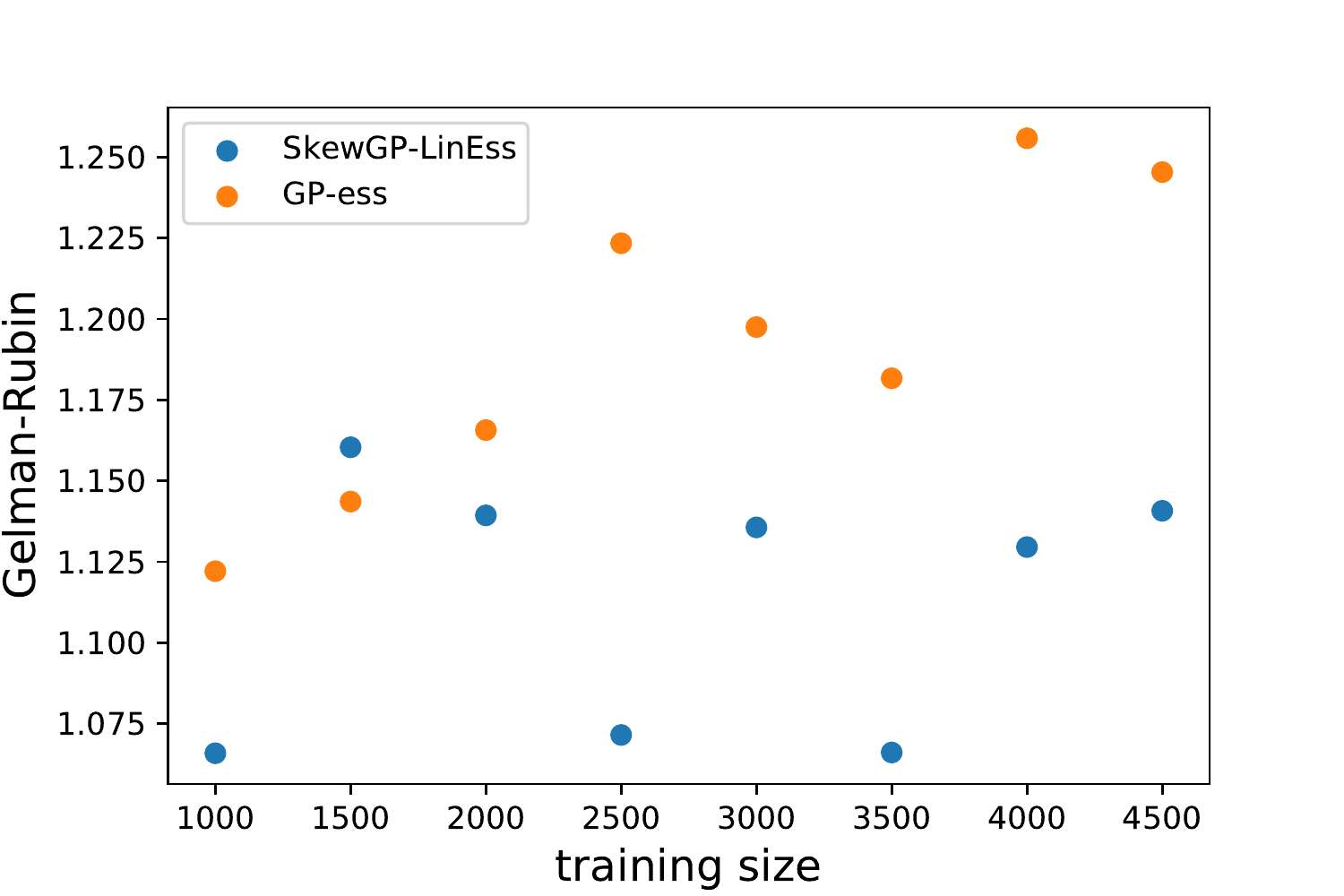}
	\caption{Left: running time for sampling from the posterior latent function at the training point  as a function of the number of training points for the five methods. Right: Gelman-Rubin statistics for GP-ess and SkewGP-ess.}
	\label{fig:comparsiontime}
    \end{figure}

The computational time in seconds (on a standard laptop) for the 5 methods is shown in Figure \ref{fig:comparsiontime}(left).
It can be noticed that GP-HMC is the slowest method and
GP-LP is the fastest; however it is well-known that GP-LP provides a  bad approximation of the posterior.
Among the remaining three methods, SkewGP-LinEss
is the fastest for $n>2200$.
The computational cost of drawing one sample using LinEss
 is $O(n^2)$. This is comparable with that of ess. However, for large $n$, the constant can
be much smaller for Lin-Ess because, contrarily to ess, Lin-Ess does not need to compute the likelihood at each iteration.This is also the main reason why both GP-EP and GP-ess become much slower at the increase of $n$.  It is \textbf{important} to notice that this computational advantage derives by our analytical formulation of the posterior process as a SkewGP.

Our goal was only to compare the running time to obtain $3000$ samples but we also  checked the convergence of the posterior. Figure \ref{fig:comparsiontime}(right)
reports the Gelman-Rubin (GR) \citep{gelman1992inference} statistics for GP-ess and SkewGP-LinEss.\footnote{To compute the Gelman-Rubin statistics, we have sampled an extra chain ($3100$ samples) in each experiment.}   \citep{gelman1992inference} suggest that GR values greater than $1.2$ should indicate nonconvergence.
The GR statistics is below $1.2$ for SkewGP-LinEss, while for GP-ess is greater than $1.2$ for $n>2500$. This means that  GP-ess needs additional tuning for $n>2500$ (so it is even slower). 
 
 It is also worth to notice that, for SkewGP, two alternative approaches to approximate the posterior mean  have been proposed in \citep{cao2020scalable}: (i)  using a tile-low-rank Monte Carlo methods for computing multivariate Gaussian probabilities; (ii) approximating functionals of multivariate truncated normals which via a mean-field variational approach.  
The cost of the tile-low-rank version of is  expensive in high dimensional setting \citep{cao2020scalable}, while the variational approximation scales up to large $n$.
The drawback of the variational approximation is that it is query dependent (it approximates an expectation). Conversely, the \textit{lin-ess} based approach is both efficient and agnostic of the query.
\paragraph{Hyperparameters selection}
In the GP literature \citep{rasmussen2006gaussian}, a GP is usually parametrised with a zero mean function $\xi({\bf x})=0$ and a  covariance kernel $\Omega({\bf x},{\bf x}')$ which depends on hyperparameters $\theta \in \Theta$. Typically $\theta$ contains lengthscale parameters and a variance parameter.
For instance, the RBF kernel in \eqref{eq:RBF}
has ${\boldsymbol \theta}=[\ell_1,\ell_2,\ell_3,\sigma,\sigma_v]$, where $\sigma_v^2$ is the variance of the measurement noise.
To fully define a SkewGP, we must also select the latent dimensions $s$, the additional parameters ${\bf \gamma},\Gamma$ and the skewness function $\Delta({\bf x})$. The function $\Delta({\bf x})$ may also have hyperparameters, which we denote by ${\bf u}$,  so that 
  ${\boldsymbol \theta}=[{\boldsymbol \ell},\sigma,\sigma_v,{\bf \gamma},\Gamma,{\bf u}]$.
  A choice of  ${\bf \gamma},\Gamma,{\bf u}$ is discussed in \citep{Benavoli_etal2020} for classification. 
 
 The parameters ${\boldsymbol \theta}$ are chosen by maximizing the marginal likelihood, that for the general mixed problem \eqref{eq:ml_normal_mix}, involves computing the CDF $\Phi_{s+m_a}$ of a multivariate normal distribution, whose dimensions  grows with $m_a$.
Quasi-randomized Monte Carlo methods \citep{genz1992numerical,genz2009computation,botev2017normal}  have been proposed to calculate $\Phi_{s+m_a}$ for small $m_a$ (few hundreds observations).
These approaches  are not in general suitable 
for medium and large $m_a$ (apart from special cases
\citet{phinikettos2011fast,genton2018hierarchical,azzimonti2018estimating}). 
Another approach is to use EP approximations as in  \citep{cunningham2013gaussian}, which has a similar computational load of GP-EP for learning the hyperparameters.

We overcome this issue by using  the  approximation introduced in \cite[Prop.2] {Benavoli_etal2020}:
 
\begin{proposition}
	\label{prop:approx}
Lower bound of the CDF:
	\begin{equation}
 \Phi_{m}(\tilde{\bgamma};~\tilde{\Gamma}) ~~~\geq  \sum_{i=1}^{b} \Phi_{|B_i|}(\tilde{\bgamma}_{B_i};~\tilde{\Gamma}_{B_i})-(b-1),
	\label{eq:marginalLikelihoodapprox}
	\end{equation}
\end{proposition}
where $B_1,\dots,B_b$ are a partition of the training dataset  into $b$ random disjoint subsets, $|B_i|$ denotes the number of observations  in the i-th element of the partition, 
$\tilde{\bgamma}_{B_i},~\tilde{\Gamma}_{B_i}$ are the parameters
of the posterior computed using only the subset $B_i$ of the data (in the experiments $|B_i|=70$).

To compute $\Phi_{|B_i|}(\cdot)$  in \eqref{eq:marginalLikelihoodapprox}, we use the routine
proposed in \cite{trinh2015bivariate}, that computes multivariate normal probabilities using bivariate conditioning.  This is very fast (fractions of second for $|B_i|=70$) and returns a deterministic quantity.
We therefore optimise  the hyperparameters of the kernel by maximising the lower bound in \eqref{eq:marginalLikelihoodapprox} using the  \textit{Dual Annealing} optimization routine.\footnote{For  local search, we use the  L-BFGS algorithm with numerical computation of the gradient.}
%

Another possible way to learn the hyperparameters is to
exploit the following result.
\begin{proposition}
\label{th:derivative}
 Assume that $\tilde{\bgamma}$ does not depend on $\theta_i$\footnote{The result can easily be extended to the case where $\tilde{\bgamma}$ depends on $\theta_i$ by using a change of variables.}, the derivative of $\log  \Phi_{m}(\tilde{\bgamma};~\tilde{\Gamma})$ with respect to $\theta_i$ is
 $$
 \frac{\partial}{ \partial \theta_i}\log\Phi_{m}(\tilde{\bgamma};~\tilde{\Gamma})=-\frac{1}{2}\text{Tr}\left( \tilde{\Gamma}^{-1}\tilde{\Gamma}'_{\theta_i}\right)+\frac{1}{2}\text{Tr}\left(\tilde{\Gamma}^{-1}\tilde{\Gamma}'_{\theta_i}\tilde{\Gamma}^{-1} N\right),
 $$
 where $\tilde{\Gamma}'_{\theta_i}=\frac{\partial}{ \partial \theta_i}\tilde{\Gamma}$
and $N= \frac{1}{\Phi_{m}(\tilde{\bgamma};~\tilde{\Gamma})}\iint_{-\infty}^{\tilde{\bgamma}} {\bf z}{\bf z}^T \frac{1}{|2 \pi {\Gamma}|}e^{-\frac{1}{2}{\bf z}^T \tilde{\Gamma}^{-1} {\bf z}} d{\bf z}$.
\end{proposition}
Note that, $M$ is the second order moment matrix of a multivariate truncated normal. We plan to investigate the use of (batch) stochastic gradient descent methods based on Proposition \ref{th:derivative} to learn the hyperparameters on future work. In the next sections, we will use the approximation in \eqref{eq:marginalLikelihoodapprox} to optimise the hyperparameters.

\section{Application to active learning and optimisation}
\label{sec:applications}
The ability of providing a calibrated measure of uncertainty is fundamental for a Bayesian model.
In the previous one-dimensional examples, we showed that the true posterior (SkewGP) can be highly skewed and so very different from LP and EP's posteriors. 
LP tends underestimating the uncertainty and EP tends overestimating it. Both methods are not able to capture the skewness (asymmetry) of the posterior which, in turn,  can lead to significant differences in the computed posterior means.
In this section we will compare LP, EP and SkewGP in two tasks, \textit{Bayesian Active Learning} and \textit{Bayesian Optimization}, where a wrong representation of  uncertainty can lead to a significant performance degradation.
In all experiments, we employ a RBF kernel and we estimate its parameters by maximising the respective marginal likelihood  for  LP, EP and SkewGP.
The Python implementation of SkewGPs for regression, classification and mixed problems is available at \url{https://github.com/benavoli/SkewGP}.
Although the derivations in the previous sections were carried out for a generic SkewGP, in the numerical  experiments below we will consider as a prior a SkewGP with latent dimension $s=0$, that is a GP prior. 
In this way the only difference between LP, EP and SkewGP is the computation  of the posterior. For all the models (LP, EP and SkewGP) we compute the acquisition functions via Monte Carlo sampling from the posterior (using 3000 samples).\footnote{Although for LP and EP some analytical formulas are available (for instance to compute the Upper Credible Bound used in the Bayesian Optimisation section), by using random sampling  for all models we  remove any  advantage of SkewGP over LP and EP due to the additional exploration effect of Monte Carlo sampling.}

\subsection{Bayesian Active Learning}
A key problem in machine learning and statistics is data efficiency.  Active learning   is a powerful technique for achieving data efficiency. Instead of  collecting and labelling a large dataset, which often can be very costly, in active learning one sequentially acquires labels from an expert only for the most informative data points from a pool of available unlabelled data. 
After each acquisition step, the newly labelled point is  added to the training set, and the model is retrained. This process is repeated until a suitable level of accuracy is achieved.   
As for optimal experimental design, the goal of active learning is to produce the best model with the least possible data. In a nutshell, Bayesian Active Learning consists of four  steps. 
\begin{enumerate}
 \item Train a Bayesian model on the labeled training dataset.
 \item Use the trained model to select the next input from the unlabeled data pool.
 \item  Send the selected input to be labeled (by human experts).
 \item     Add the labeled samples to the training dataset and repeat the steps.
\end{enumerate}
In step 2, the next input point is usually selected by maximising an information criterion.
In this paper, we  focus on active learning for binary classification and, consider the \textit{Bayesian Active Learning by Disagreement} (BALD) information criterion introduced in \citep{houlsby2011bayesian}:\footnote{It corresponds to the information gain computed in  $y$ space.}
\begin{equation}
 \label{eq:bald}
\text{Bald}({\bf x})=h\left(E_{f \sim p(f|W)}\left(\Phi\left(f({\bf x}\right)\right)\right)-E_{f \sim p(f|W)}\left(h\left(\Phi\left(f({\bf x})\right)\right)\right)
 \end{equation}
with $h(p)=-p \log(p)-(1-p)\log(1-p)$ being the binary entropy function of the probability  $p$. The next point ${\bf x}$ is therefore selected by maximising $\text{Bald}({\bf x})$. 

As Bayesian model we consider a GP classifier (GP prior and probit likelihood) and we compute  the  expectations in \eqref{eq:bald} via Monte Carlo sampling from the posterior distribution of the latent function $f$.

\begin{figure}[htp]
	\centering
	\begin{tabular}{ll}
			\begin{minipage}{6cm}
			\includegraphics[height=4.0cm,trim={0.0cm 0.0cm 0.0cm 0.0cm }, clip]{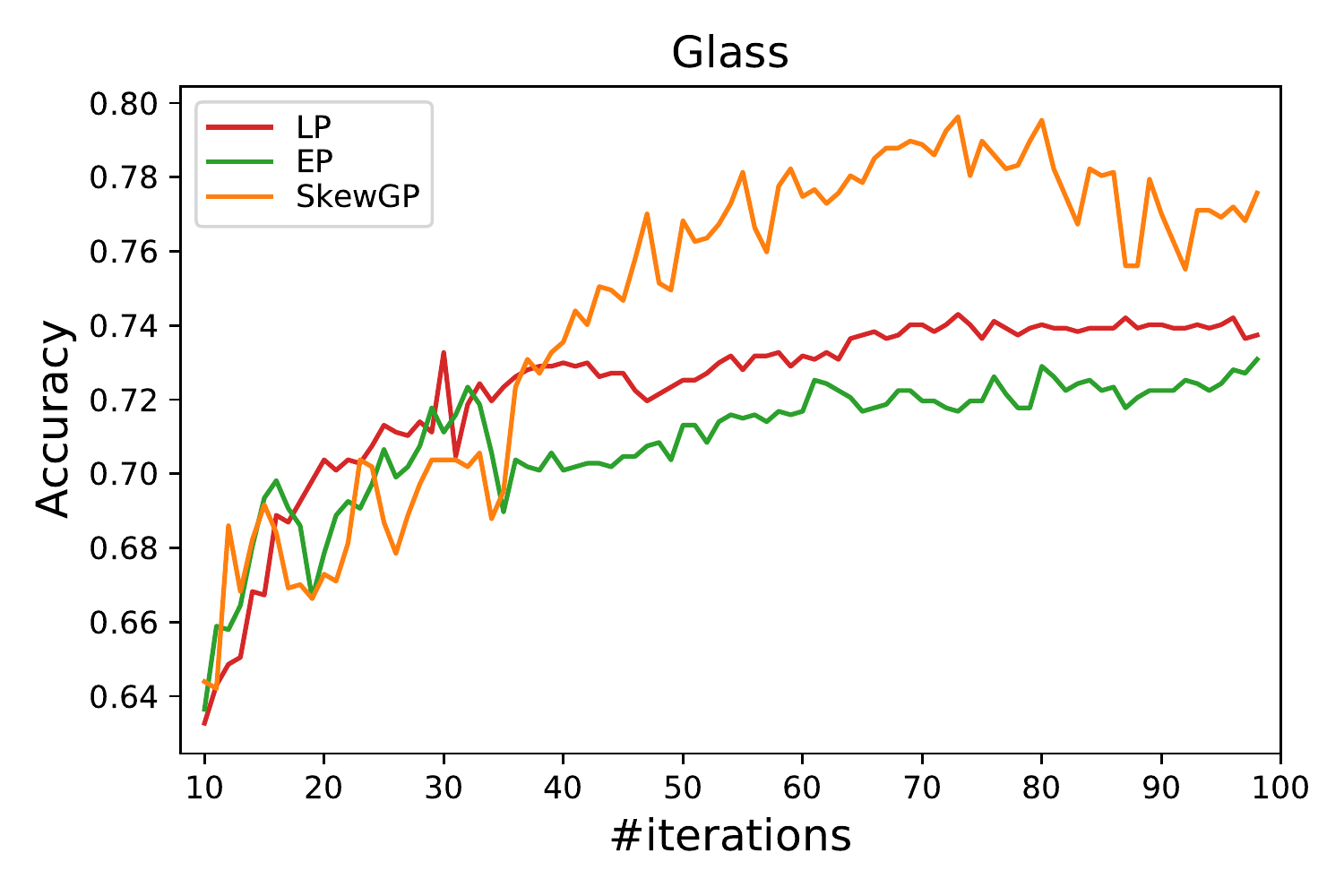}
		\end{minipage}&
		\begin{minipage}{6cm}
			\includegraphics[height=4.0cm,trim={0.0cm 0.0cm 0.0cm 0.0cm }, clip]{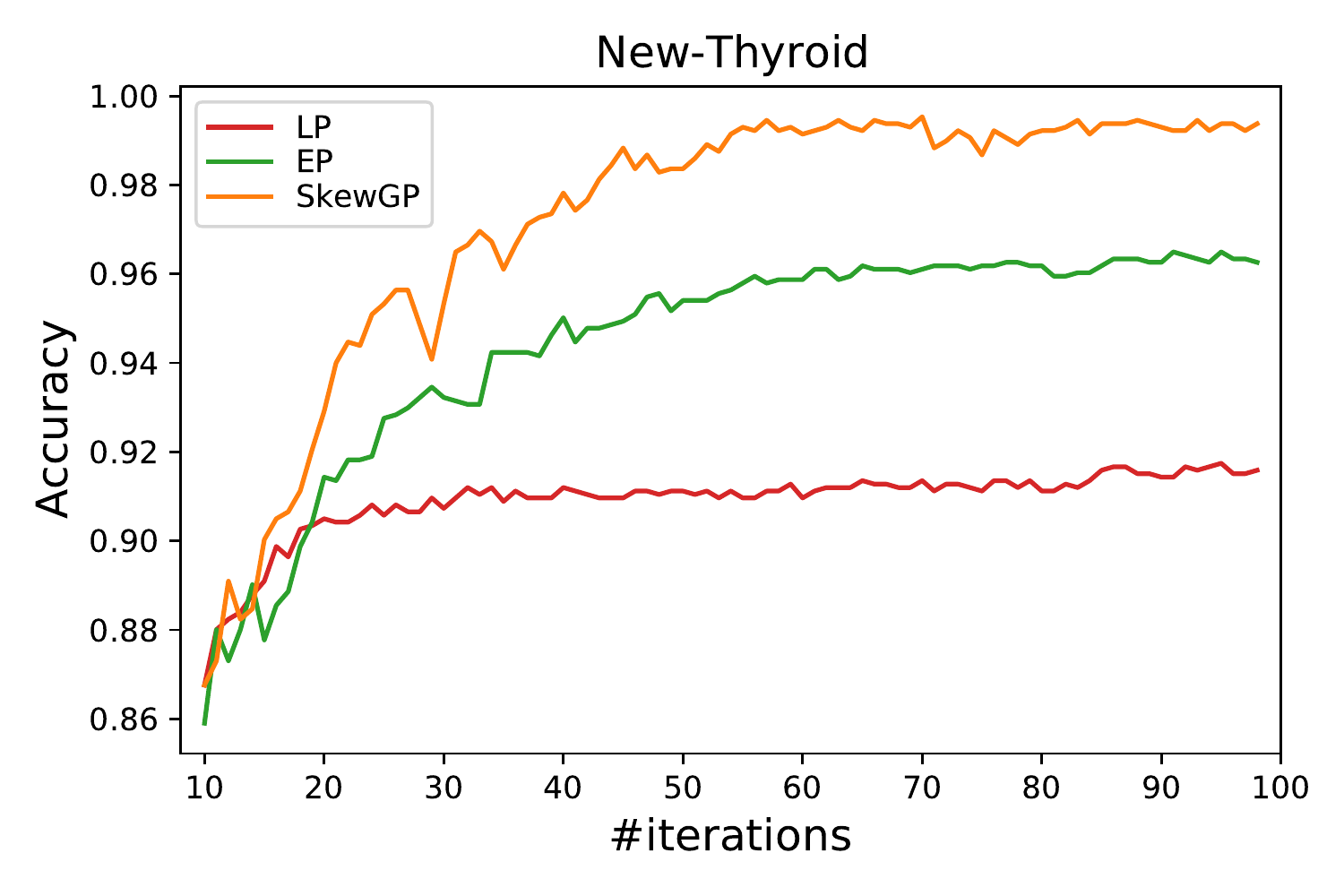}
		\end{minipage}  \\
		\begin{minipage}{6cm}
			\includegraphics[height=4.0cm,trim={0.0cm 0.0cm 0.0cm 0.0cm }, clip]{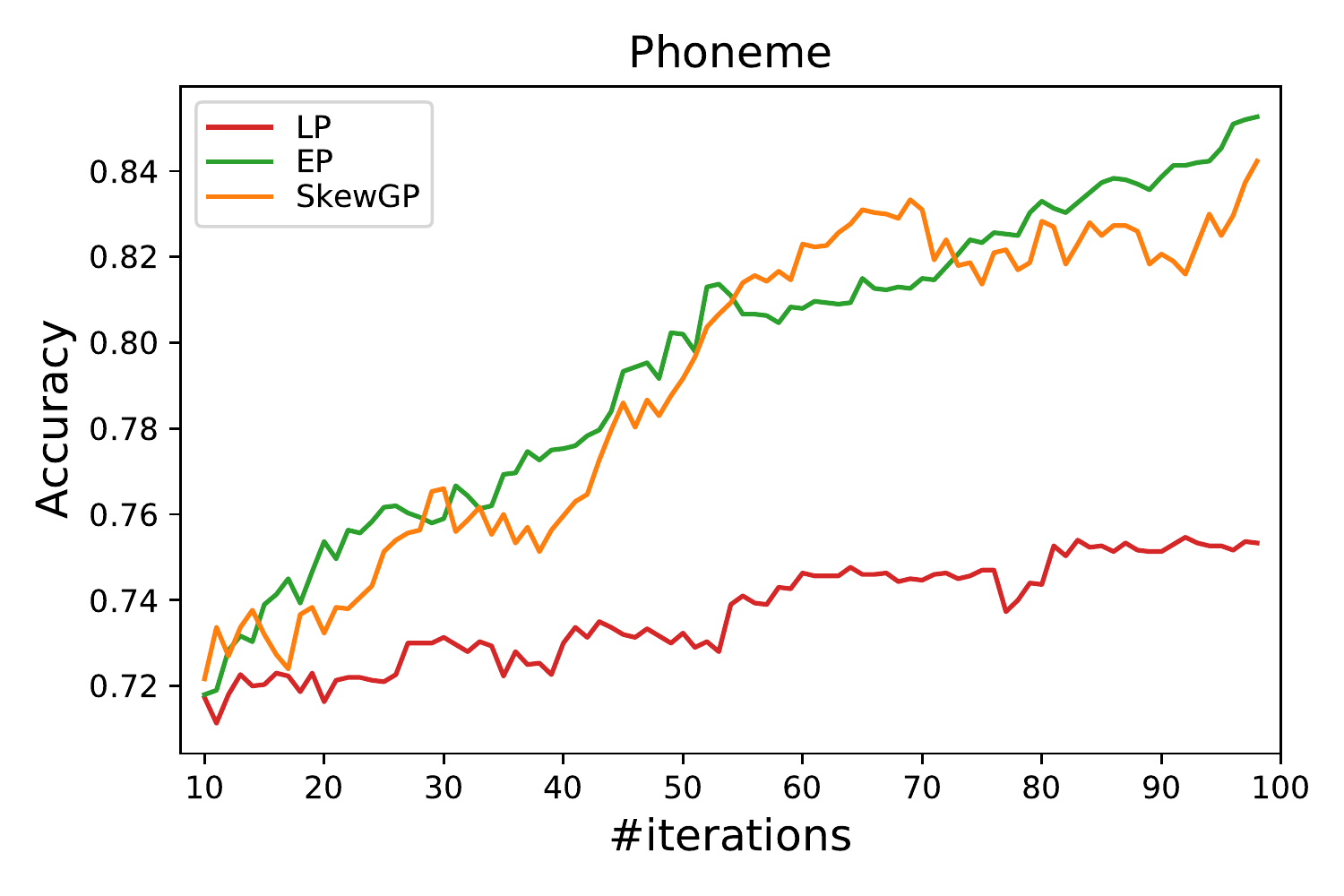}
		\end{minipage} &
		\begin{minipage}{6cm}
			\includegraphics[height=4.0cm,trim={0.0cm 0.0cm 0.0cm 0.0cm }, clip]{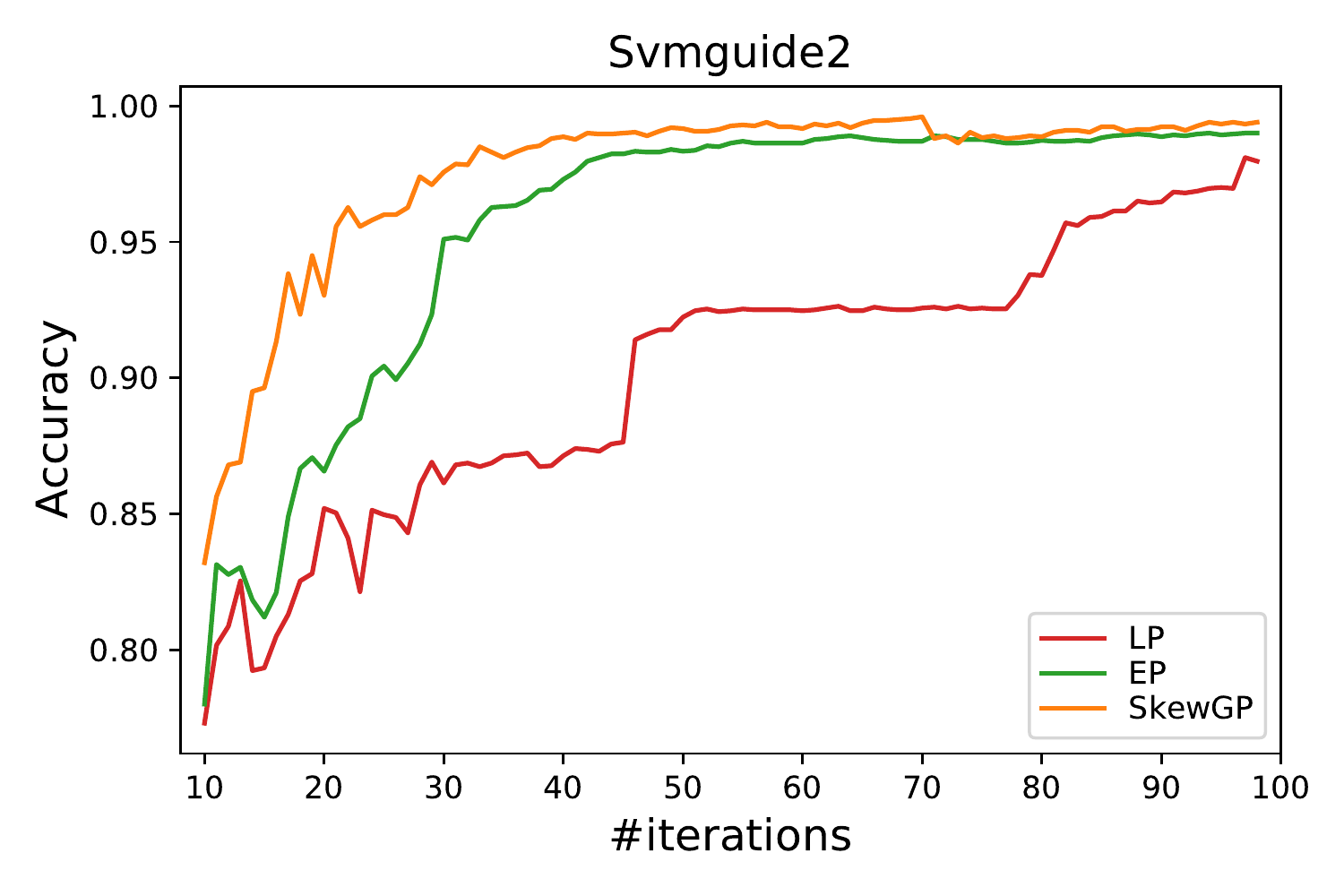}
		\end{minipage}\\
		\begin{minipage}{6cm}
			\includegraphics[height=4.0cm,trim={0.0cm 0.0cm 0.0cm 0.0cm }, clip]{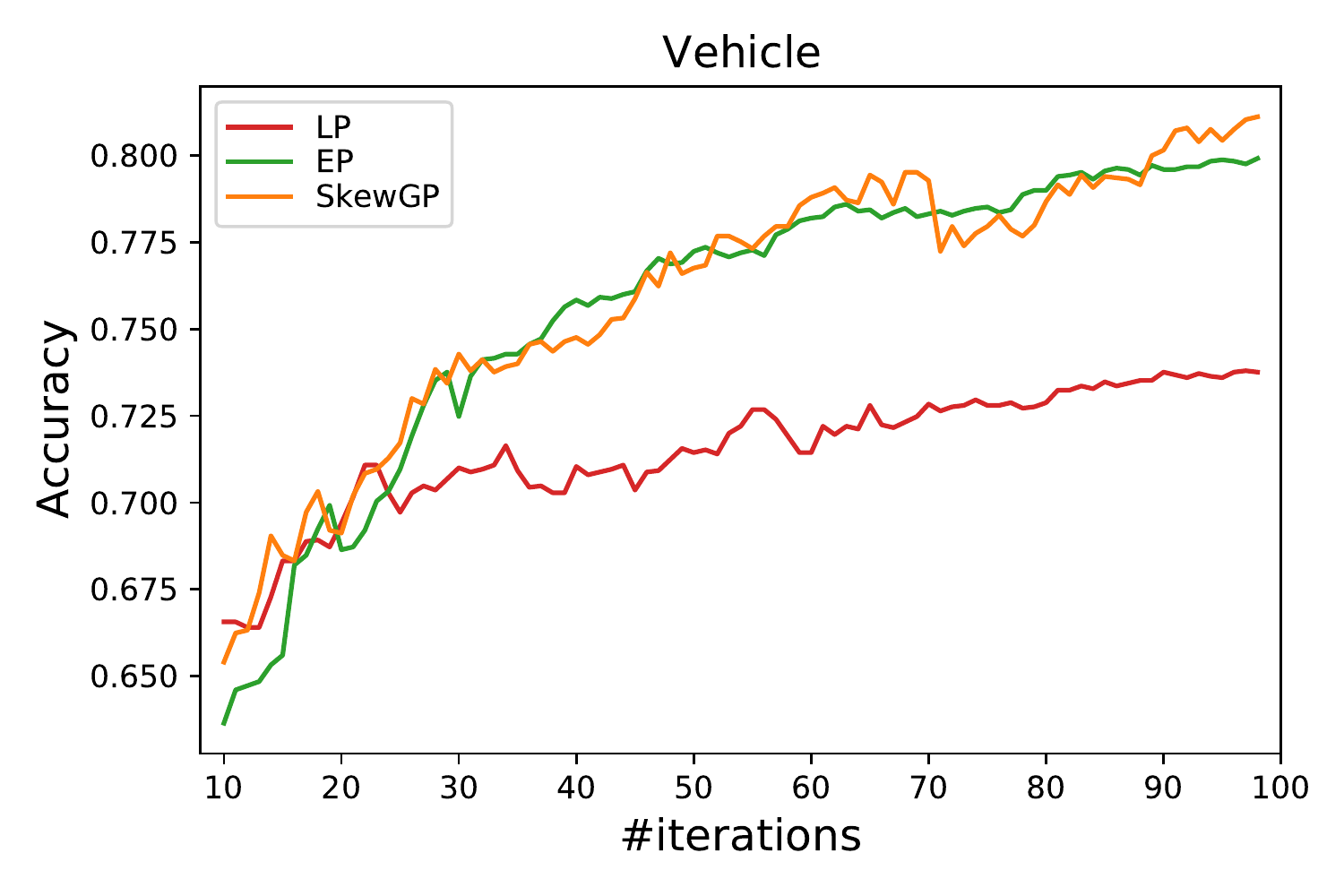}
		\end{minipage}& 
		\begin{minipage}{6cm}
			\includegraphics[height=4.0cm,trim={0.0cm 0.0cm 0.0cm 0.0cm }, clip]{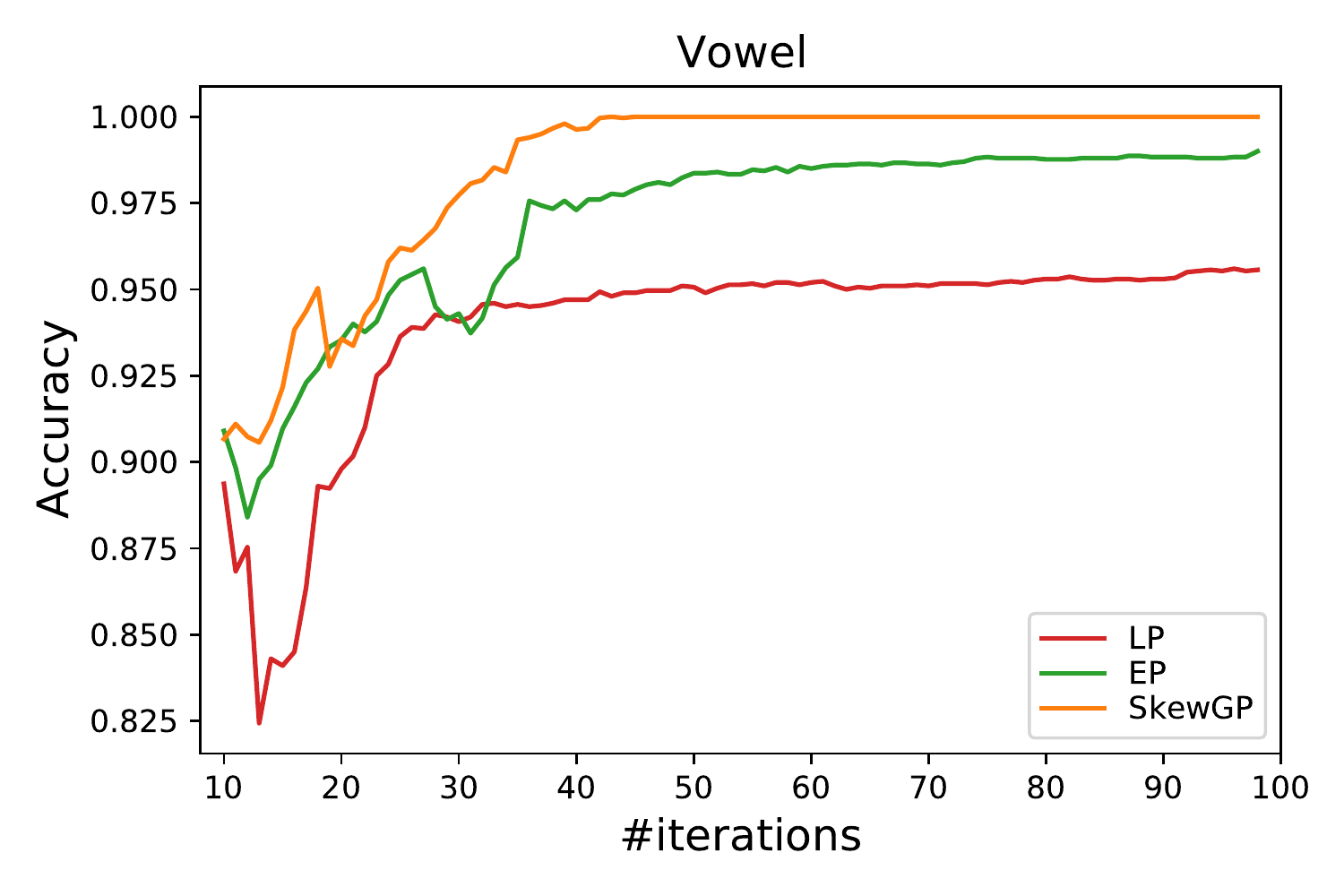}
		\end{minipage}\\
		
				\begin{minipage}{6cm}
			\includegraphics[height=4.0cm,trim={0.0cm 0.0cm 0.0cm 0.0cm }, clip]{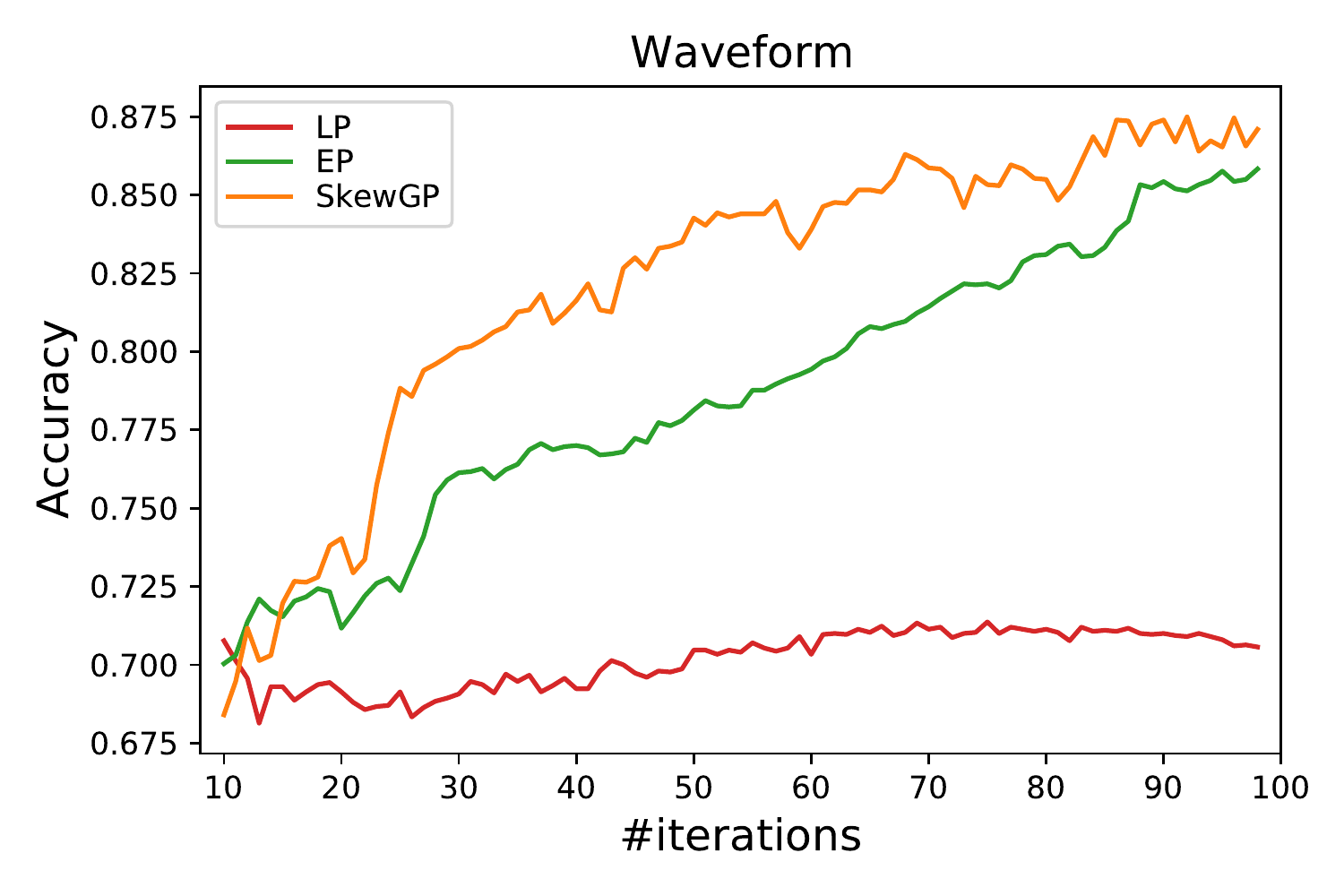}
		\end{minipage}& 
		\begin{minipage}{6cm}
			\includegraphics[height=4.0cm,trim={0.0cm 0.0cm 0.0cm 0.0cm }, clip]{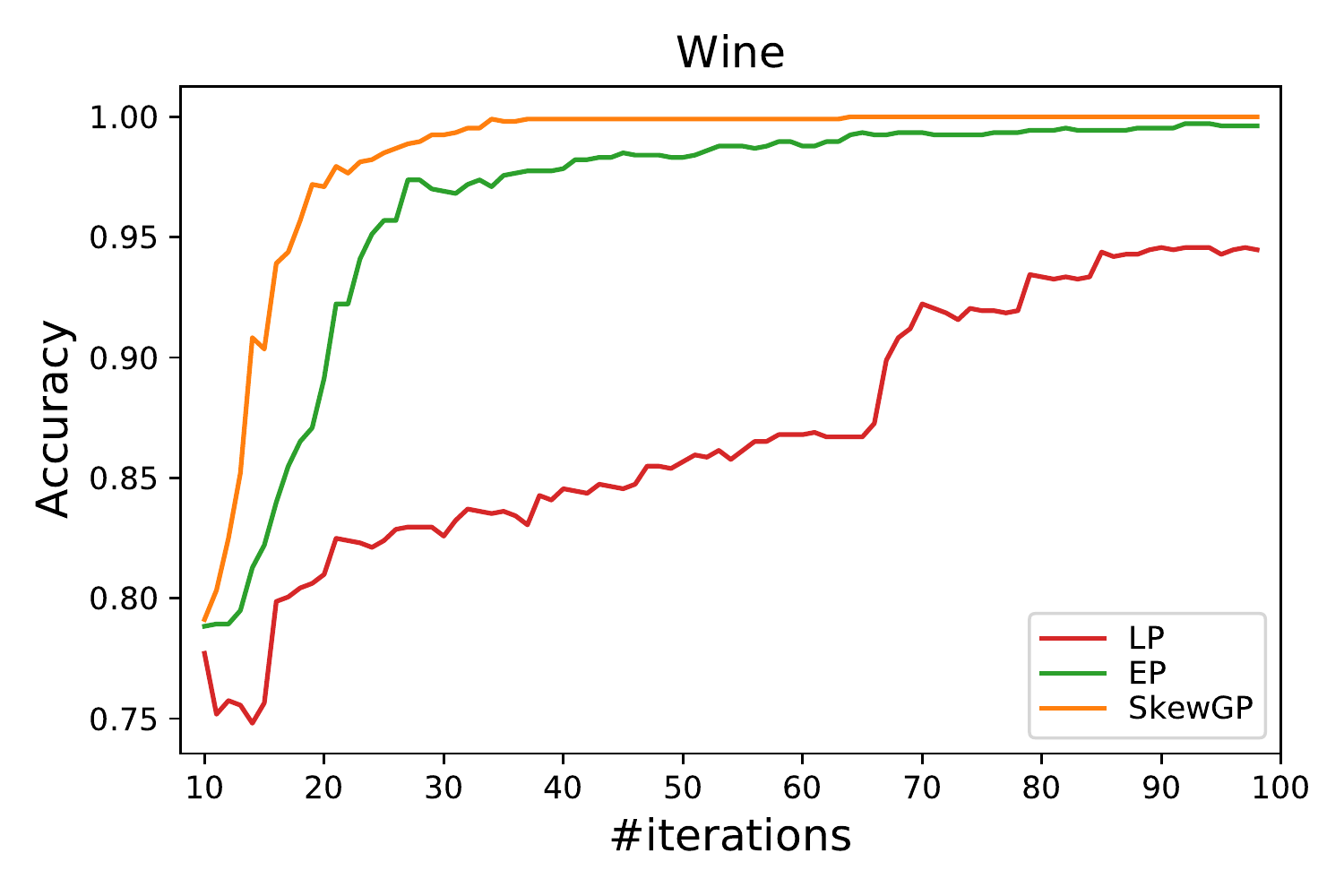}
		\end{minipage}\\
	\end{tabular}
	\caption{Averaged results over 10 trials for LP, EP and SkewGP on the 8 UCI datasets.  The x-axis represents the
		number of iterations and the y-axis represents the classification accuracy.}
	\label{fig:al}
\end{figure}

Our aim is to compare the two approximations for the posterior LP, EP versus the full model SkewGP in the above active learning task.   
We consider eight UCI classification datasets.\footnote{These datasets have recently been used for  GP classification  in \citep{villacampa2020multi}.} Table \ref{tab:charac} displays the characteristics of the considered datasets. 
\begin{table}[H]
		\begin{center}
			{\small
				\begin{tabular}{lccc}
					\hline
					{\bf Dataset} & {\bf \#Instances} & {\bf \#Attributes} & {\bf \#Classes} \\
					\hline
					Glass & 214 & 9 & 6 \\
					New-thyroid & 215 & 5 & 3 \\
					Satellite & 6435 & 36 & 6 \\
					Svmguide2 & 391 & 20 & 3 \\
					Vehicle & 846 & 18 & 4 \\
					Vowel & 540 & 10 & 6 \\
					Waveform & 5000 & 21 & 3 \\
					Wine & 178 & 13 & 3 \\
					\hline
				\end{tabular}
			}
		\end{center}
		\caption{Characteristics of the datasets.}
		\label{tab:charac}
	\end{table}
They are multiclass dataset which we transformed into binary classification problems considering the first class (class $0$) versus rest (class $1$).
We start with a randomly selected initial labelled pool of 10 data points and we run active learning sequentially for 90 steps. We perform 10 trials by starting from a different randomly selected initial pool. Figure \ref{fig:al} shows, for each dataset, the  accuracy (averaged over the 10 trials) as a function of the number of iterations for LP, EP and SkewGP.  

\begin{figure}[htp]
	\centering
		\includegraphics[height=3.3cm,trim={0.0cm 0.0cm 0.0cm 0.0cm }, clip]{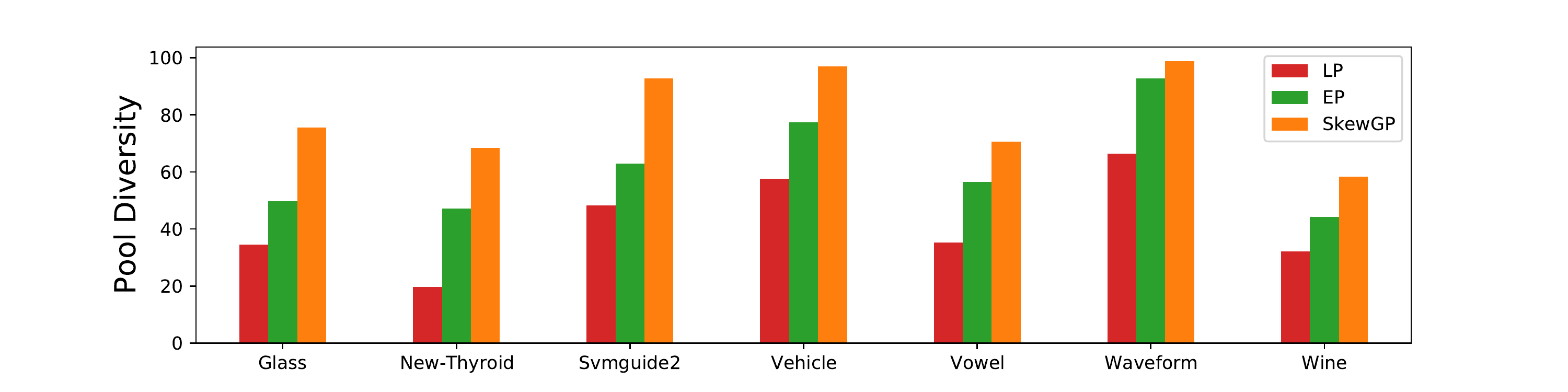}
	\caption{Averaged pool diversity (percentage) in the active learning task.}
	\label{fig:al1}
\end{figure}

LP performs poorly, in many cases its accuracy does not increase with the number of iterations. SkewGP is always the best algorithm and  clearly outperforms EP in $6$ datasets.
The reason of this significant difference in performance  can be understood from  Figure \ref{fig:al1} which reports, for each dataset, 
the percentage (averaged over 10 trials) of unique input instances in the final labelled pool (that is at the end of the 90 iterations). We call it  \textit{pool diversity}. A low pool diversity means that the same instance has been included in the pool more than once, which is the result of a poor representation of uncertainty.
Note in fact that LP has always the lowest  pool diversity. EP, which provides a better approximation of the posterior, performs better, but SkewGP has always the highest pool diversity.
SkewGP performs a better exploration of the input space.

\subsection{Bayesian Optimisation}
We consider the problem of finding the global maximum of an unknown function which is expensive to evaluate. For instance,  evaluating the function requires conducting an experiment.
Mathematically, for a function $g$ on a domain $X$, the goal is to find a global maximiser ${\bf x}^o$:
$$
{\bf x}^o =\arg \max_{{\bf x} \in X} g({\bf x}).
$$
Bayesian  optimisation (BO) poses this  as  a   sequential  decision  problem -- a trade-off between learning about the underlying function $g$  (exploration) and capitalizing on this information in order to find the optimum ${\bf x}^o$ (exploitation).

In BO, it is usually assumed that $g({\bf x})$ can be evaluated  directly (numeric (noisy) observations). However, in many applications,  measuring $g({\bf x})$ can be either costly or not always possible. In these cases, the  objective function $g$ may only be accessed via preference judgments, such as ``this is better than that'' between two candidate solutions ${\bf x}_i,{\bf x}_j$ (like in A/B tests or recommender systems).  In such situations,  Preferential Bayesian optimization (PBO) \citep{shahriari2015taking,gonzalez2017preferential} or more general algorithms for active preference learning should be adopted~\citep{BDG08,pmlr-v32-zoghi14,sadigh2017active,bemporad2019active}. These approaches require the users to  simply  compare the final outcomes of two different candidate solutions and indicate which they prefer. 

There are also applications where either
\begin{enumerate}
 \item numeric (noisy) measurements and preference, or
 \item numeric (noisy) measurements and binary observations,
\end{enumerate}
may be available together;

In the next sections, we show how we can use SkewGP as surrogated model for BO in these situations and how it outperforms BO based on LP and EP.

\subsubsection{Preferential Optimisation}
\label{sec:exper_preferece}
The state-of-the-art approach for PBO \citep{shahriari2015taking,gonzalez2017preferential}  uses a GP as a prior distribution of the latent  preference function and a probit  likelihood to model the observed pairwise comparisons. The  posterior distribution of the preference function is approximated via the LP approach. In a recent paper \citep{benavoli2020preferential}, we  showed that, by computing the  exact  posterior (SkewGP), we can outperform LP in PBO.  In this section, we further compare SkewGP with  EP (but we also report LP for completeness). For EP, we use the  formulation for preference learning   discussed by \citep{houlsby2011bayesian}, which shows that GP preference learning is equivalent to GP classification with a particular transformed  kernel function.
 Therefore, we compare three different implementation of Bayesian preferential optimisation based on LP, EP, SkewGP.\footnote{For LP and EP, we use \citep{gpy2014}.}

 In PBO, since $g$ can only be queried via preferences, the next candidate solution ${\bf x}$  is selected by optimizing (w.r.t.\ ${\bf x}$) a \textit{dueling acquisition function} $\alpha({\bf x},{\bf x}_r)$, where ${\bf x}_r$ (reference point) is the best point found so far.\footnote{More precisely, we assume that preferential observation compares  the current input with the best input found so far ${\bf x}_r$.} . By optimizing  $\alpha({\bf x},{\bf x}_r)$, one aims to find a point that is better than ${\bf x}_r$ (but also considering the trade-off between exploration and exploitation). After computing the optimum of the the acquisition function, denoted with ${\bf x}_n$, we query the black-box function for ${\bf x}_n\stackrel{?}{\succ} {\bf x}_r$. If    ${\bf x}_n\succ {\bf x}_r$  then ${\bf x}_n$ becomes the new reference point (${\bf x}_r$) for the next iteration.

We consider two dueling acquisition functions: (i) Upper Credible Bound (UCB); (ii)  Expected Improvement Info Gain (EIIG). 

 {\bf UCB:} The dueling UCB acquisition function is defined as the upper bound of the minimum width $\gamma$\% (in the experiments we use $\gamma=95$) credible interval of $f({\bf x})-f({\bf x}_r)$.

{\bf EIIG:} The dueling EIIG was proposed in \citep{benavoli2020preferential} by combining  the expected  probability of improvement (in log-scale) and the dueling BALD \citep{houlsby2011bayesian} (information gain): 
$$
k \log\left(E_{f \sim p(f|W)}\left(\Phi\left(\tfrac{f({\bf x})-f({\bf x}_r)}{\sqrt{2}\sigma}\right)\right)\right) - IG({\bf x},{\bf x}_r),$$
where
$
IG({\bf x},{\bf x}_r)=h\left(E_{f \sim p(f|W)}\left(\Phi\left(\tfrac{f({\bf x})-f({\bf x}_r)}{\sqrt{2}\sigma}\right)\right)\right)-E_{f \sim p(f|W)}\left(h\left(\Phi\left(\tfrac{f({\bf x})-f({\bf x}_r)}{\sqrt{2}\sigma}\right)\right)\right).
$   
This last acquisition function balances exploration-exploitation by means of the nonnegative scalar $k$ (in the  experiments we use $k=0.1$).

\begin{figure}[htp]
	\centering
	\begin{tabular}{ll}
			\begin{minipage}{6cm}
			\includegraphics[height=2.7cm,trim={0.6cm 0.0cm 0.0cm 0.0cm }, clip]{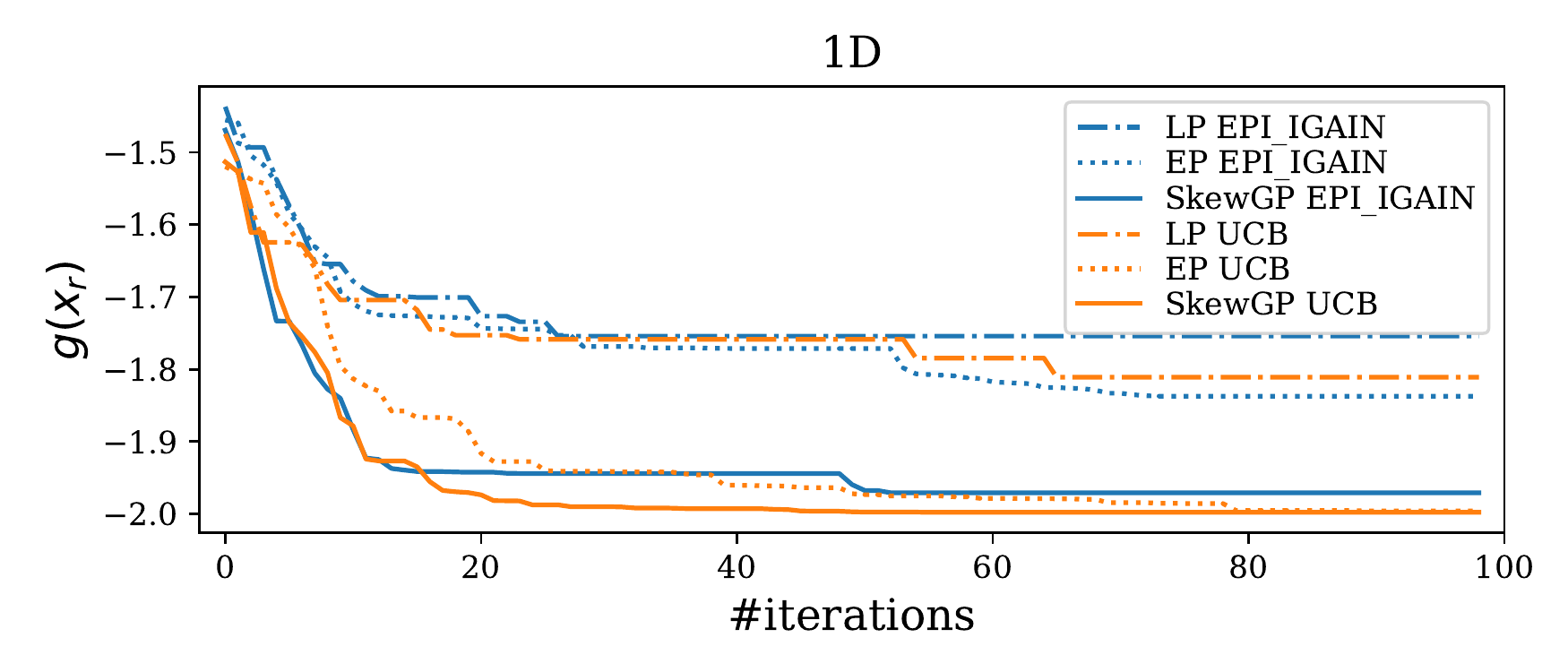}
		\end{minipage}&
		\begin{minipage}{6cm}
			\includegraphics[height=2.7cm,trim={0.6cm 0.0cm 0.0cm 0.0cm }, clip]{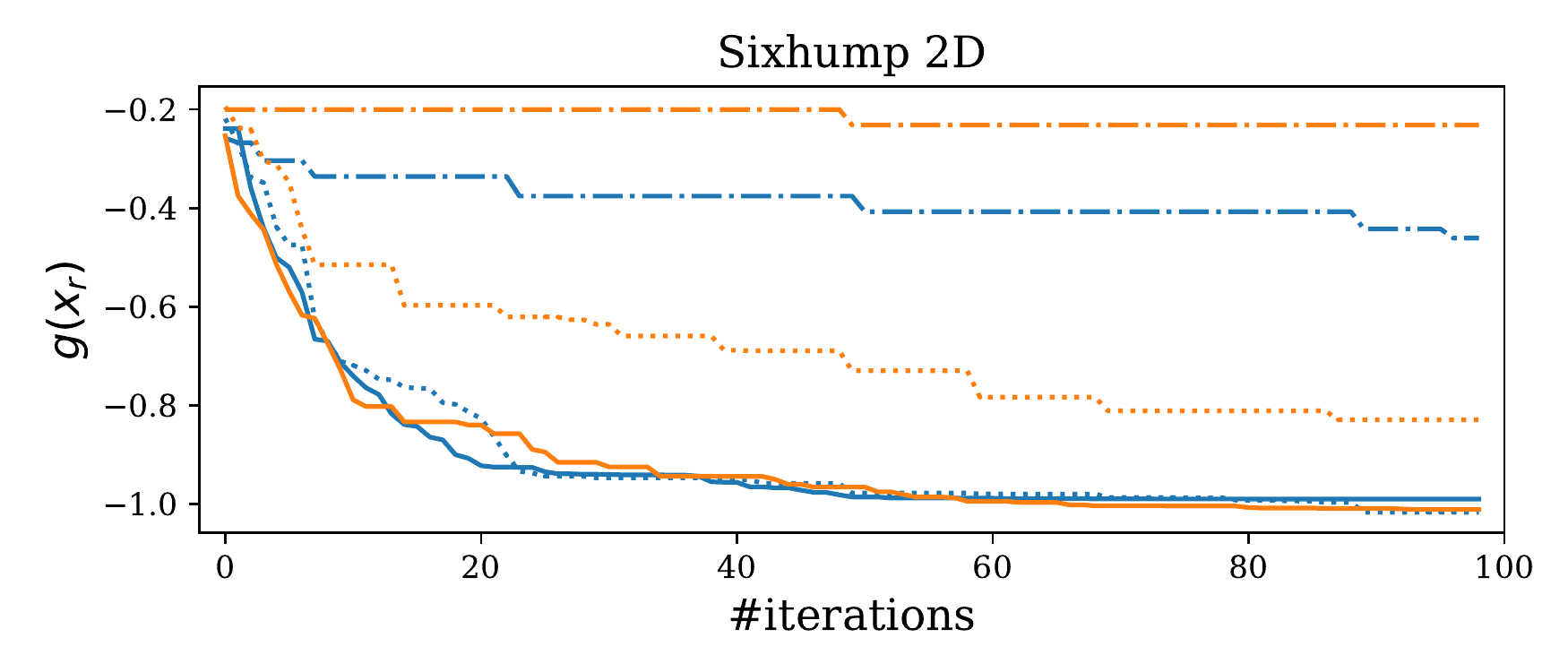}
		\end{minipage}  \\
		\begin{minipage}{6cm}
			\includegraphics[height=2.7cm,trim={0.4cm 0.0cm 0.0cm 0.0cm }, clip]{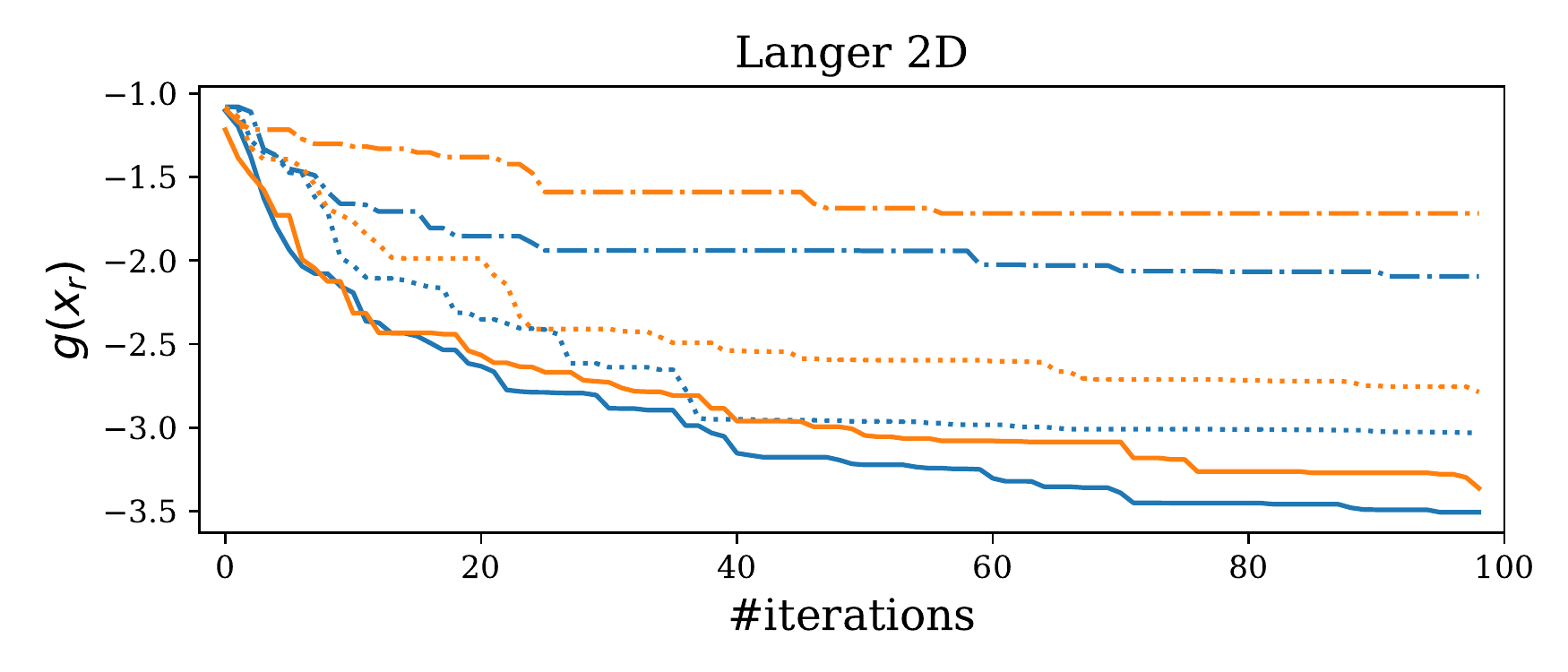}
		\end{minipage} &
		\begin{minipage}{6cm}
			\includegraphics[height=2.7cm,trim={0.4cm 0.0cm 0.0cm 0.0cm }, clip]{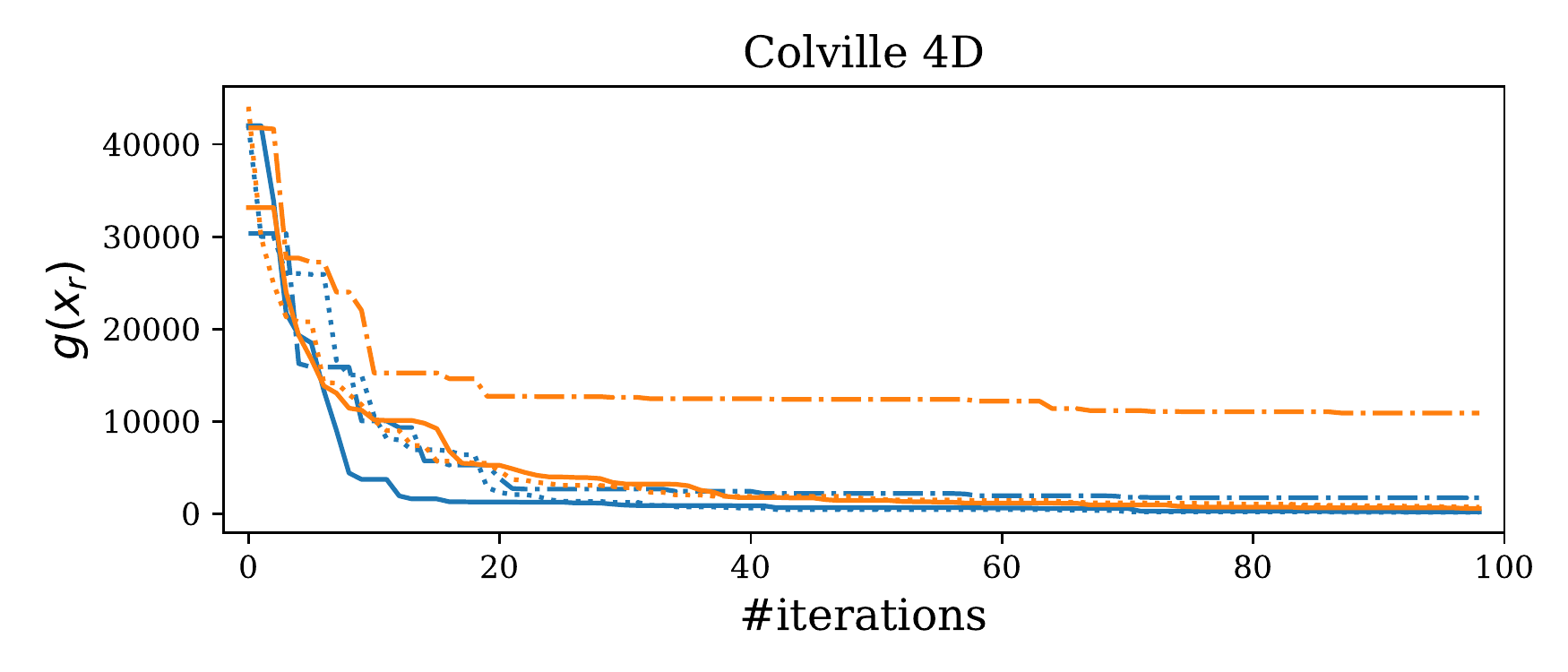}
		\end{minipage}\\
		\begin{minipage}{6cm}
			\includegraphics[height=2.7cm,trim={0.4cm 0.0cm 0.0cm 0.0cm }, clip]{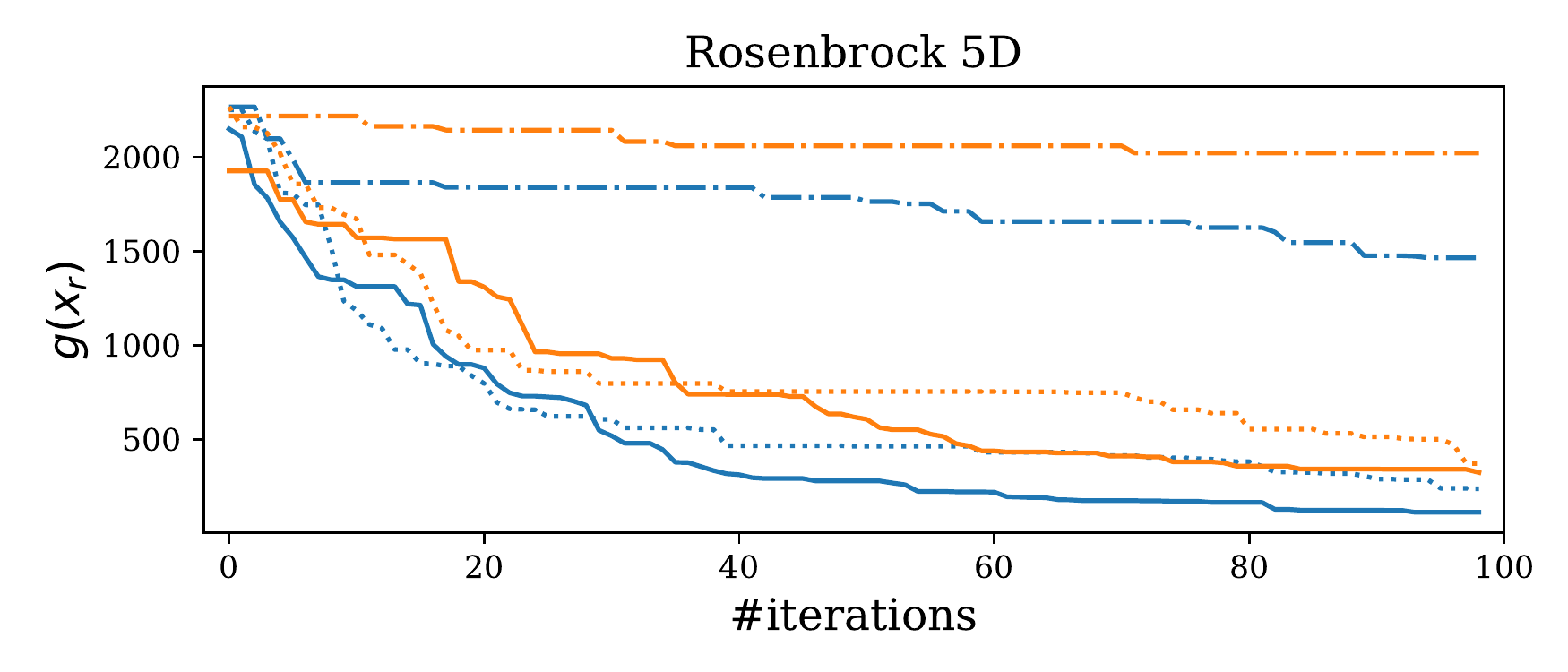}
		\end{minipage}& 
		\begin{minipage}{6cm}
			\includegraphics[height=2.7cm,trim={0.4cm 0.0cm 0.0cm 0.0cm }, clip]{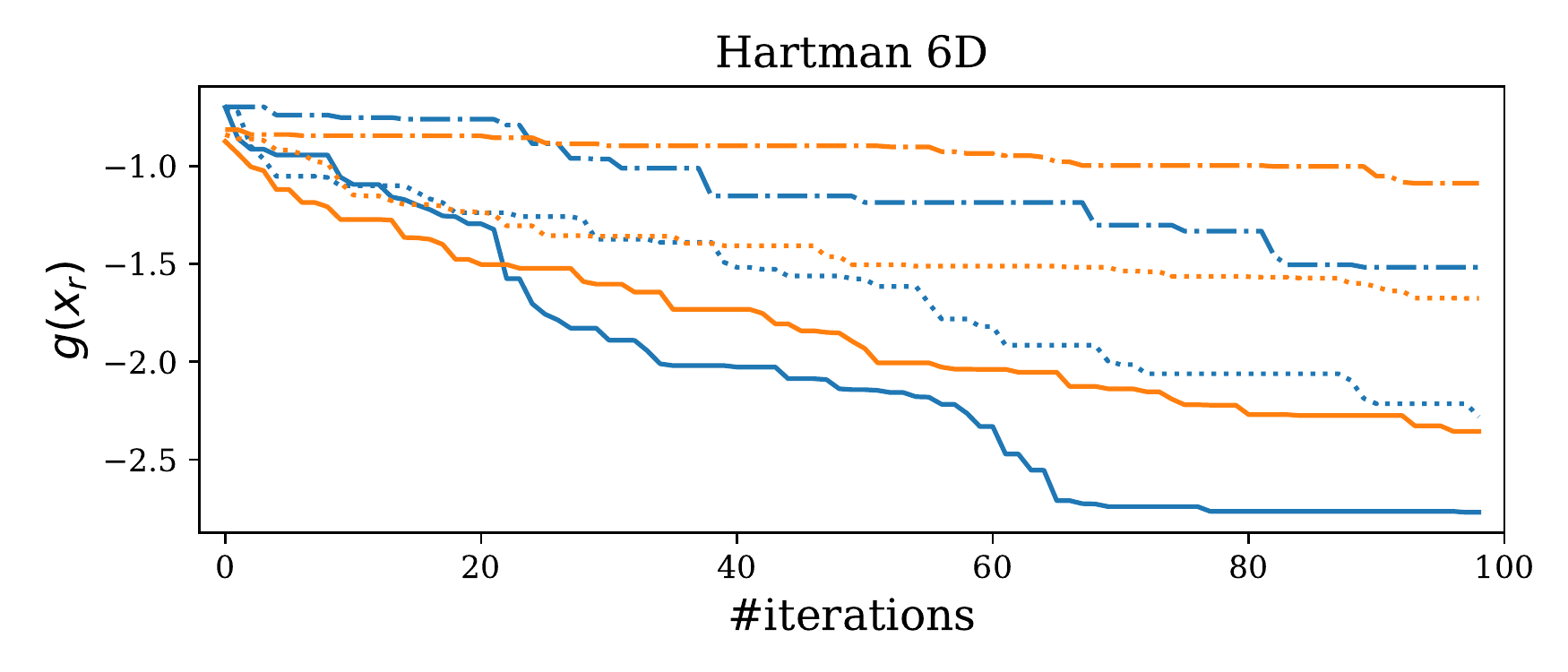}
		\end{minipage}\\
	\end{tabular}
	\caption{Averaged results over 20 trials for PBO for LP, EP and SkewGP on the 6 benchmark functions considering 2 different acquisition functions.  The x-axis represents the
		number of iterations and the y-axis represents the value  of the true objective function at the current optimum ${\bf x}_r$.}
	\label{fig:4}
\end{figure}
We have considered for $g({\bf x})$ six benchmark functions: $g(x)=cos(5x)+e^{-\frac{x^2}{2}}$  denoted as `1D' (1D),  `Six-Hump
Camel' (2D), `Langer' (2D), `Colville' (4D),
`Rosenbrock5' (5D) and `Hartman6' (6D). These are minimization problems.\footnote{We converted them into maximizations
	so that the acquisition functions are well-defined.}
	Each experiment begins with $5$ initial (randomly selected) duels and a total budget of $100$ duels are run.
Further, each experiment is repeated 20 times  with different initialization (the same for all methods) as in \citep{gonzalez2017preferential,benavoli2020preferential}.

Figure \ref{fig:4} reports the performance of the 
different methods.
Consistently across all benchmarks 
PBO-SkewGP outperforms both PBO-LP and PBO-EP (no matter the acquisition function.)
This  confirms for PBO what previously noticed for active learning: a wrong uncertainty representation leads to a non optimal exploration of the input space and, therefore, to a slower convergence in active learning tasks. 

\subsubsection{Mixed numerical and preferential BO}
We repeat the previous experiments considering  mixed type observations, that is $g$ can be queried  directly (numeric data) and via preferences.
Each experiment begins with $5$ initial (randomly selected) duels (preferences) and 
$5$ numeric (scalar) observations of $g$. 
A total budget of $100$ iterations is considered and we assume that at the iteration $4,8,12,16,\dots$ $g$ is queried directly and in all the other iterations is queried via preferences (with respect to $x_r$). Each experiment is repeated 20 times  with different initialization (the same for all methods).
In this case, we only consider the UCB acquisition function, which is valid for both numeric and preference data.\footnote{Observe in fact that the Bald criterion assumes binary observations.}

\begin{figure}
	\centering
	\begin{tabular}{ll}
			\begin{minipage}{6cm}
			\includegraphics[height=2.7cm,trim={0.6cm 0.0cm 0.0cm 0.0cm }, clip]{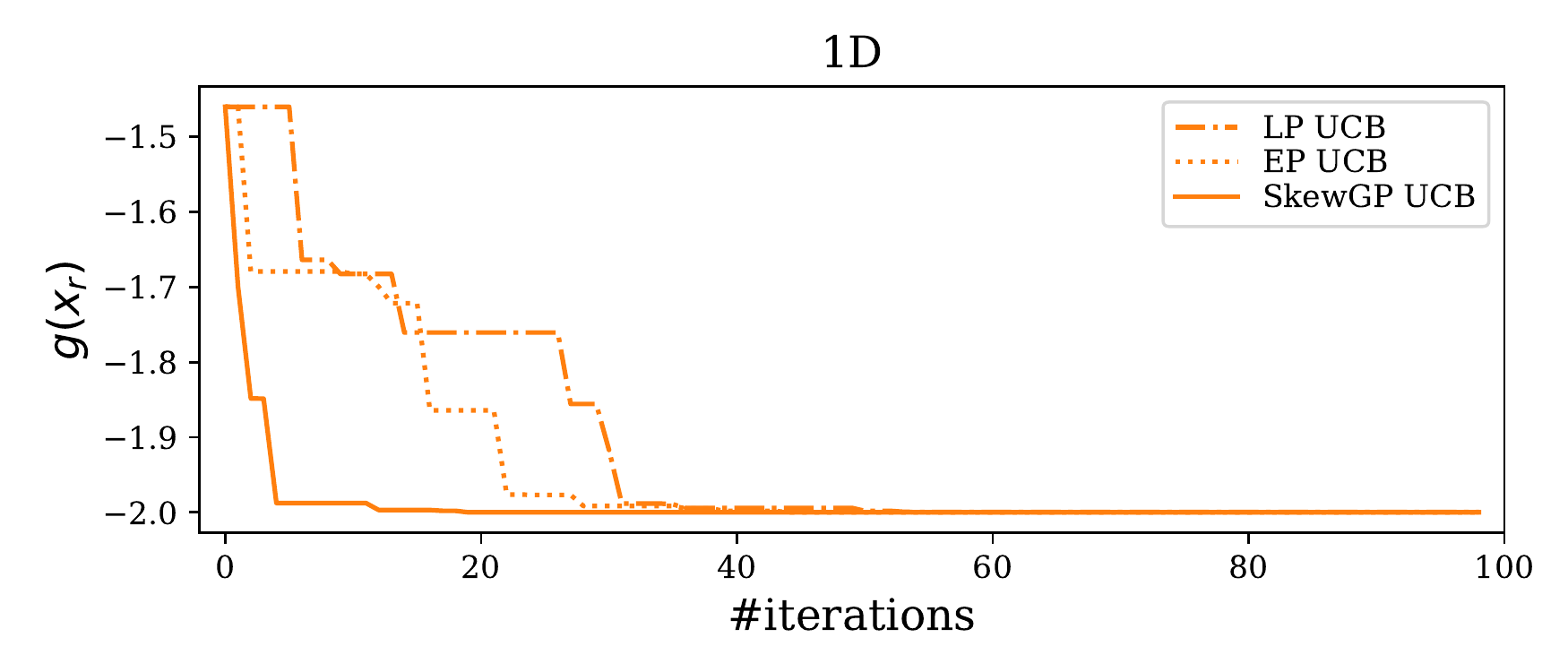}
		\end{minipage}&
		\begin{minipage}{6cm}
			\includegraphics[height=2.7cm,trim={0.6cm 0.0cm 0.0cm 0.0cm }, clip]{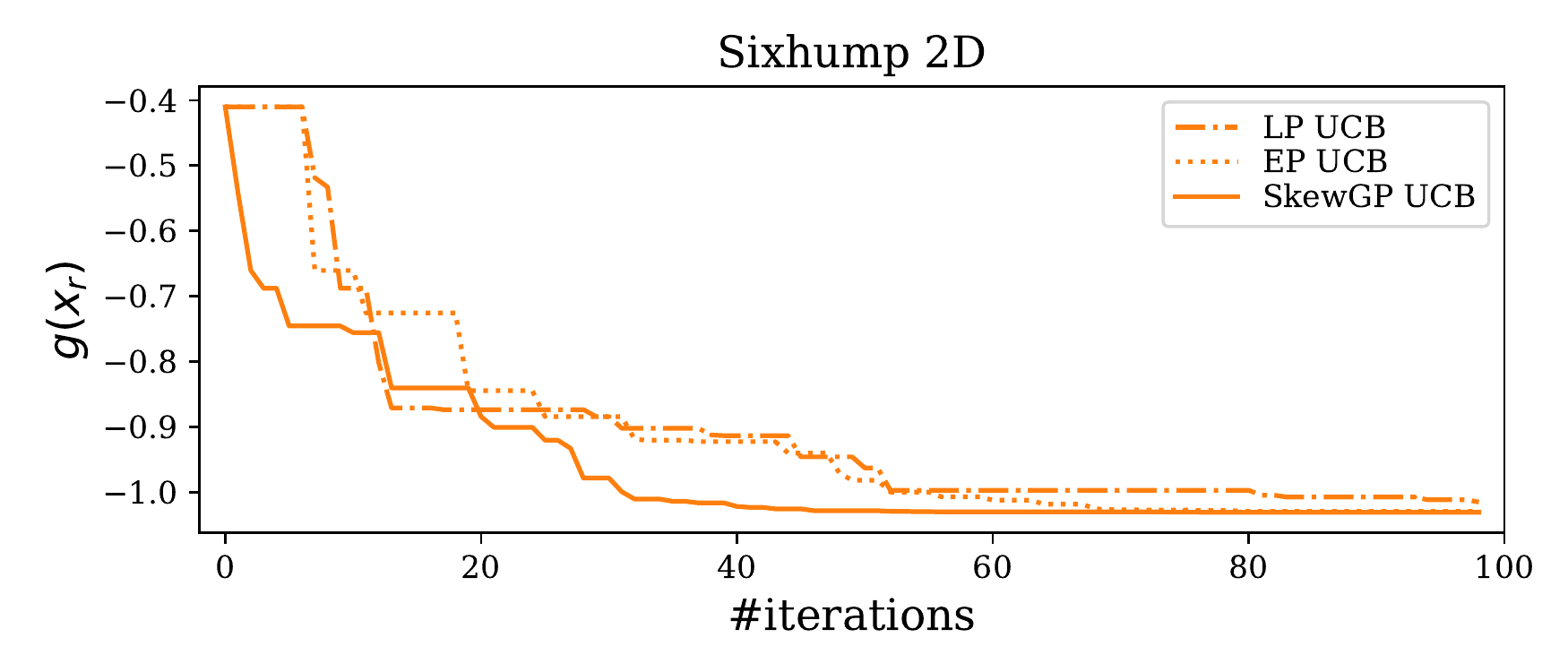}
		\end{minipage}  \\
		\begin{minipage}{6cm}
			\includegraphics[height=2.7cm,trim={0.6cm 0.0cm 0.0cm 0.0cm }, clip]{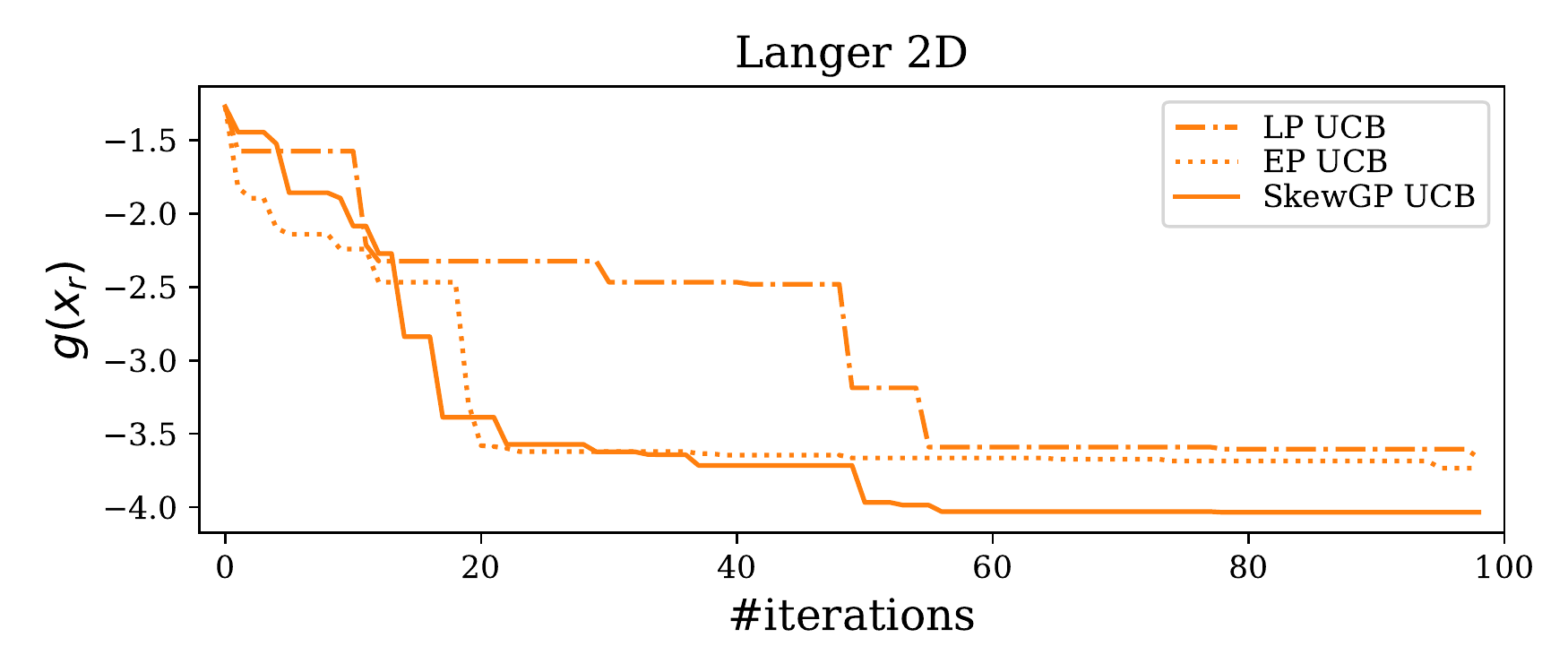}
		\end{minipage} &
		\begin{minipage}{6cm}
			\includegraphics[height=2.7cm,trim={0.6cm 0.0cm 0.0cm 0.0cm }, clip]{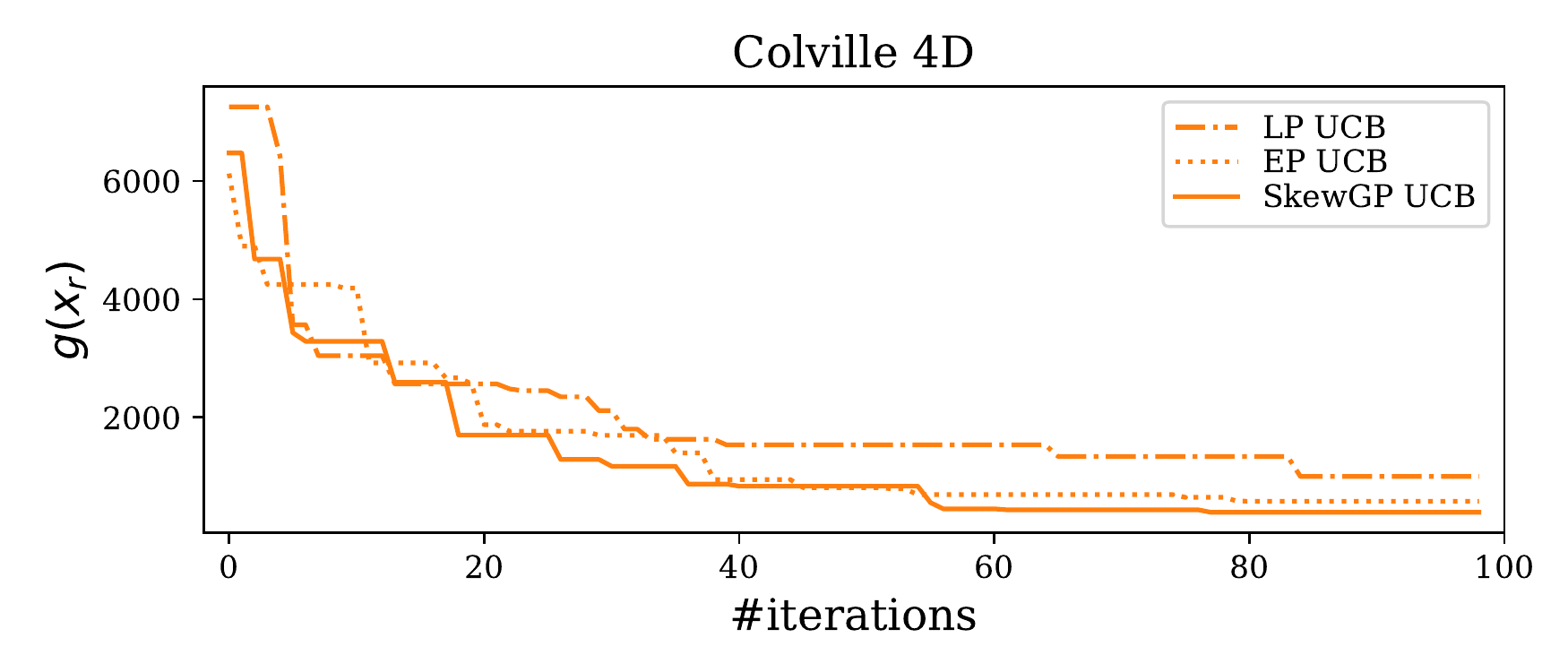}
		\end{minipage}\\
		\begin{minipage}{6cm}
			\includegraphics[height=2.7cm,trim={0.6cm 0.0cm 0.0cm 0.0cm }, clip]{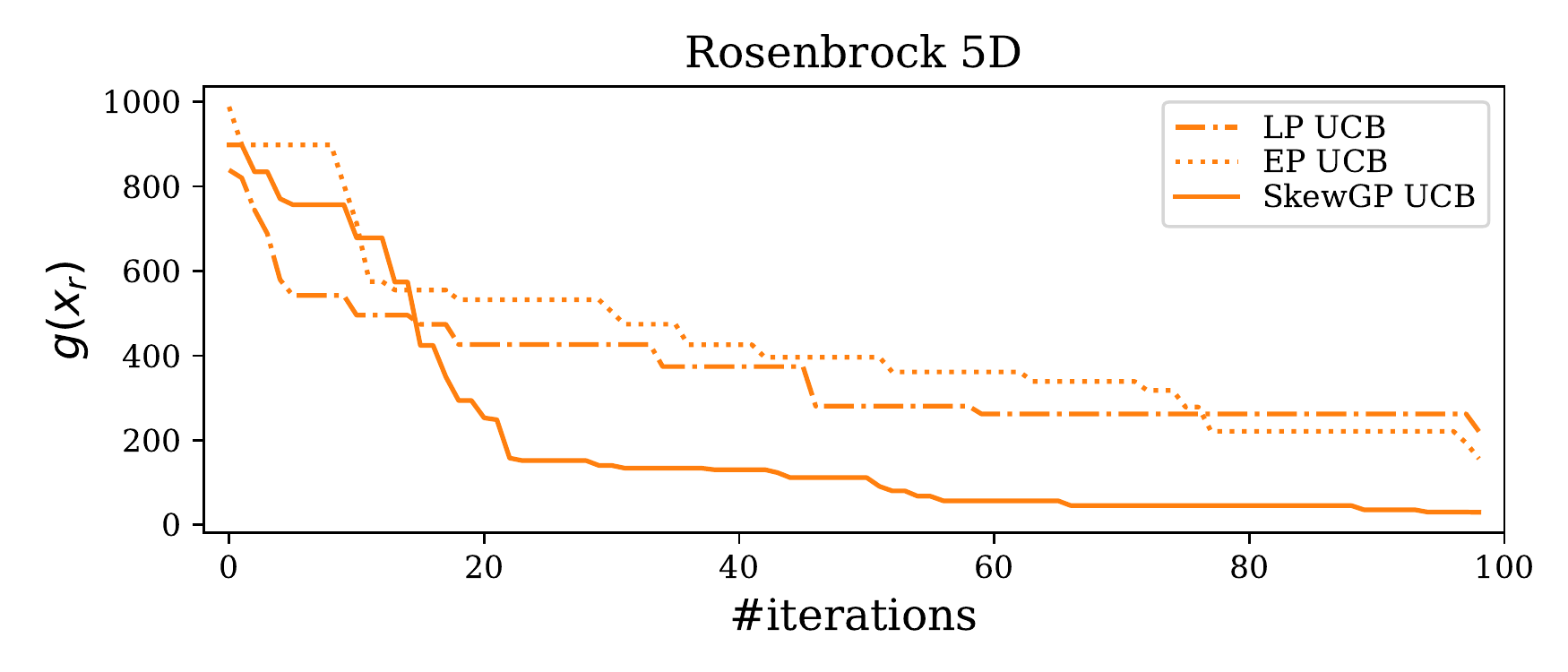}
		\end{minipage}& 
		\begin{minipage}{6cm}
			\includegraphics[height=2.7cm,trim={0.6cm 0.0cm 0.0cm 0.0cm }, clip]{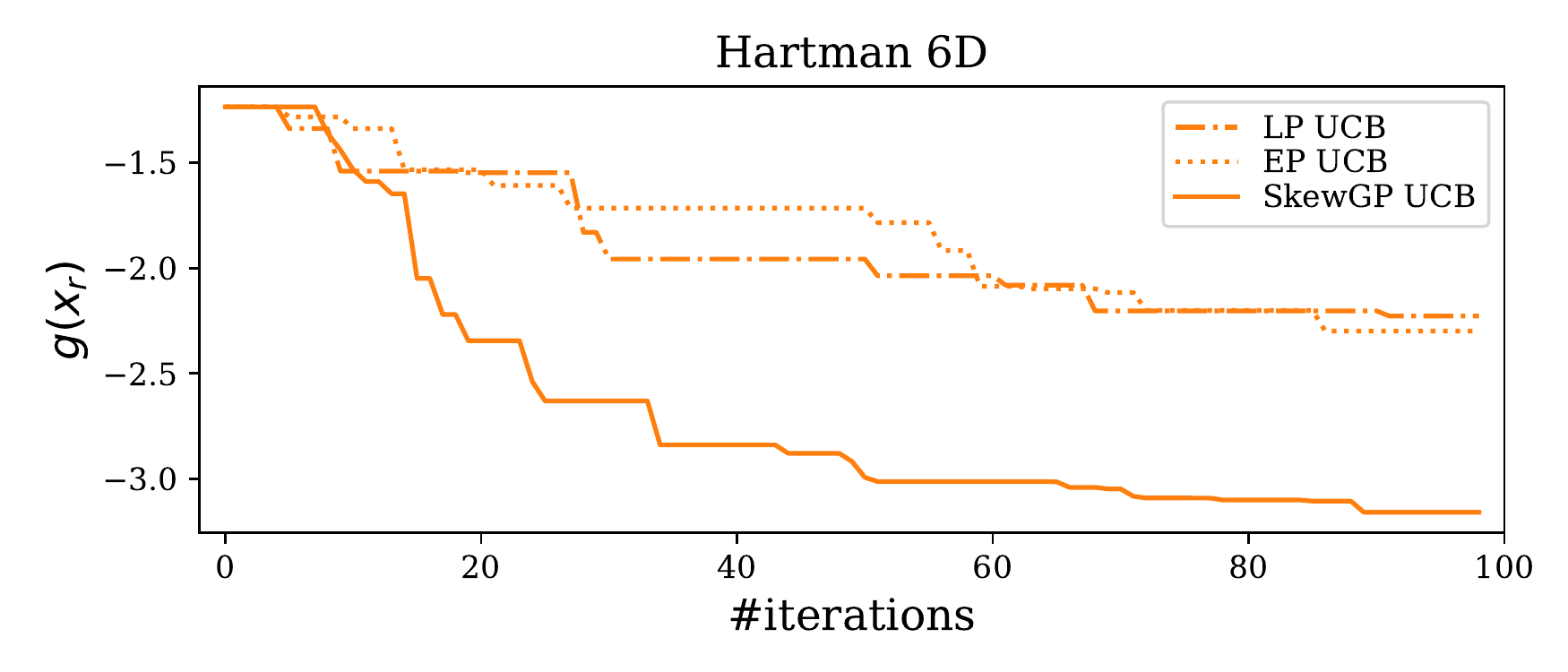}
		\end{minipage}\\
	\end{tabular}
	\caption{Averaged results over 20 trials for mixed-PBO for LP, EP and SkewGP on the 6 benchmark functions considering the UCB acquisition function.  The x-axis represents the
		number of iterations and the y-axis represents the value  of the true objective function at the current optimum ${\bf x}_r$.}
	\label{fig:5}
\end{figure}

Figure \ref{fig:5} shows the performance of the 
different methods. As expected,\footnote{LP, EP and SkewGP coincides in case the observations are all numeric.} the differences between the three approaches tend to be smaller than preference-only BO, especially in the low dimensional problems. However,  also in this case, BO-SkewGP outperforms both BO-LP and BO-EP.

\subsubsection{Safe Bayesian Optimisation}
Safe BO \citep{sui2015safe} is an extension of BO, which aims
to solve the constrained optimization problem:
$$
\begin{aligned}
{\bf x}^o =\arg \max_{{\bf x} \in X} g({\bf x}) ~~\text{s.t.}~~~g({\bf x})\geq h,
\end{aligned}
$$
where $g$ is an unknown function and $h \in \mathbb{R}$ is a safety threshold. The set
$\mathscr{S}=\{{\bf x}  \in X \mid  g({\bf x})\geq h\}$ is called the \textit{safe} set. This safe set is not known initially (because $g$ is unknown), but is estimated after each function evaluation.
Note that, in safe optimisation,  the  algorithm  should avoid (with high probability) unsafe points.
Therefore, the challenge is to find an appropriate strategy, which at each iteration not only aims to find the global maximum within the currently known safe set $\hat{\mathscr{S}}$ (exploitation), but also aims to increase the size of $\hat{\mathscr{S}}$ including inputs ${\bf x}$ that are known to be safe (exploration). Different strategies and approaches are discussed in \citep{sui2015safe}.

In Safe BO, it is usually assumed that $g$  can be queried directly. Given a certain input ${\bf x}_k$, we instead assume  that when $g({\bf x}_k)  < h$   no output is produced. As discussed for \eqref{eq:likevalid}, in this case   the space of possibility is $\{(\text{valid},g({\bf x}_k)),~~(\text{non-valid},None)\}$ -- an input ${\bf x} $ is  \textit{valid} whether ${\bf x} \in \mathscr{S}$.

We can address this BO problem using the general framework \citep{berkenkamp2016safe,berkenkamp2016bayesian}, that is we employ a GP regression model to learn $g$ using only \textit{valid} data and a GP   classifier (with two classes \textit{valid} and \textit{non-valid}) to learn the sign of the constraint.\footnote{We have used the SafeOpt library \citep{berkenkamp2016safe,berkenkamp2016bayesian} with $\beta=0.85$ and used the EP approximation for GP classification.}

We compare the above approach with a BO strategy that uses SkewGP as conjugate prior for the likelihood \eqref{eq:likevalid} and, therefore, addresses the two types of observations as a mixed numeric and binary regression problem. As acquisition function we use UCB plus a a term that penalises violations of the constraint with high probability $1000(P[f<0]<0.7)$, where $P$ is computed by sampling the function $f$ from SkewGP.

In the experiments, we sample (using rejection sampling) a set of 100 random functions $g$ from a zero-mean GP with RBF kernel (with variance $2$ and lengthscale uniformly sampled in the interval $[1,2]$), which satisfies
$g(0)\geq0$. This ensures that $x=0$ is an initial safe input. This point serves as a starting point for both the algorithms.

The dependence between the kernel hyperparameters and the acquired data can lead to poor results in BO (especially in the first iterations). In fact, hyperparameters estimated by maximising the marginal likelihood can lead to a  GP estimate which does not have a calibrated uncertainty. In Safe BO \citep{berkenkamp2016safe,berkenkamp2016bayesian}, one critically relies on the uncertainty  to guarantee safety (avoiding constraint violation with high probability). As a consequence,  the hyperparameters are kept fixed and we treat the kernel as a prior over functions in the true Bayesian sense -- the kernel hyperparameters encode our prior knowledge about the functions. 
Therefore, in the experiment we set the variance of the RBF kernel to $2$ and the lengthscale to $1.5$.\footnote{The lengthscale can be different from the true one which is uniformly sampled in $[1,2]$.}

\begin{figure}
	\centering
	\begin{tabular}{|ll|}
		\hline
	\rotatebox{90}{{\small Iteration 0}} 
			\begin{minipage}{6cm}
		\includegraphics[height=2.3cm,trim={0.6cm 0.0cm 0.0cm 0.0cm }, clip]{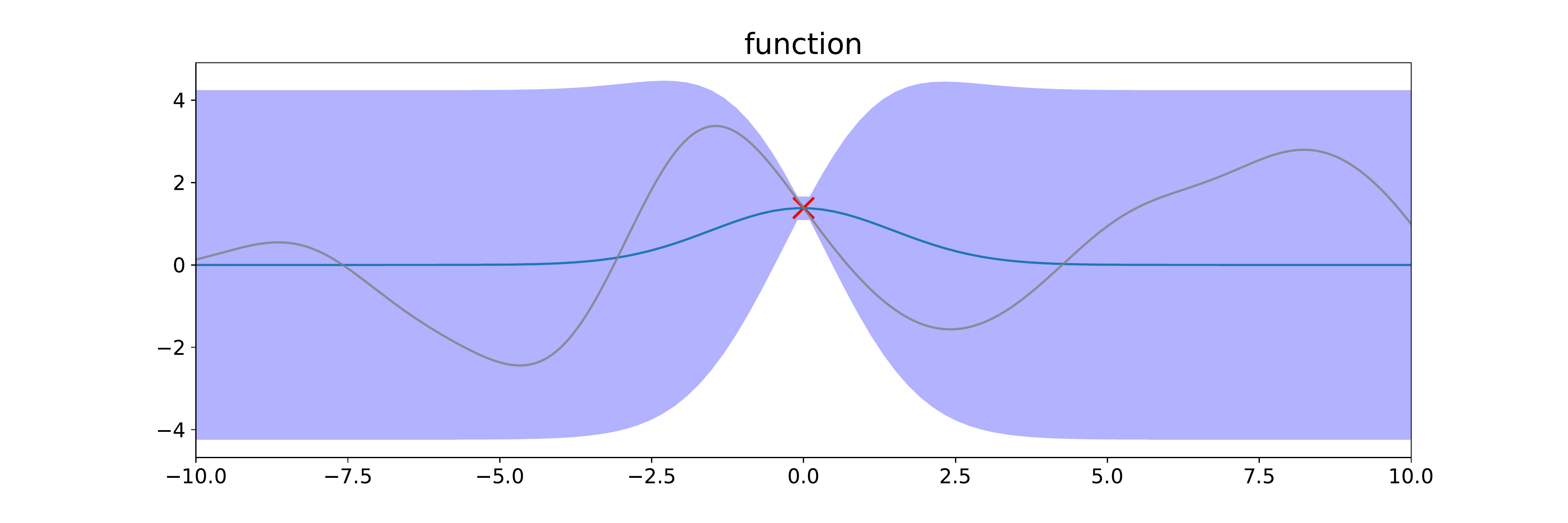}\vspace{-0.22cm}	\\		        \includegraphics[height=2.3cm,trim={0.6cm 0.cm 0.0cm 0.0cm }, clip]{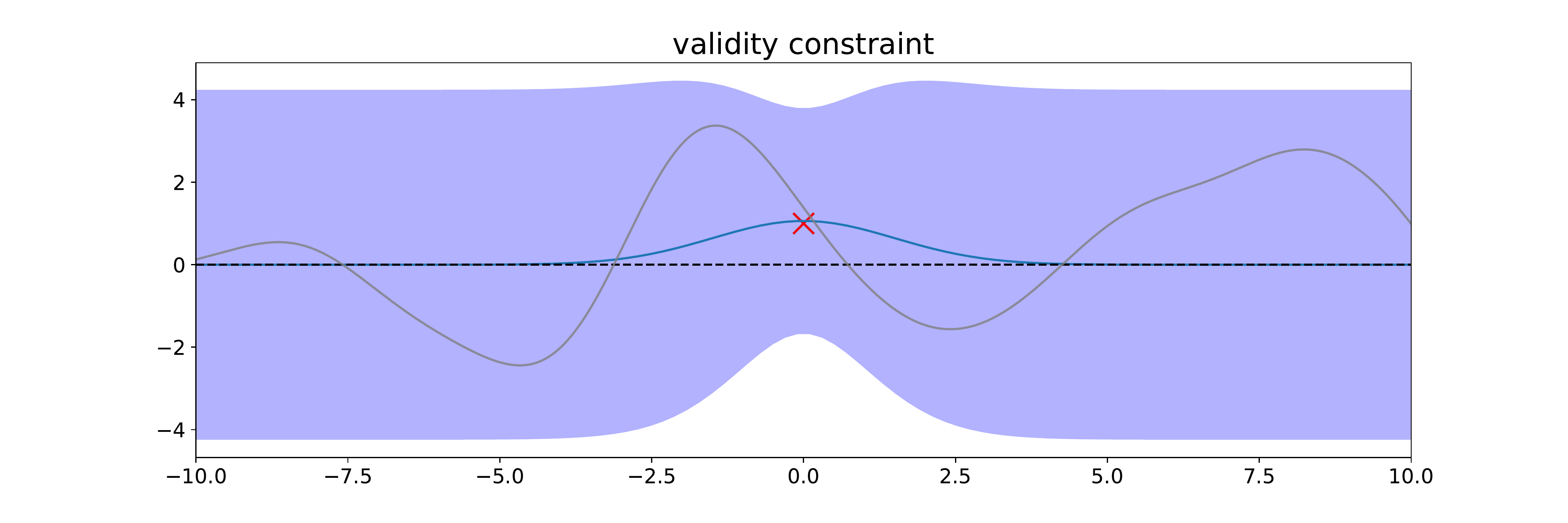}
		\end{minipage}&
		\begin{minipage}{6cm}
		\includegraphics[height=2.3cm,trim={0.0cm 0.0cm 0.0cm 0.0cm }, clip]{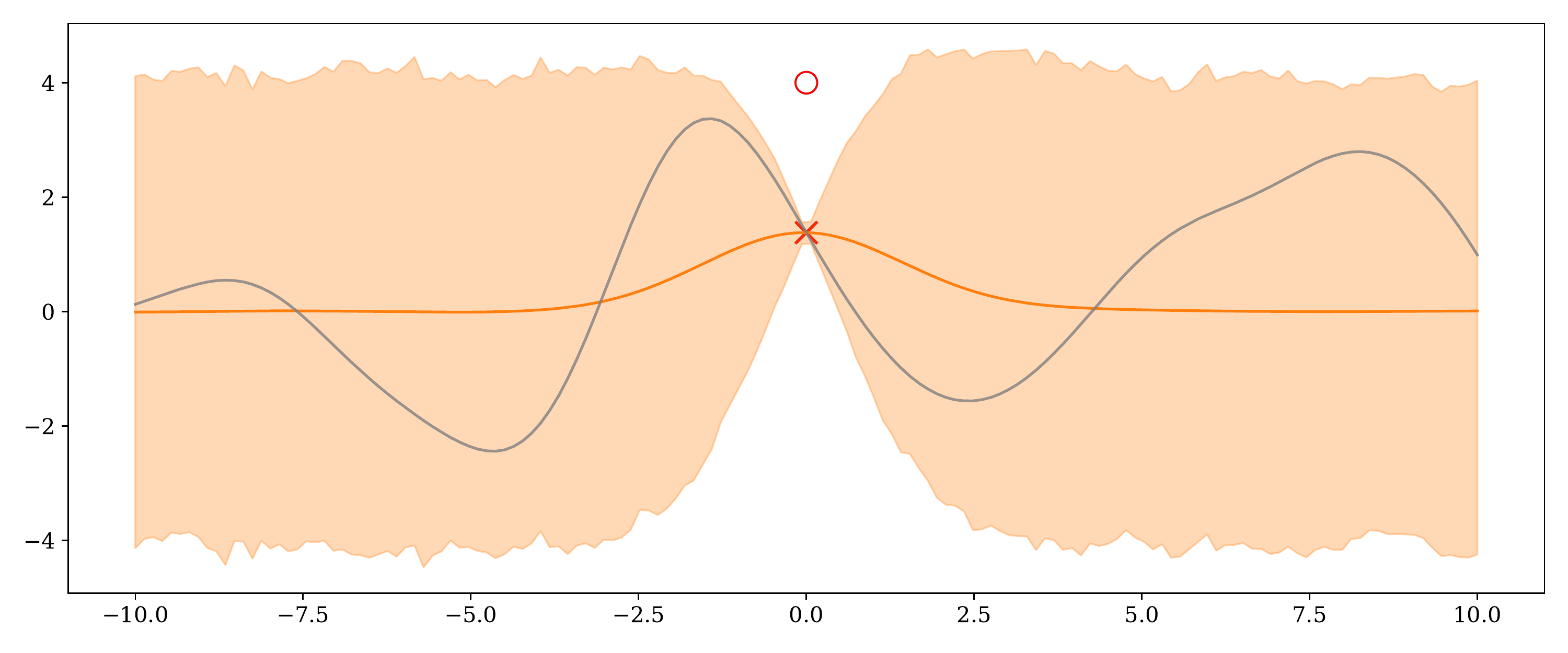}
		\end{minipage} \\
			\hline
				\rotatebox{90}{{\small Iteration 1}} 
					\begin{minipage}{6cm}
		\includegraphics[height=2.3cm,trim={0.6cm 0.0cm 0.0cm 0.0cm }, clip]{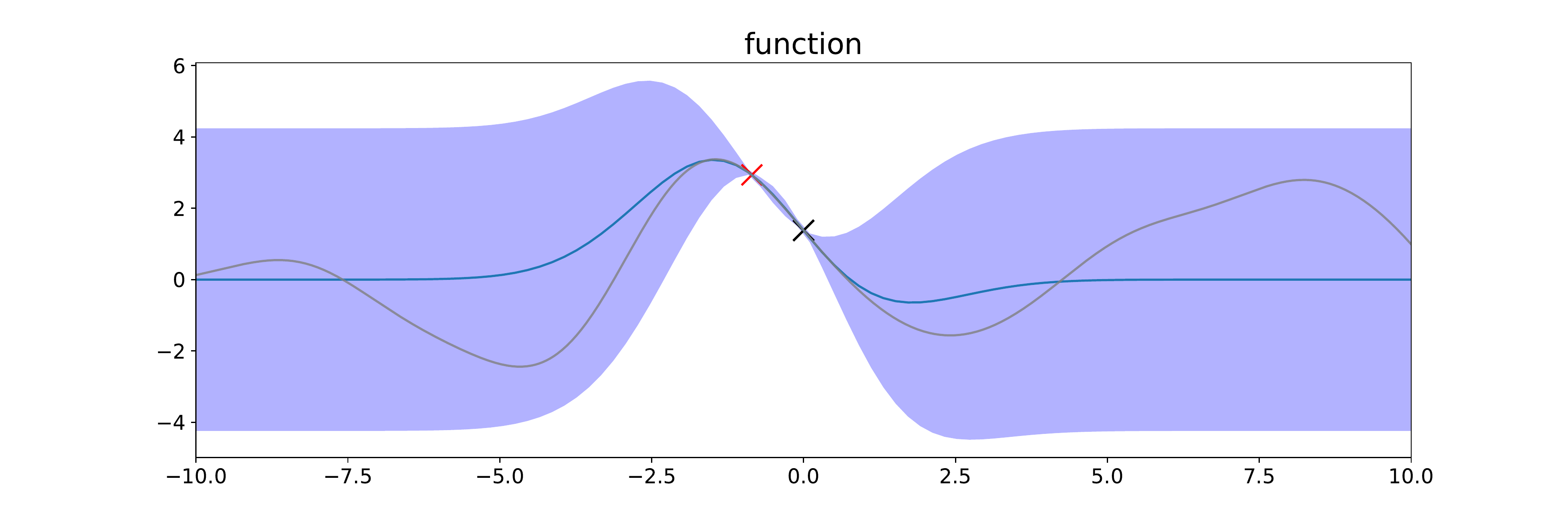}\vspace{-0.22cm}	\\		        \includegraphics[height=2.3cm,trim={0.6cm 0.cm 0.0cm 0.0cm }, clip]{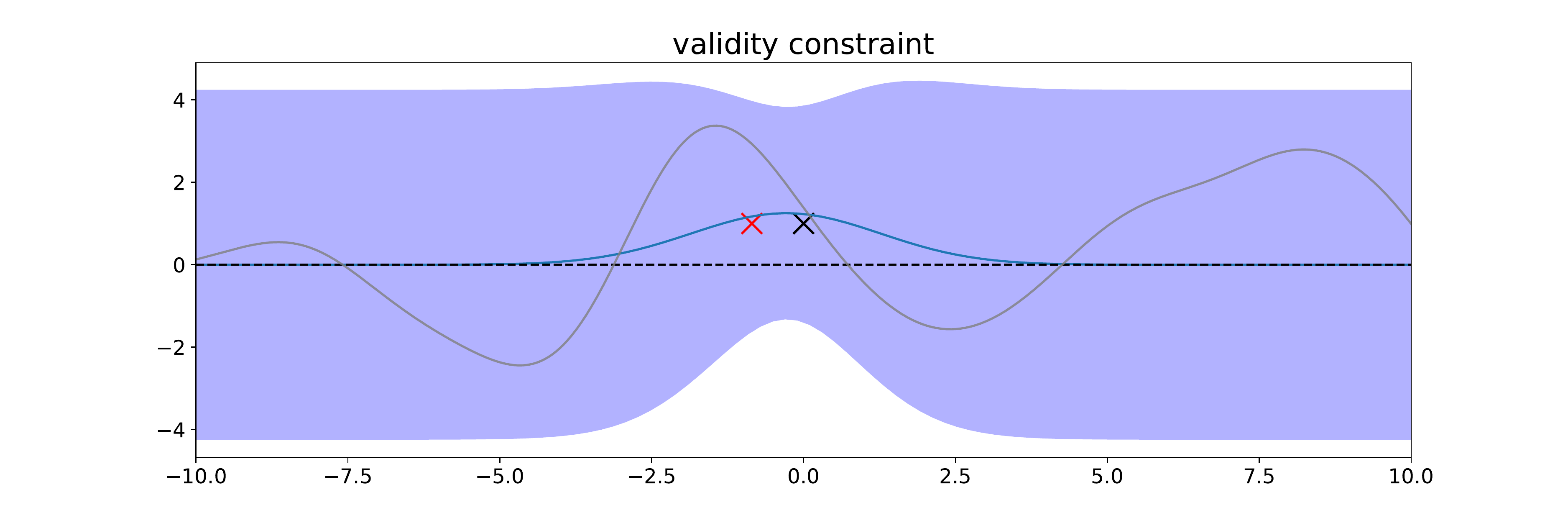}
		\end{minipage}&
		\begin{minipage}{6cm}
		\includegraphics[height=2.3cm,trim={0.0cm 0.0cm 0.0cm 0.0cm }, clip]{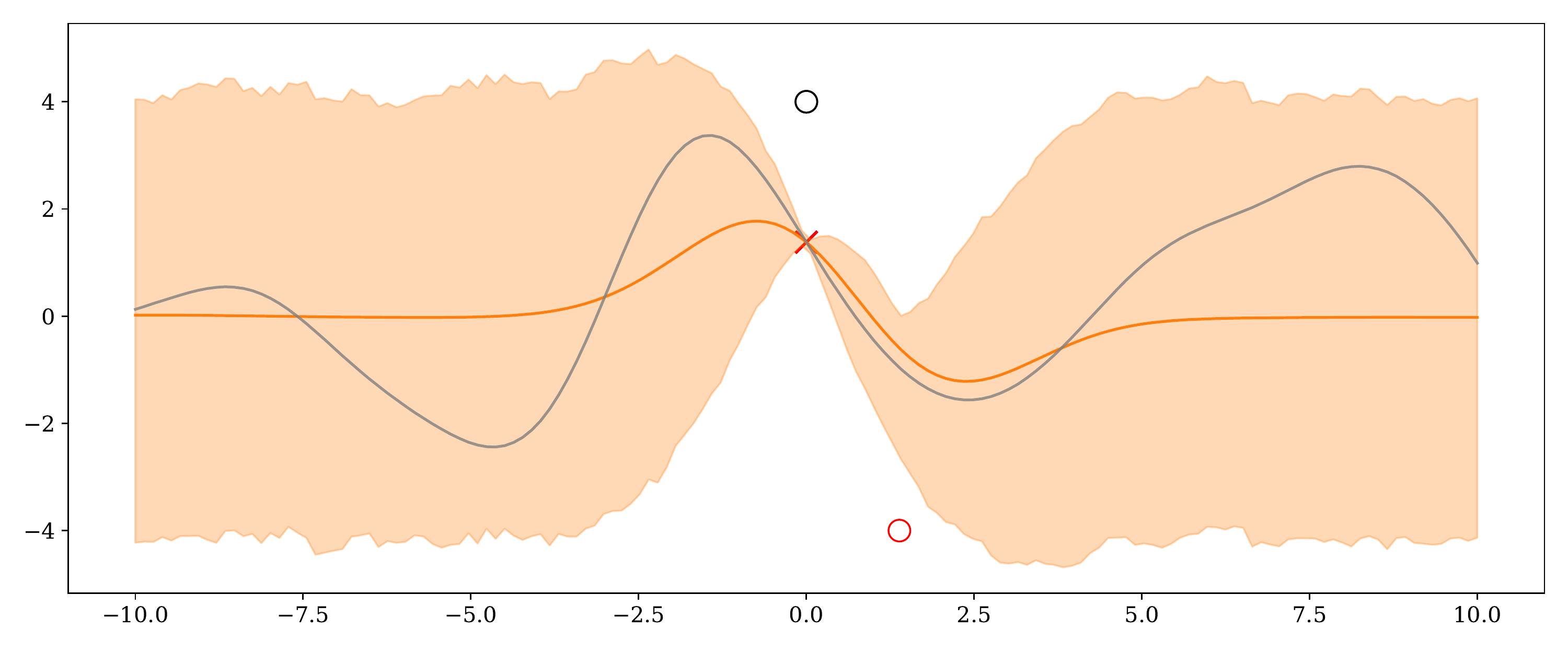}
		\end{minipage}  \\
		\hline
			\rotatebox{90}{{\small Iteration 2}} 
					\begin{minipage}{6cm}
		\includegraphics[height=2.3cm,trim={0.6cm 0.0cm 0.0cm 0.0cm }, clip]{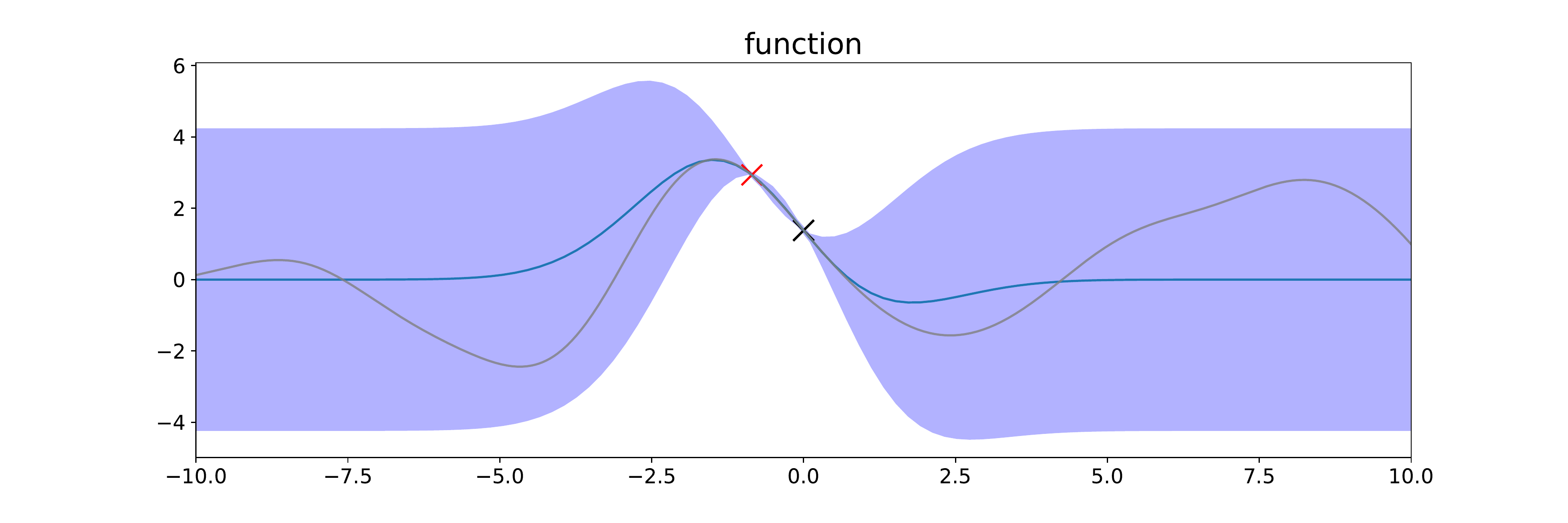}\vspace{-0.22cm}	\\		        \includegraphics[height=2.3cm,trim={0.6cm 0.cm 0.0cm 0.0cm }, clip]{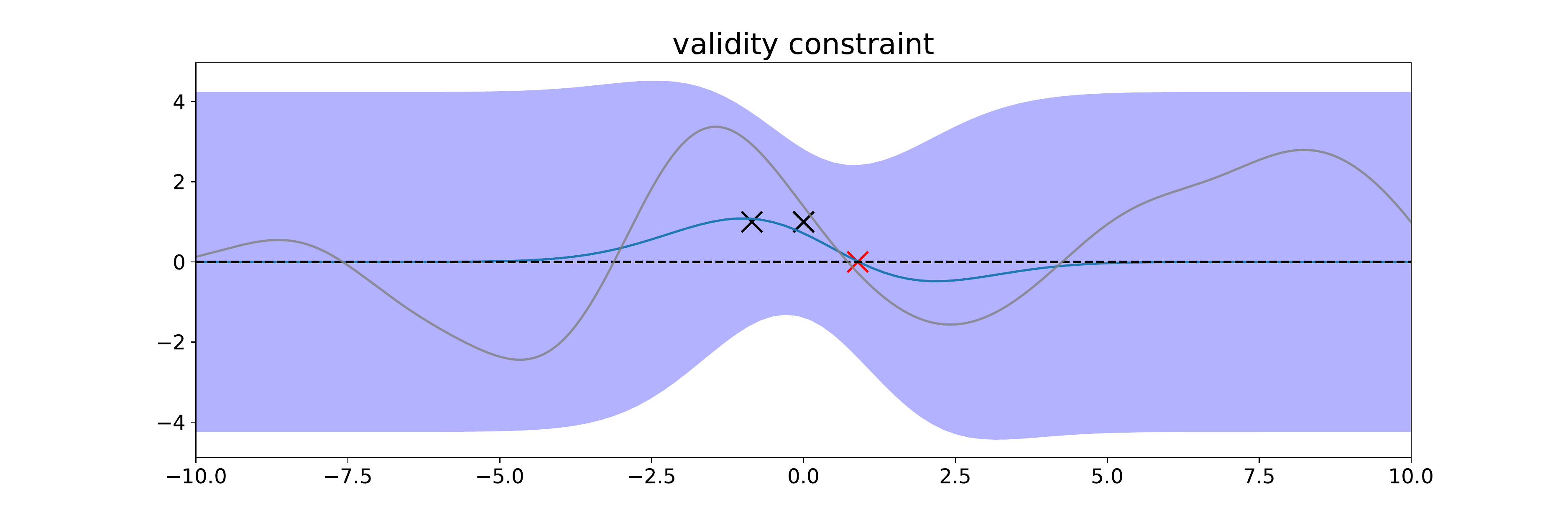}
		\end{minipage}&
		\begin{minipage}{6cm}
		\includegraphics[height=2.3cm,trim={0.0cm 0.0cm 0.0cm 0.0cm }, clip]{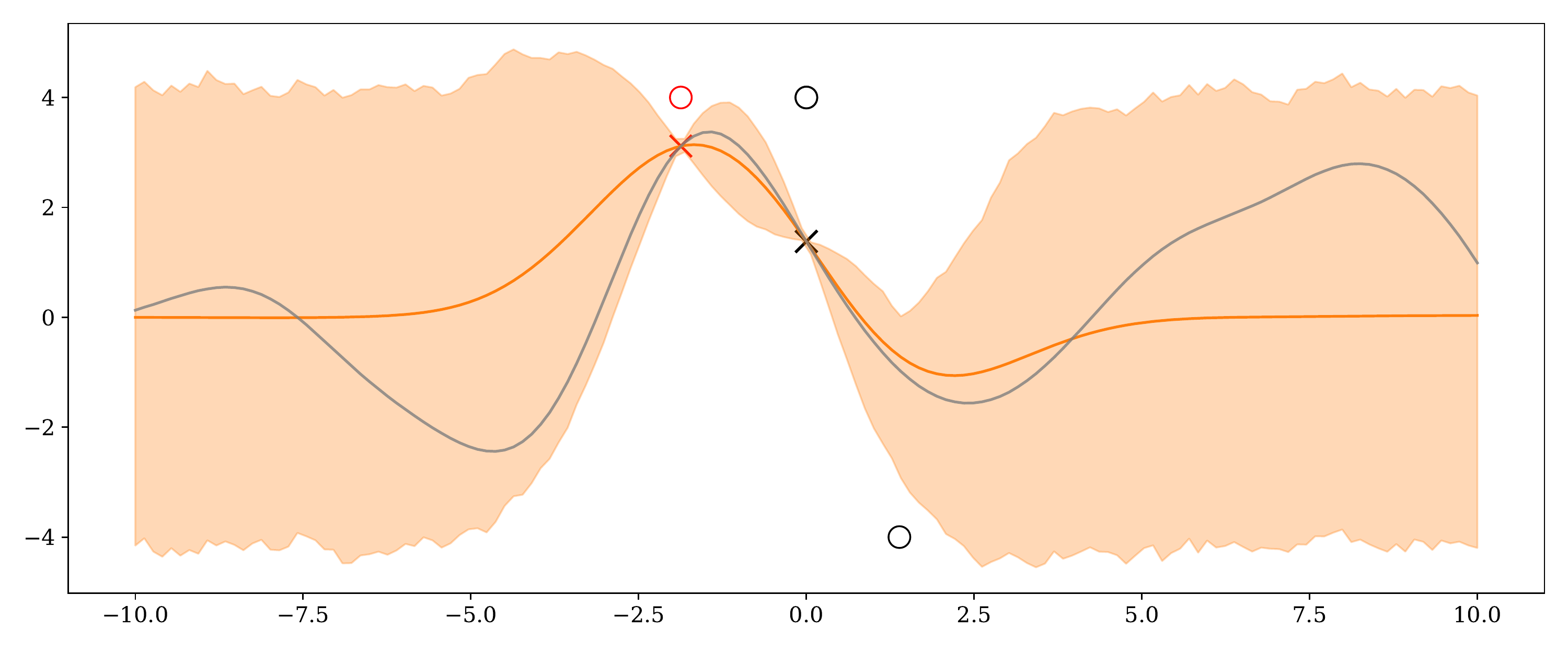}
		\end{minipage}  \\
				\hline
					\rotatebox{90}{{\small Iteration 3}} 
					\begin{minipage}{6cm}
		\includegraphics[height=2.3cm,trim={0.6cm 0.0cm 0.0cm 0.0cm }, clip]{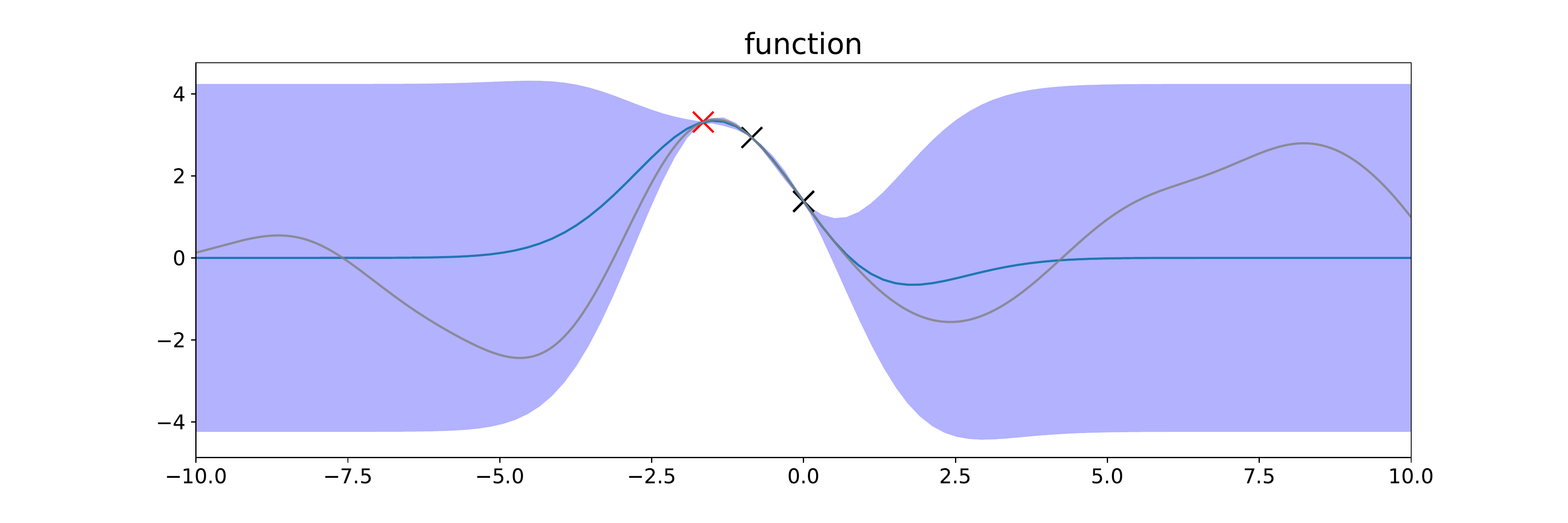}\vspace{-0.22cm}	\\		        \includegraphics[height=2.3cm,trim={0.6cm 0.cm 0.0cm 0.0cm }, clip]{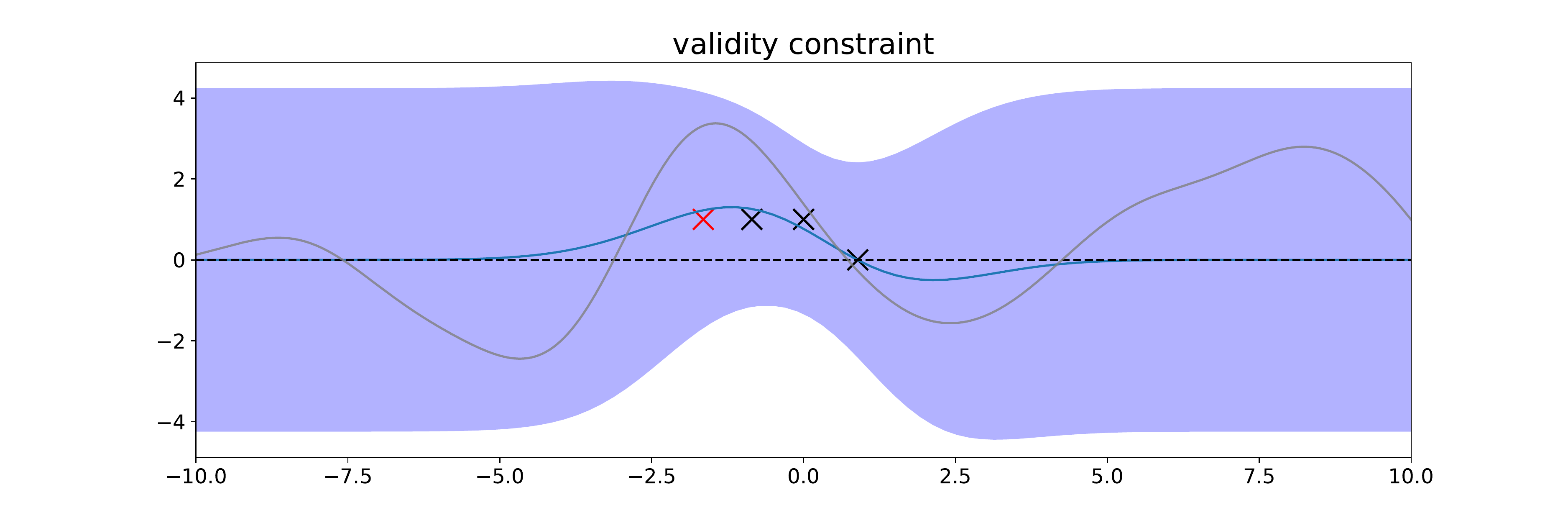}
		\end{minipage}&
		\begin{minipage}{6cm}
		\includegraphics[height=2.3cm,trim={0.0cm 0.0cm 0.0cm 0.0cm }, clip]{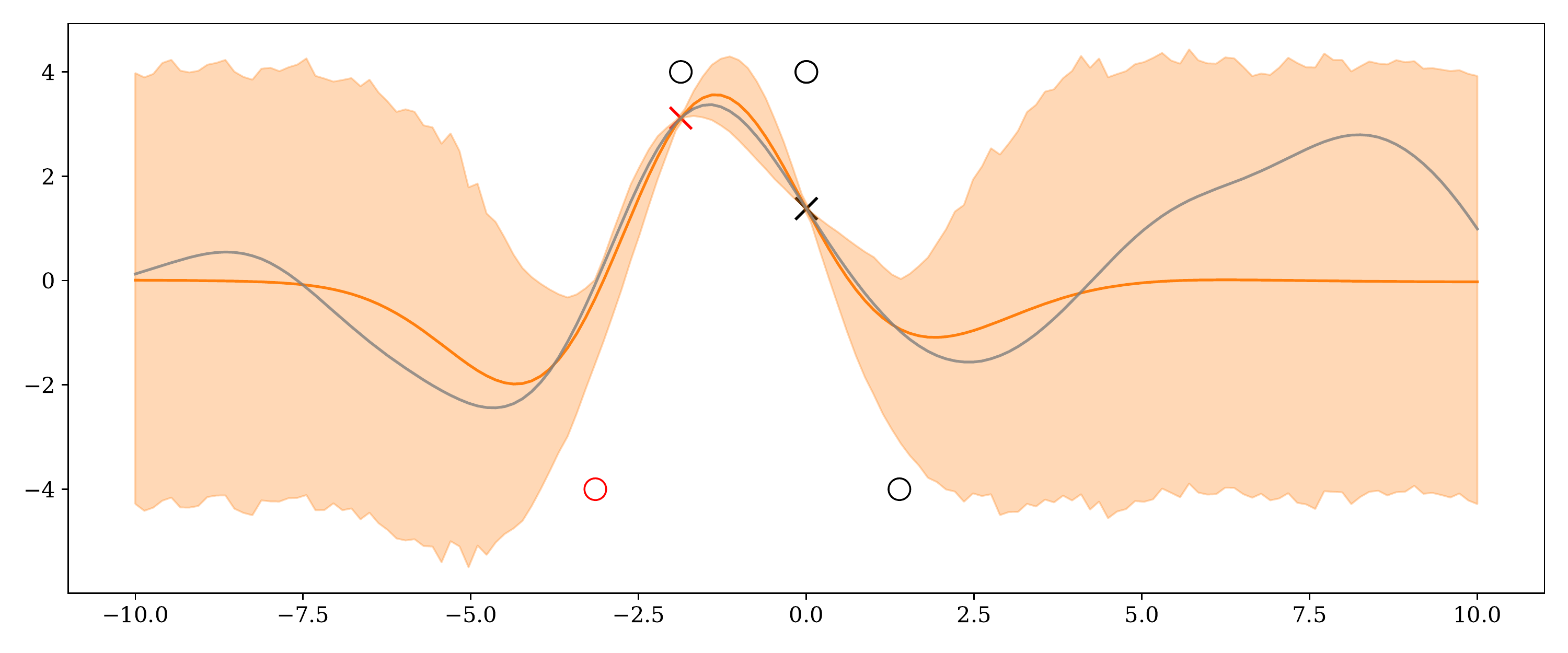}
		\end{minipage}  \\
			\hline
		\end{tabular}
	\caption{Iterations 0-3, GP-SafeOpt (left) and SkewGP-SafeOpt (right).}
	\label{fig:6}
\end{figure}

\begin{figure}
	\centering
	\begin{tabular}{|ll|}
	\hline
		\rotatebox{90}{{\small Iteration 4}} 
			\begin{minipage}{6cm}
		\includegraphics[height=2.3cm,trim={0.6cm 0.0cm 0.0cm 0.0cm }, clip]{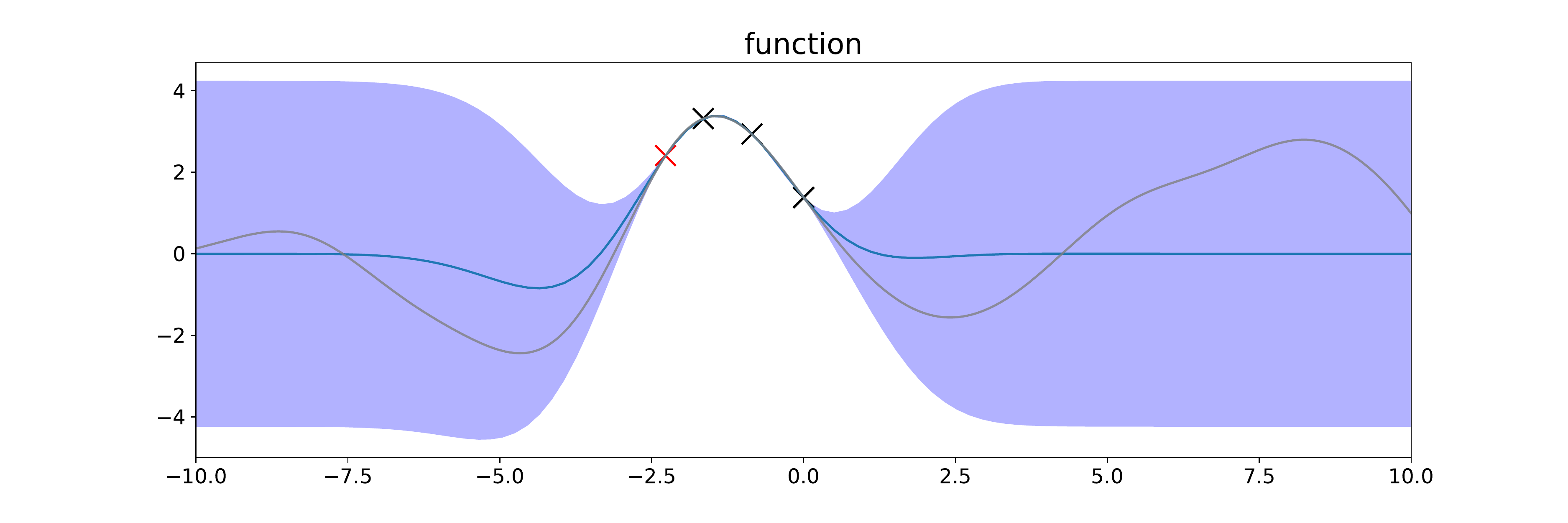}\vspace{-0.22cm}	\\		        \includegraphics[height=2.3cm,trim={0.6cm 0.cm 0.0cm 0.0cm }, clip]{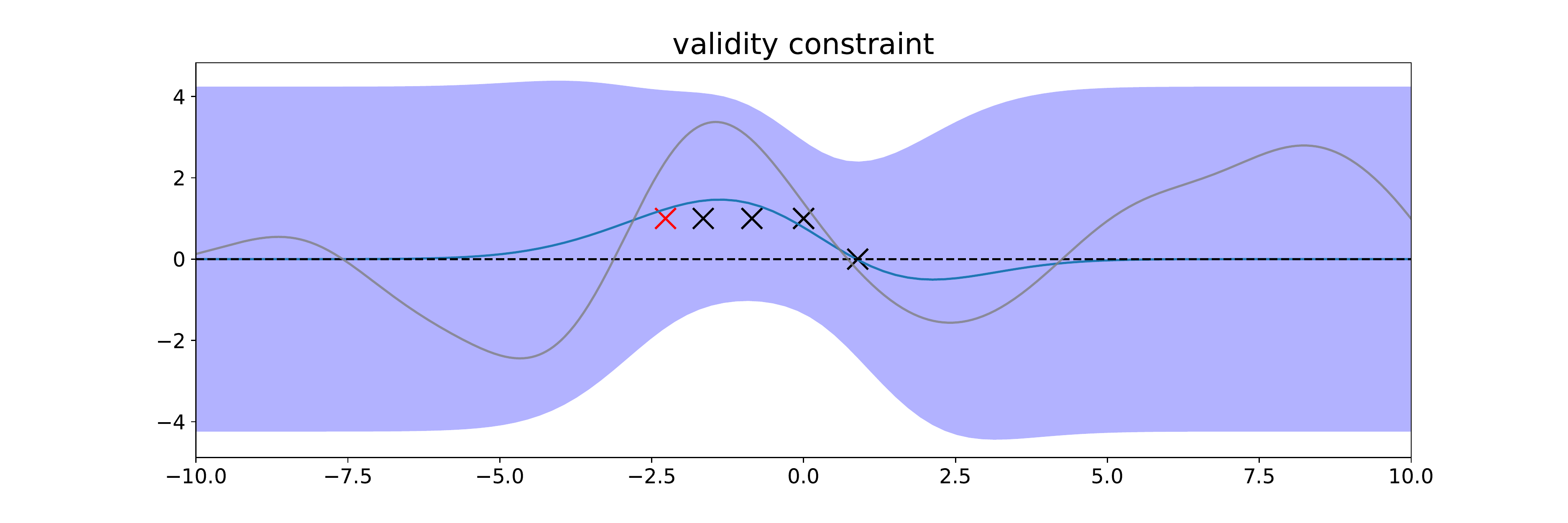}
		\end{minipage}&
		\begin{minipage}{6cm}
		\includegraphics[height=2.3cm,trim={0.0cm 0.0cm 0.0cm 0.0cm }, clip]{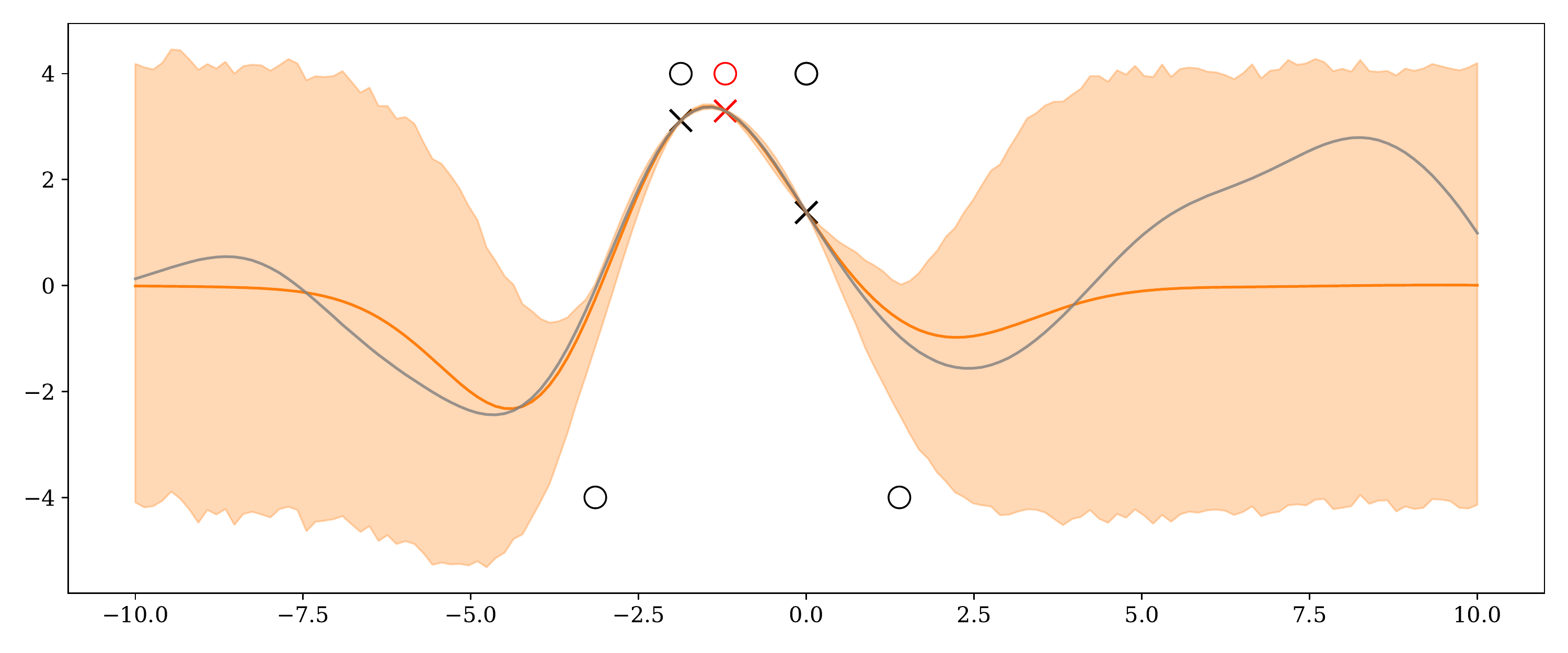}
		\end{minipage} \\
			\hline
				\rotatebox{90}{{\small Iteration 5}} 
					\begin{minipage}{6cm}
		\includegraphics[height=2.3cm,trim={0.6cm 0.0cm 0.0cm 0.0cm }, clip]{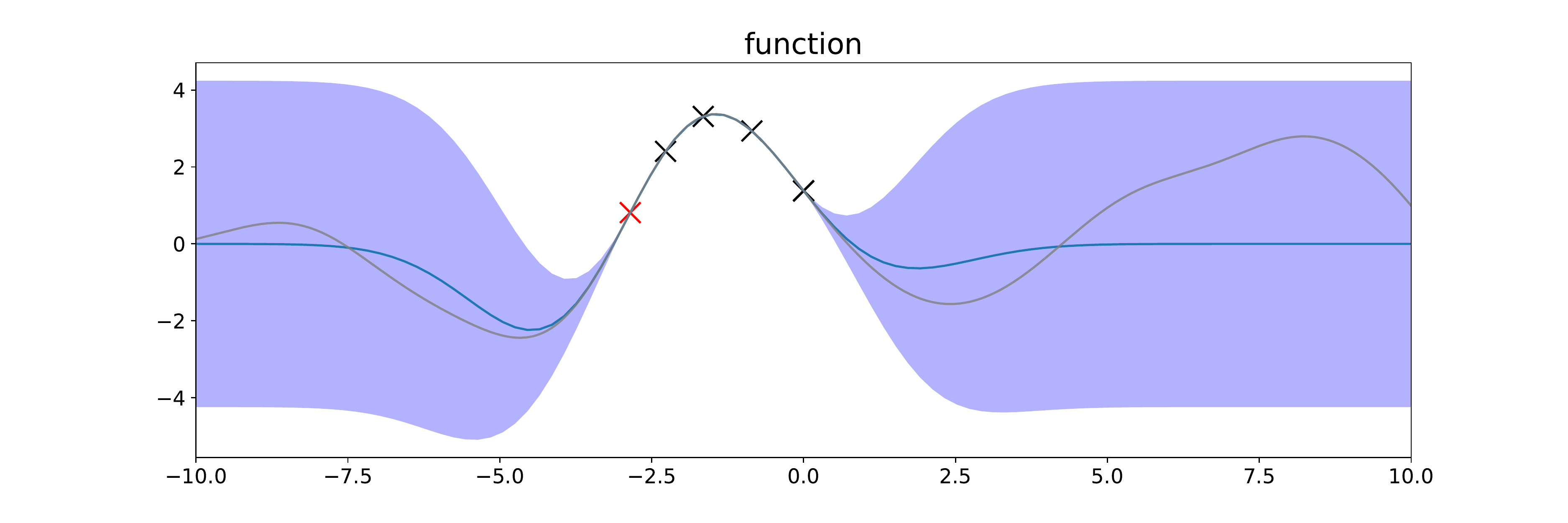}\vspace{-0.22cm}	\\		        \includegraphics[height=2.3cm,trim={0.6cm 0.cm 0.0cm 0.0cm }, clip]{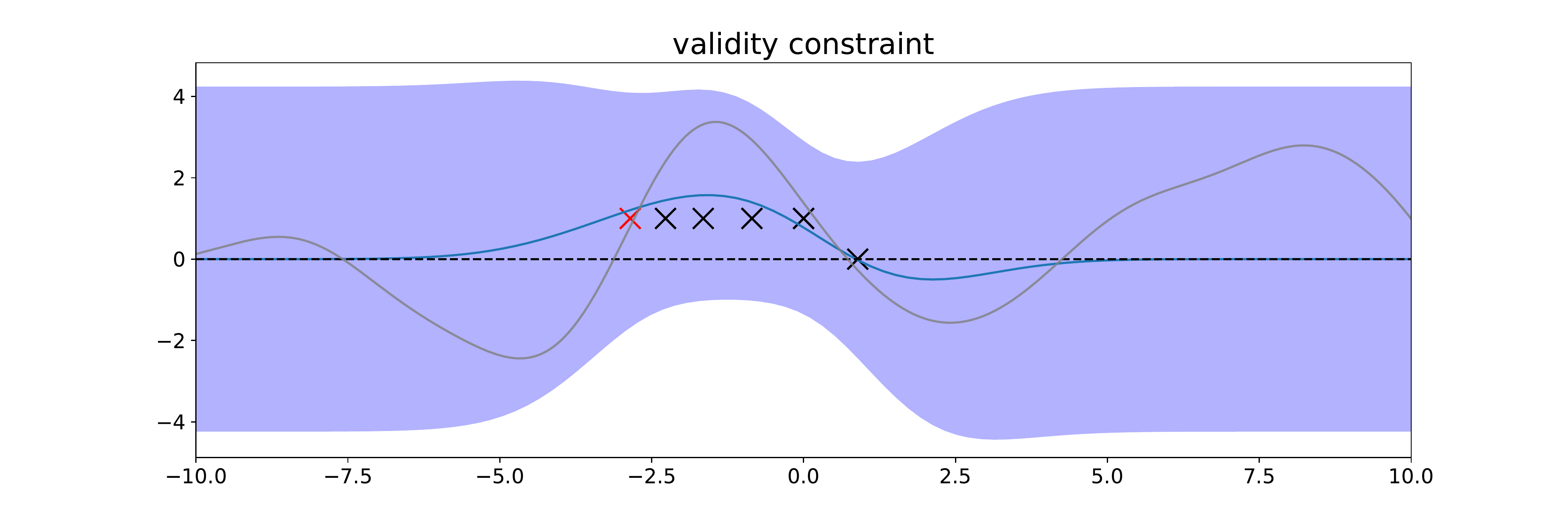}
		\end{minipage}&
		\begin{minipage}{6cm}
		\includegraphics[height=2.3cm,trim={0.0cm 0.0cm 0.0cm 0.0cm }, clip]{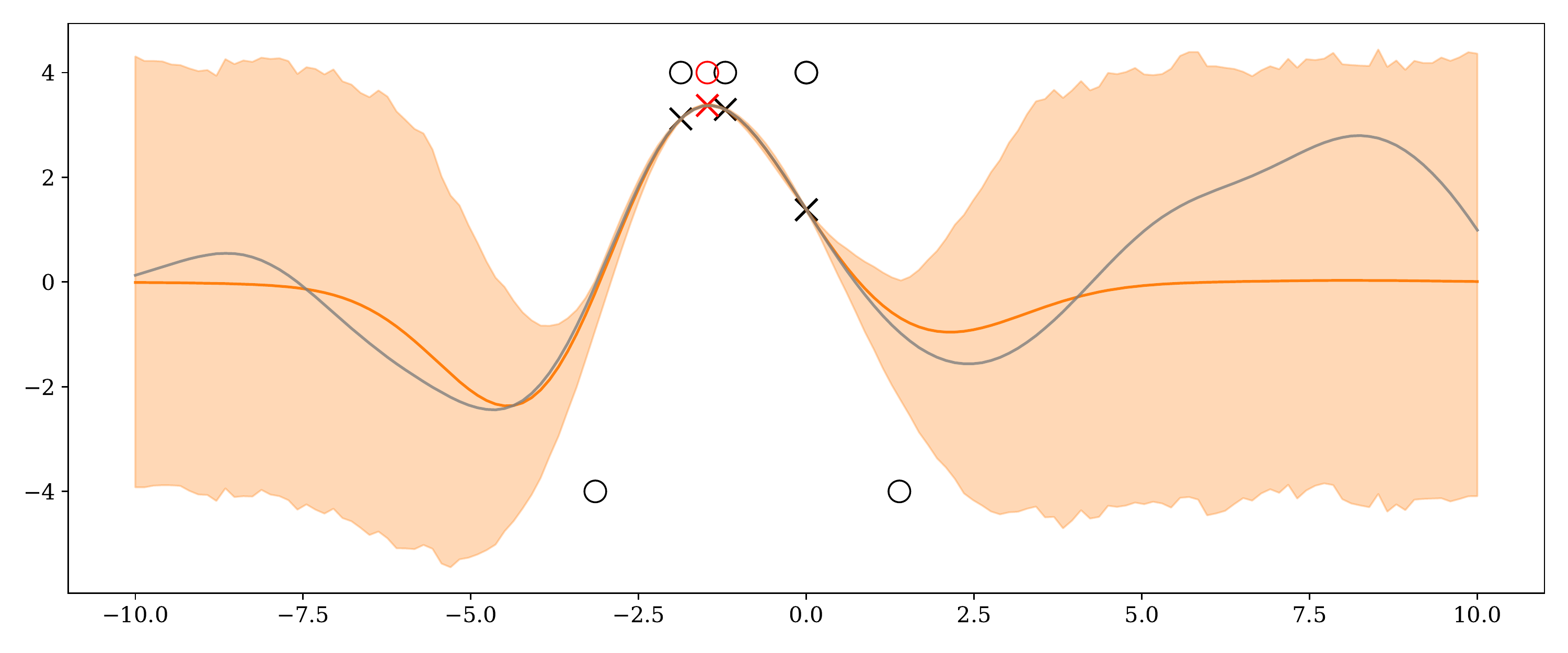}
		\end{minipage}  \\
		\hline
			\rotatebox{90}{{\small Iteration 6}} 
					\begin{minipage}{6cm}
		\includegraphics[height=2.3cm,trim={0.6cm 0.0cm 0.0cm 0.0cm }, clip]{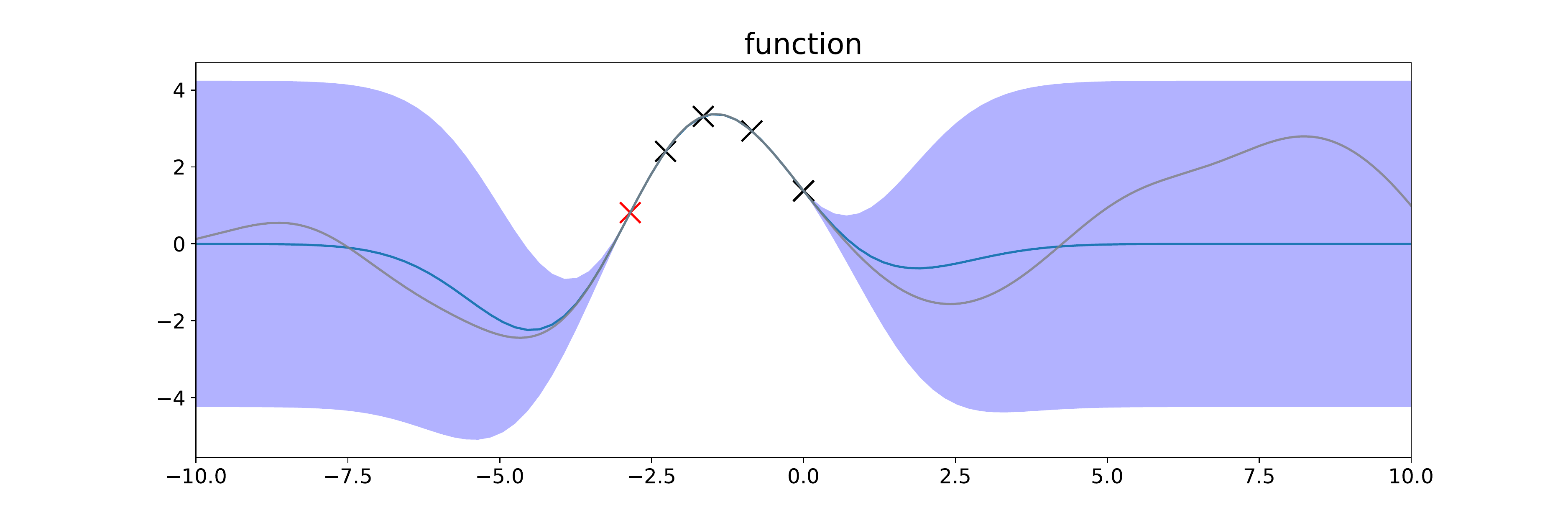}\vspace{-0.22cm}	\\		        \includegraphics[height=2.3cm,trim={0.6cm 0.cm 0.0cm 0.0cm }, clip]{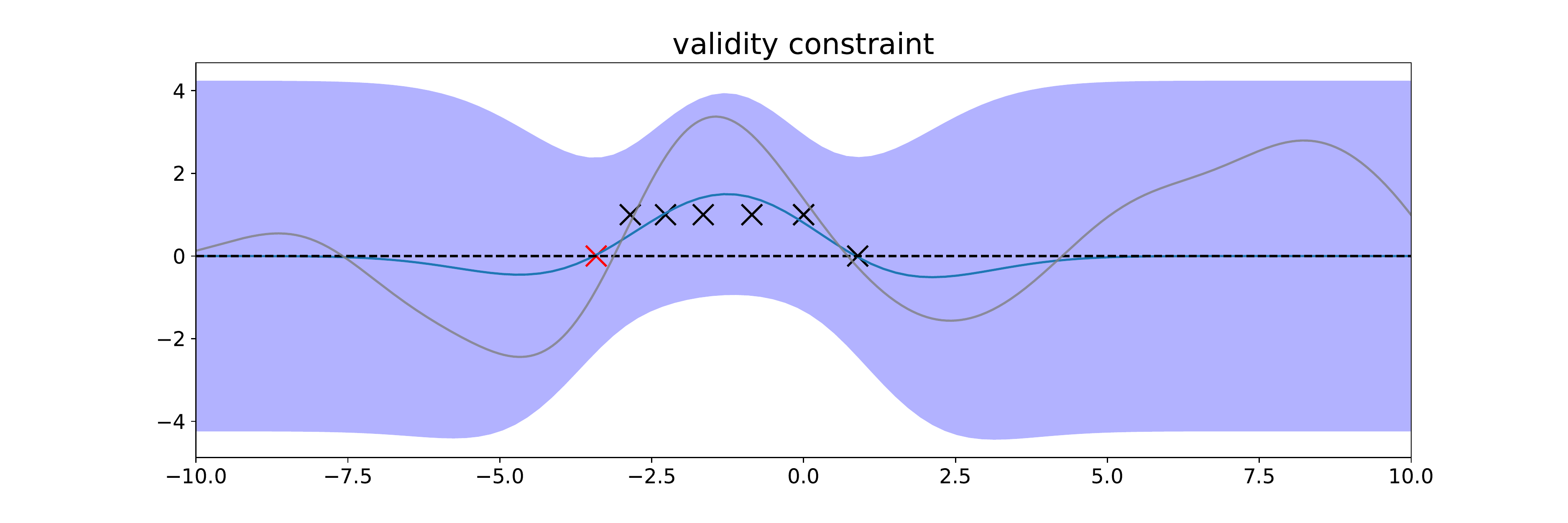}
		\end{minipage}&
		\begin{minipage}{6cm}
		\includegraphics[height=2.3cm,trim={0.0cm 0.0cm 0.0cm 0.0cm }, clip]{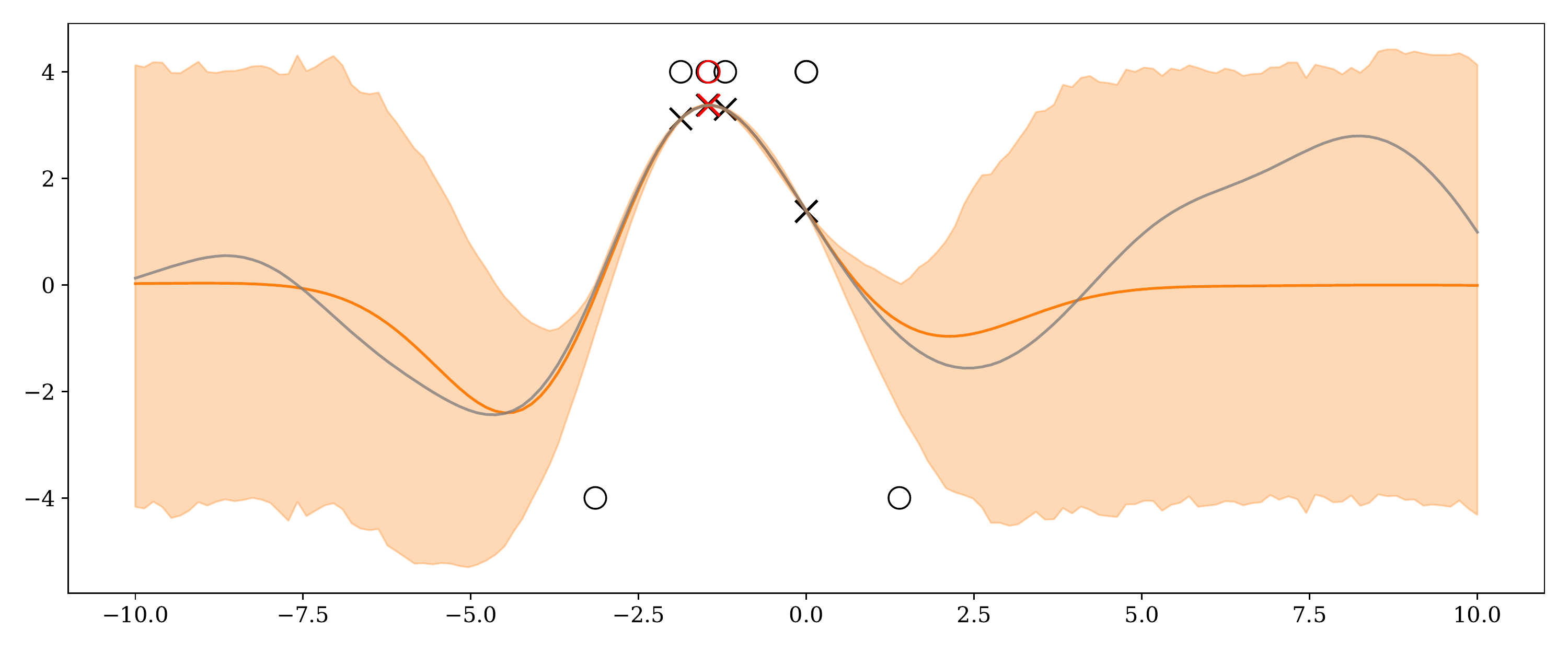}
		\end{minipage}  \\
				\hline
					\rotatebox{90}{{\small Iteration 7}} 
					\begin{minipage}{6cm}
		\includegraphics[height=2.3cm,trim={0.6cm 0.0cm 0.0cm 0.0cm }, clip]{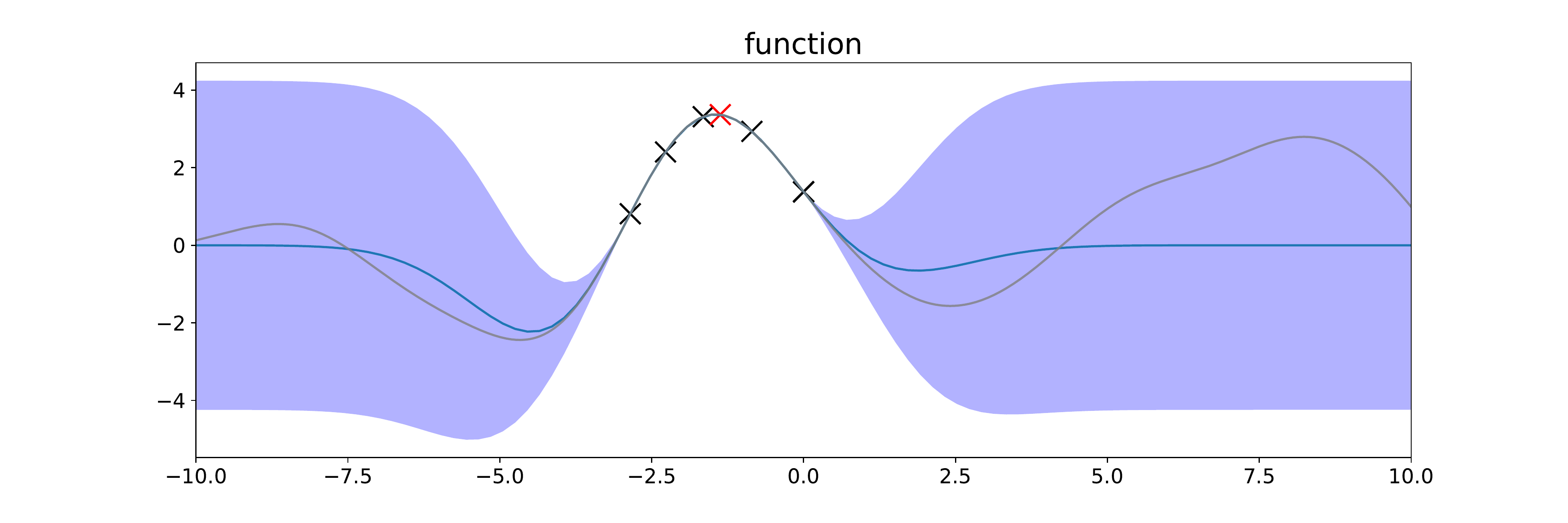}\vspace{-0.22cm}	\\		        \includegraphics[height=2.3cm,trim={0.6cm 0.cm 0.0cm 0.0cm }, clip]{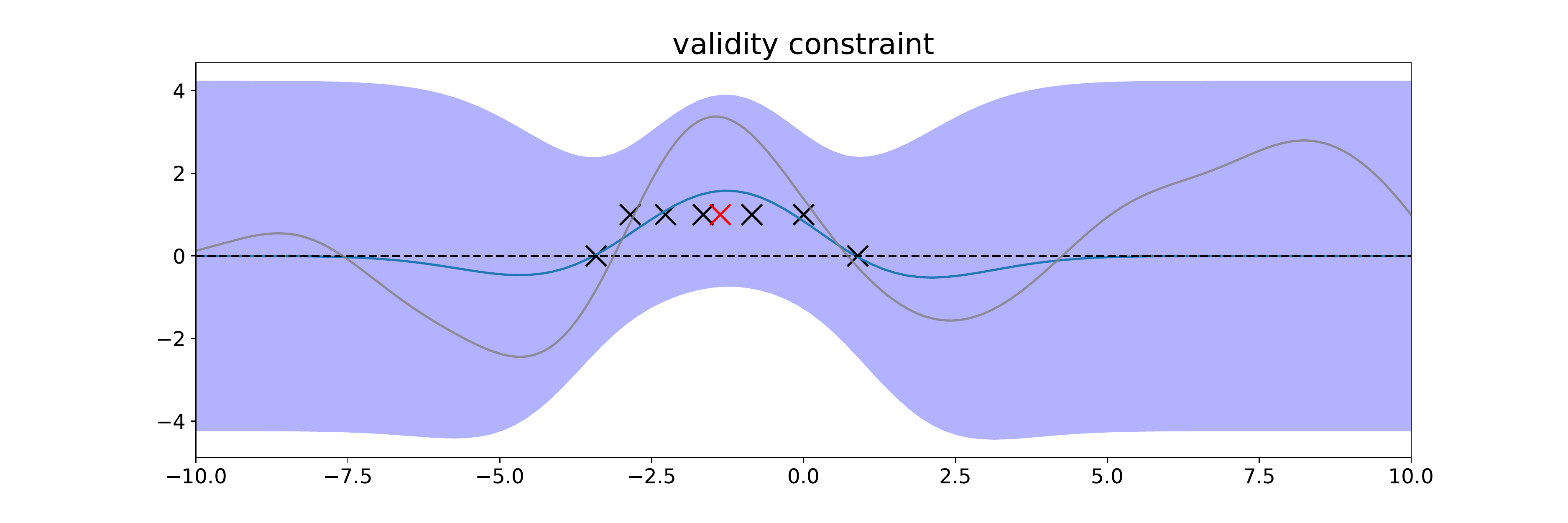}
		\end{minipage}&
		\begin{minipage}{6cm}
		\includegraphics[height=2.3cm,trim={0.0cm 0.0cm 0.0cm 0.0cm }, clip]{img/Skewgp_safe6.pdf}
		\end{minipage}  \\
	\hline
		\end{tabular}
	\caption{Iterations 4-7, GP-SafeOpt (left) and SkewGP-SafeOpt (right).}
	\label{fig:7}
\end{figure}

Figure \ref{fig:6} and \ref{fig:7} report 7 iterations of the two approaches, which we call as SafeOpt and SkewGP-BO, for one of the 100 trials. Figure \ref{fig:6}(Iter.\ 0, left) shows
the true function $g$ in gray; the estimated GP regression model (mean and 99.7 credible interval)
after observing the safe point $x=0$ (top plot); the estimated GP classifier (mean and 99.7 credible interval of the latent function) after observing the valid point $x=0$ (bottom plot). The x in the plots corresponds the value of $g(0)$ in the top-plot and to the binary observation $1$ (which means valid) in the bottom-plot.
Figure \ref{fig:6}(Iter.\ 0, right) shows the posterior mean and 99.7 credible interval for SkewGP, which solves the mixed numeric and binary regression problem. We use  xs to mark  numeric observations and circle to mark binary observations (the ordinate $+4$ corresponds to valid  and  $-4$ to non-valid). 

At iteration 1, SkewGP-BO selects a \textit{non-valid} point Figure \ref{fig:6}(Iter.\ 1, right), while SafeOpt selects a \textit{valid} one, finding a better maximum Figure \ref{fig:6}(Iter.\ 1, left).

At iteration 2, SkewGP-BO selects a \textit{valid} point and finds a better maximum Figure \ref{fig:6}(Iter.\ 2, right), while GP-SafeOpt violates the constraint Figure \ref{fig:6}(Iter.\ 2, left). 

At iteration 3, SkewGP-BO selects a \textit{non-valid} point Figure \ref{fig:6}(Iter.\ 3, right), while SafeOpt selects a \textit{valid} point \ref{fig:6}(Iter.\ 3, left). 

At iteration 4 and 5, SkewGP-BO finds the maximum  Figure \ref{fig:7}(Iter.\ 4 and 5, right), while SafeOpt explores the \textit{valid} region Figure \ref{fig:7}(Iter.\ 4 and 5, left). 

At iteration 6, SafeOpt violates the constraint Figure \ref{fig:7}(Iter.\ 6, left).

At iteration 7, SafeOpt finally finds the maximum Figure \ref{fig:7}(Iter.\ 7, left).

\begin{figure}[htp!]
\centering
   \includegraphics[width=8.5cm]{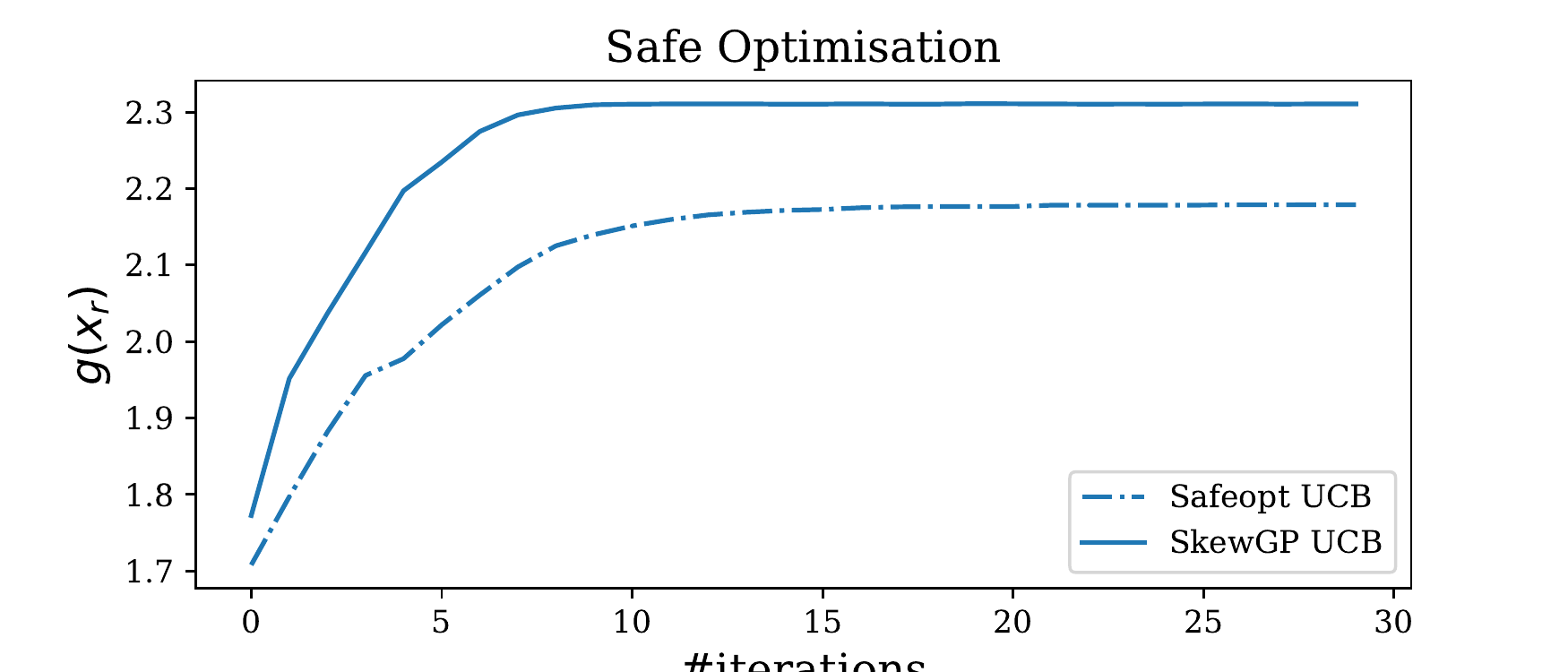}
   \caption{Averaged results over 100 trials for GP-SafeOpt versus SkewGP.  The x-axis represents the number of iterations and the y-axis represents the value  of the true objective function at the current optimum ${\bf x}_r$. }
   \label{fig:8}
\end{figure}

Figure \ref{fig:8} reports the averaged results over $100$ trials for SafeOpt versus SkewGP-BO.  The x-axis represents the number of iterations and the y-axis represents the value  of the true objective function at the current optimum ${\bf x}_r$. SkewGP-BO achieves the best performance. 
Both SafeOpt and SkewGP-BO are very unlikely to explore the unsafe region  ($51$  violations for SafeOpt and  $53$ for SkewGP-BO in the $3000$ acquisitions). This shows that, when Safe BO can be modelled as a mixed problem,\footnote{The framework  for safe BO in \citep{berkenkamp2016safe,berkenkamp2016bayesian} is more general and cannot always be expressed as a mixed problem.} SkewGP provides faster convergence than using two  separated surrogated models (GPs) for objective and constraint.

\section{Conclusions}
\label{sec:conclusions}
	We  have shown that Skew Gaussian process (SkewGP) are conjugate to the normal likelihood, the probit affine likelihood and with their product, and provided  marginals and closed form conditionals. SkewGPs allow us to compute the exact posterior in all these cases and, therefore, provide an accurate representation of the uncertainty. From a theoretical point of view, this shows that SkewGPs encompass GPs in nonparametric regression, classification, preference learning and mixed problems. From a practical point of view, we compared SkewGPs with GPs (Laplace and Expectation Propagation approximation) in two tasks, Bayesian Active Learning and Bayesian Optimization, where a wrong representation of  uncertainty can lead to a significant performance degradation. SkewGP  achieved an  improved  performance over GPs (Laplace's method and Expectation Propagation approximations).
	
	As future work, we plan to investigate the possibility of using inducing points, as for sparse GPs \citep{quinonero2005unifying,snelson2006sparse,pmlrv5titsias09a,Hensman2013,hernandez2016scalable,bauer2016understanding,SCHURCH2020}, to reduce the computational load for matrix operations (complexity $O(n^3)$ with storage demands of $O(n^2)$). We also plan to derive tighter approximations of the marginal likelihood.
	Moreover, we plan to study different ways to parametrize the skewness matrix $\Delta$ and select the prior parameters ${\bm \gamma},\Gamma$, which allows one to fully exploit the flexibility of SkewGPs.
		
\appendix
\clearpage


\section{Proofs of the results in the paper}

\paragraph{Proof Lemma \ref{lemma:Normal} and Corollary
\ref{co:Normal}}
The likelihood is $\phi_{m_r}(Y-Cf;R)$, 
where $Y\in \mathbb{R}^{m_a}$ is the vector of observations;
$f\in \mathbb{R}^n$ is the vector of function values at the input points $\bx_i$ for $i=1,\dots,n$ and the prior is:
\begin{equation}
\label{eq:sunpp}
p(f) = \phi_n(f-\bxi;\Omega)\frac{\Phi_s\left(\bgamma+\Delta^T\bar{\Omega}^{-1}\Domega^{-1}(f-\bxi);\Gamma-\Delta^T\bar{\Omega}^{-1}\Delta\right)}{\Phi_s\left(\bgamma;\Gamma\right)}. 
\end{equation}

First, note that
$$
\phi_{m_r}(Y-Cf;R)\phi_{n}(f;\bxi,\Omega)\propto \phi_n(f-\bxi_p;\Omega_p),
$$
with 
\begin{align}
\bxi_p  &=\bxi+\Omega C^T(C\Omega C^T+R)^{-1}(Y-C\xi),\\
\Omega_p &= \Omega-\Omega C^T(C\Omega C^T+R)^{-1}C\Omega.
\end{align}
Now consider
$\Phi_s\left(\gamma+\Delta^T\bar{\Omega}^{-1}\Domega^{-1}(f-\xi);\Gamma-\Delta^T\bar{\Omega}^{-1}\Delta\right)$
and observe that
$$
\begin{aligned}
 \gamma+\Delta^T\bar{\Omega}^{-1}\Domega^{-1}(f-\xi)
 &=\gamma+\Delta^T\bar{\Omega}^{-1}\Domega^{-1}(f-\xi_p+\xi_p-\xi)\\
  &=\gamma+\Delta^T\bar{\Omega}^{-1}\Domega^{-1}(\xi_p-\xi)\\&+\Delta^T\bar{\Omega}^{-1}\Domega^{-1}(f-\xi_p)\\
    &=\gamma+\Delta^T\bar{\Omega}^{-1}\Domega^{-1}(\xi_p-\xi)\\
    &+\Delta^T\bar{\Omega}^{-1}\Domega^{-1}{\Domega}_p\bar{\Omega}_p\bar{\Omega}_p^{-1}{\Domega}_p^{-1}(f-\xi_p)\\
     &=\gamma_p +\Delta_p^T\bar{\Omega}_p^{-1}{\Domega}_p^{-1}(f-\xi_p)\\
\end{aligned}
$$
with 
\begin{align}
\gamma_p  &= \gamma+\Delta^T\bar{\Omega}^{-1}\Domega^{-1}(\xi_p-\xi)\\
\Delta_p^T &= \Delta^T\bar{\Omega}^{-1}\Domega^{-1}{\Domega}_p\bar{\Omega}_p
\end{align}
Finally observe that
$$
\begin{aligned}
\Gamma-\Delta^T\bar{\Omega}^{-1}\Delta &= \Gamma-\Delta^T\bar{\Omega}^{-1}\Delta +\Delta_p^T\bar{\Omega}_p^{-1}\Delta_p -\Delta_p^T\bar{\Omega}_p^{-1}\Delta_p=\Gamma_p -\Delta_p^T\bar{\Omega}_p^{-1}\Delta_p
\end{aligned}
$$
with $
\Gamma_p  = \Gamma-\Delta^T\bar{\Omega}^{-1}\Delta +\Delta_p^T\bar{\Omega}_p^{-1}\Delta_p
$. 
Putting everything together
$$
p(f|Y)=\phi_n(f-\xi_p;\Omega_p)\frac{\Phi_s\left(\gamma_p+\Delta_p^T\bar{\Omega}_p^{-1}{\Domega}_p^{-1}(f-\xi_p);\Gamma_p-\Delta_p^T\bar{\Omega}_p^{-1}\Delta_p\right)}{\Phi_s\left(\gamma_p;\Gamma_p\right)}
$$
%

By considering the derivations  of the above expression, we can derive the marginal likelihood
$$
\phi_n(Y-C\bxi;C\Omega C^T+R)\frac{\Phi_s(\gamma_p,\Gamma_p)}{\Phi_s(\gamma,\Gamma)}.
$$

\paragraph{Proof Theorem \ref{th:1}}

Consider the test point $\bx\in \mathbb{R}^d$ and the vector $\hat{f} = \begin{bmatrix}
f(X) \\
f(\bx)
\end{bmatrix} := [\mathbf{f}~~f_{*}]$ we have 
\begin{equation*}
p(\mathbf{f},f_{*}) = \text{SkewGP}\left(\begin{bmatrix}
\xi(X) \\
\xi(\bx)
\end{bmatrix}, \begin{bmatrix}
\Omega(X,X) & \Omega(X,\bx) \\
\Omega(\bx,X) & \Omega(\bx,\bx) \\
\end{bmatrix}, \begin{bmatrix}
 \Delta(X) \\
\Delta(\bx) \\
\end{bmatrix}, \bgamma, \Gamma\right)
\end{equation*}
and the predictive distribution is by definition
\begin{align*}
p(f_{*} \mid Y) &= \int p(f_* \mid \mathbf{f}) p(\mathbf{f} \mid Y)d\mathbf{f} \\
&=  \int p(f_* \mid \mathbf{f}) \frac{p(Y \mid \mathbf{f})p(\mathbf{f})}{p(Y)}d\mathbf{f} \\
&\propto \int p(f_*, \mathbf{f}) p(Y \mid \mathbf{f})d\mathbf{f}
\end{align*}
We can then apply Lemma~\ref{lemma:Normal} with $\hat{f}$ and the likelihood $p([C \mid \mathbf{0}] \mid \hat{f}) = \phi_n\left(Y-[C \mid \mathbf{0}] \begin{bmatrix}
f(X) \\
f(\bx)
\end{bmatrix}; R\right)$ which results in a posterior distribution 
\begin{equation*}
p\left(\begin{bmatrix}
f(X) \\
f(\bx)
\end{bmatrix} \mid [Y\mid \mathbf{0}]\right) = \text{SUN}_{n+1,s}(\hat{\bxi},\hat{\Omega},\hat{\Delta},\hat{\gamma},\hat{\Gamma})
\end{equation*}
with  
\begin{align}
\hat{\bxi} &= \begin{bmatrix} \xi(X)\\
               \xi(\bx)
              \end{bmatrix}
 +\begin{bmatrix}
\Omega(X,X)  \\
\Omega(\bx,X) \\
\end{bmatrix}C^T(C\Omega C^T+R)^{-1}(Y-C\xi(X))\\
\hat{\Omega} &= \begin{bmatrix}
\Omega(X,X) & \Omega(X,\bx) \\
\Omega(\bx,X) & \Omega(\bx,\bx) \\
\end{bmatrix}-\begin{bmatrix}
\Omega(X,X)  \\
\Omega(\bx,X) \\
\end{bmatrix}C^T (C\Omega C^T+R)^{-1} C
\begin{bmatrix}
\Omega(X,X)  & \Omega(X,\bx) \\
\end{bmatrix}
\end{align}
For $\hat{\Delta}$, from Lemma \ref{lemma:Normal}, one has 
$$
\begin{aligned}
\hat{\Delta}&= D_{\hat{\Omega}}^{-1} \hat{\Omega} \begin{bmatrix}
\Omega(X,X) & \Omega(X,\bx) \\
\Omega(\bx,X) & \Omega(\bx,\bx) \\
\end{bmatrix}^{-1}  \Domega
 \begin{bmatrix}
 \Delta(X) \\
 \Delta(\bx)  \\
 \end{bmatrix}=D_{\hat{\Omega}}^{-1}  \begin{bmatrix}
A_1 & A_2 \\
A_3 & A_4 \\
\end{bmatrix}  \Domega
 \begin{bmatrix}
 \Delta(X) \\
 \Delta(\bx)  \\
 \end{bmatrix}\\
&=\begin{bmatrix}
D_{\hat{\Omega}(X,X)}^{-1}A_1D_{\Omega(X,X)}\Delta(X)+ D_{\hat{\Omega}(X,X)}^{-1}A_2 D_{\Omega(\bx,\bx)}\Delta(\bx) \\
D_{\hat{\Omega}(\bx,\bx)}^{-1}A_3 D_{\Omega(X,X)}\Delta(X)+ D_{\hat{\Omega}(\bx,\bx)}^{-1}A_4D_{\Omega(\bx,\bx)}\Delta(\bx) \\
\end{bmatrix} 
\end{aligned}
$$
where
$$
\begin{aligned}
&\begin{bmatrix}
A_1 & A_2 \\
A_3 & A_4 \\
\end{bmatrix}= \hat{\Omega} \begin{bmatrix}
\Omega(X,X) & \Omega(X,\bx) \\
\Omega(\bx,X) & \Omega(\bx,\bx) \\
\end{bmatrix}^{-1}  \\
&=\hat{\Omega}    \left[ \begin{smallmatrix}\Omega(X,X) ^{-1}+\Omega(X,X) ^{-1}\Omega(X,{\bf x}) \left(\Omega({\bf x},{\bf x}) -\Omega({\bf x},X)\Omega(X,X) ^{-1}\Omega(X,{\bf x}) \right)^{-1}\Omega({\bf x},X)\Omega(X,X) ^{-1}&-\Omega(X,X) ^{-1}\Omega(X,{\bf x}) \left(\Omega({\bf x},{\bf x}) -\Omega({\bf x},X)\Omega(X,X) ^{-1}\Omega(X,{\bf x}) \right)^{-1}\\-\left(\Omega({\bf x},{\bf x}) -\Omega({\bf x},X)\Omega(X,X) ^{-1}\Omega(X,{\bf x}) \right)^{-1}\Omega({\bf x},X)\Omega(X,X) ^{-1}&\left(\Omega({\bf x},{\bf x}) -\Omega({\bf x},X)\Omega(X,X) ^{-1}\Omega(X,{\bf x}) \right)^{-1}\end{smallmatrix}\right]
\end{aligned}
$$
Since we are interested in the marginal, we only need to compute $A_3$ and $A_4$:
$$
\begin{aligned}
A_3&=
\Big(\Omega({\bf x},X)-\Omega({\bf x},X)C^T (C\Omega(X,X) C^T+R)^{-1} C\Omega(X,X) \Big) \\&\Big(\Omega(X,X) ^{-1}+\Omega(X,X) ^{-1}\Omega(X,{\bf x}) \left(\Omega({\bf x},{\bf x}) -\Omega({\bf x},X)\Omega(X,X) ^{-1}\Omega(X,{\bf x}) \right)^{-1}\Omega({\bf x},X)\Omega(X,X) ^{-1} \Big)  \\
&-\Big(\Omega({\bf x},{\bf x})-\Omega({\bf x},X)C^T (C\Omega(X,X) C^T+R)^{-1} C\Omega(X,{\bf x}) \Big) \\&\left(\Omega({\bf x},{\bf x})-\Omega({\bf x},X)\Omega(X,X) ^{-1}\Omega(X,{\bf x}) \right)^{-1}\Omega({\bf x},X)\Omega(X,X) ^{-1} \\
&=-\Omega({\bf x},X)C^T (C\Omega(X,X) C^T+R)^{-1} C
    \end{aligned}
$$
and
$$
\begin{aligned}
A_4&=
-\Big(\Omega({\bf x},X)-\Omega({\bf x},X)C^T (C\Omega(X,X) C^T+R)^{-1} C\Omega(X,X) \Big) \\&
\Omega(X,X) ^{-1}\Omega(X,{\bf x}) \left(\Omega({\bf x},{\bf x}) -\Omega({\bf x},X)\Omega(X,X) ^{-1}\Omega(X,{\bf x}) \right)^{-1}\\
&+\Big(\Omega({\bf x},{\bf x})-\Omega({\bf x},X)C^T (C\Omega(X,X) C^T+R)^{-1} C\Omega(X,{\bf x}) \Big) \\& \left(\Omega({\bf x},{\bf x}) -\Omega({\bf x},X)\Omega(X,X) ^{-1}\Omega(X,{\bf x}) \right)^{-1}
=I_{{\bf x}}
    \end{aligned}
$$
Finally, we will show that:
$$
\begin{aligned}
\hat{\bgamma} &= \bgamma_p,\\
\hat{\Gamma} &= \Gamma_p.
\end{aligned}
$$
By Lemma \ref{lemma:Normal}
$$
\begin{aligned}
\hat{\gamma}  &= \gamma+[\Delta(X),\Delta(\bx)]\Domega\Omega^{-1}(\xi_p-\xi)\\
&= \gamma+[\Delta(X),\Delta(\bx)]\Domega\begin{bmatrix}
\Omega(X,X) & \Omega(X,\bx) \\
\Omega(\bx,X) & \Omega(\bx,\bx) \\
\end{bmatrix}^{-1}\begin{bmatrix}
\Omega(X,X)  \\
\Omega(\bx,X) \\
\end{bmatrix}C^T(C\Omega C^T+R)^{-1}(Y-C\xi(X))
\end{aligned}
$$
Note that
$$
\begin{bmatrix}
\Omega(X,X) & \Omega(X,\bx) \\
\Omega(\bx,X) & \Omega(\bx,\bx) \\
\end{bmatrix}^{-1}\begin{bmatrix}
\Omega(X,X)  \\
\Omega(\bx,X) \\
\end{bmatrix}=\begin{bmatrix}
\Omega(X,X)^{-1}  \\
0 \\
\end{bmatrix}
$$
and so $\hat{\gamma}=\gamma_p$ as in Lemma \ref{lemma:Normal}.
Similarly, one can show that $\hat{\Gamma}=\Gamma_p$.

By exploiting the marginalization properties of the SUN distribution, see Section \ref{sec:closure} and in particular  \eqref{eq:marginalFinDim}, we can derive that
\begin{align}
\label{eq:marginalization}
p\left( f(\bx) \mid C,Y,f(X) \right) = SUN_{1,s}\left(\tilde{\xi}(\bx),\tilde{\Omega}(\bx,\bx),\tilde{\Delta}(\bx),\gamma_p,\Gamma_p\right).
\end{align}
with mean, scale, and skewness functions:
	\begin{align}
	&\tilde{\bxi}({\bf x})  =\bxi({\bf x})+\Omega({\bf x},X) C^T(C\Omega(X,X) C^T+R)^{-1}(Y-C\xi(X)),\\
	&\tilde{\Omega}({\bf x},{\bf x}) = \Omega({\bf x},{\bf x})-\Omega({\bf x},X) C^T(C\Omega(X,X) C^T+R)^{-1}C\Omega(X,{\bf x}),\\
	&\tilde{\Delta}({\bf x}) =D_{\hat{\Omega}(\bx,\bx)}^{-1}D_{\Omega(\bx,\bx)}\Delta({\bf x})-D_{\hat{\Omega}(\bx,\bx)}^{-1}\Omega({\bf x},X)C^T (C\Omega(X,X) C^T+R)^{-1} C D_{\Omega(X,X)}\Delta(X).
	\end{align}
%
%

\paragraph{Lemma~\ref{lemma:Affine}}

The joint distribution of $f(X),W,Z$ is 
\begin{align}
\nonumber
 &p(W, Z|f(X))p(f(X))\\
 \nonumber
 &=\Phi_{m_a}(Z+W\bl[f];\Sigma)\;\text{SUN}_{n,s}(\bxi,\Omega,\Delta,\bgamma,\Gamma)\\
 \nonumber
 &=\Phi_{m_a}(Z+W\bl[f];\Sigma)\;\phi_n(\bl[f]-\bxi;\Omega)\frac{\Phi_s\left(\bgamma+\Delta^T\bar{\Omega}^{-1}\Domega^{-1}(\bl[f]-\bxi);\Gamma-\Delta^T\bar{\Omega}^{-1}\Delta\right)}{\Phi_s\left(\bgamma;\Gamma\right)} \\
 \label{eq:numjoint}
 &\propto \phi_n(\bl[f]-\bxi;\Omega)\Phi_{m_a}(Z+W \bl[f];\Sigma)\Phi_s\left(\bgamma+\Delta^T\bar{\Omega}^{-1}\Domega^{-1}(\bl[f]-\bxi);\Gamma-\Delta^T\bar{\Omega}^{-1}\Delta\right),
\end{align}
where we denoted $\bl[f]= f(X) \in \mathbb{R}^n$ and omitted the dependence on $X$. First, note that 
$$
\begin{aligned}
\Phi_{m_a}(Z+W\bl[f];\Sigma)= \Phi_{m_a}\left(Z+W\bxi+(\bar{\Omega}\Domega \Dmat^T)^{T}\bar{\Omega}^{-1}\Domega^{-1}(\bl[f]-\bxi);(\Dmat\Omega \Dmat^T+\Sigma)-\Dmat \Domega \bar{\Omega}\Domega \Dmat^T \right).
\end{aligned}
$$ 
Therefore, we can write
\begin{align}
\nonumber
 &\Phi_{m_a}(Z+\Dmat\bl[f];\Sigma)\Phi_s\left(\bgamma+\Delta^T\bar{\Omega}^{-1}\Domega^{-1}(\bl[f]-\bxi);\Gamma-\Delta^T\bar{\Omega}^{-1}\Delta\right) \\
 \nonumber
 &=\Phi_{m_a}\left(Z+\Dmat\bxi+(\bar{\Omega}\Domega \Dmat^T)^{T}\bar{\Omega}^{-1}\Domega^{-1}(\bl[f]-\bxi);(\Dmat\Omega \Dmat^T+\Sigma)-\Dmat \Domega \bar{\Omega}\Domega \Dmat^T \right)\\
 \nonumber
 &\cdot\Phi_s\left(\bgamma+\Delta^T\bar{\Omega}^{-1}\Domega^{-1}(\bl[f]-\bxi);\Gamma-\Delta^T\bar{\Omega}^{-1}\Delta\right) \\
 \label{eq:prodcdf}
 &=\Phi_{s+m_a}(m;M)
\end{align}
with
$$
m=\begin{bmatrix}
   \bgamma+\Delta^T\bar{\Omega}^{-1}\Domega^{-1}(\bl[f]-\bxi)\\
   Z+\Dmat\bxi+(\bar{\Omega}\Domega \Dmat^T)^{T}\bar{\Omega}^{-1}\Domega^{-1}(\bl[f]-\bxi)
  \end{bmatrix},
$$
and
$$
M=\begin{bmatrix}
  \Gamma-\Delta^T\bar{\Omega}^{-1}\Delta & 0\\
  0 & (\Dmat\Omega \Dmat^T+\Sigma)-\Dmat \Domega \bar{\Omega}\Domega \Dmat^T 
  \end{bmatrix}.
$$
From \eqref{eq:numjoint}--\eqref{eq:prodcdf} and the definition of the PDF of the SUN distribution \eqref{eq:sun}, 
we can easily show  that we can rewrite \eqref{eq:numjoint} as a SUN distribution with updated parameters:
$$
\begin{aligned}
\tilde{\xi} & =\xi,\\
\tilde{\Omega} &= \Omega, \\
\tilde{\Delta} &=[\Delta,~~\bar{\Omega}\Domega \Dmat^T],\\
\tilde{\bgamma}& =[\bgamma,~~Z+\Dmat\xi]^T, \\
\tilde{\Gamma}&=\begin{bmatrix}
         \Gamma & \quad\Delta^T \Domega \Dmat^T\\
        \Dmat \Domega \Delta & \quad (\Dmat \Omega \Dmat^T + \Sigma) \end{bmatrix}.
\end{aligned}
$$

\paragraph{Theorem \ref{th:2}}

Consider the test point $\bx\in \mathbb{R}^d$ and the vector $\hat{f} = \begin{bmatrix}
f(X) \\
f(\bx)
\end{bmatrix} := [\mathbf{f}~~f_{*}]$ we have 
\begin{equation*}
p(\mathbf{f},f_{*}) = \text{SkewGP}\left(\begin{bmatrix}
\xi(X) \\
\xi(\bx)
\end{bmatrix}, \begin{bmatrix}
\Omega(X,X) & \Omega(X,\bx) \\
\Omega(\bx,X) & \Omega(\bx,\bx) \\
\end{bmatrix}, \begin{bmatrix}
 \Delta(X) \\
\Delta(\bx) \\
\end{bmatrix}, \bgamma, \Gamma\right)
\end{equation*}
and the predictive distribution is by definition
\begin{align*}
p(f_{*} \mid W,Z) &= \int p(f_* \mid \mathbf{f}) p(\mathbf{f} \mid Z,W)d\mathbf{f} \\
&=  \int p(f_* \mid \mathbf{f}) \frac{p(Z,W \mid \mathbf{f})p(\mathbf{f})}{p(Z,W)}d\mathbf{f} \\
&\propto \int p(f_*, \mathbf{f}) p(Z,W \mid \mathbf{f})d\mathbf{f}
\end{align*}
We can then apply Lemma~\ref{lemma:Affine} with $\hat{f}$ and the likelihood $$
p(\colvec{                                                                                Z\\
\mathbf{0}                                                                              },
[W \;\; \mathbf{0}] \mid \hat{f}) = \Phi_{m_a+1}\left(\colvec{                                                                                Z\\
\mathbf{0}                                                                              }+[W \;\; \mathbf{0}] \begin{bmatrix}
f(X) \\
f(\bx)
\end{bmatrix}; \begin{bmatrix} \Sigma & {\bf 0}\\{\bf 0} & 1\end{bmatrix}\right)
$$ 
which does not depend on $f(\bx)$. This  results in a posterior distribution 
\begin{equation*}
p\left(\begin{bmatrix}
f(X) \\
f(\bx)
\end{bmatrix} \mid \colvec{                                                                                Z\\
\mathbf{0}                                                                              },[W \;\;\mathbf{0}]\right) = \text{SUN}_{n+1,m_a}(\hat{\bxi},\hat{\Omega},\hat{\Delta},\hat{\gamma},\hat{\Gamma})
\end{equation*}
with 
\begin{align*}
\hat{\bxi} &= [\xi(X)~~\xi(\bx)]^T \\
\hat{\Omega} &= \begin{bmatrix}
\Omega(X,X) & \Omega(X,\bx) \\
\Omega(\bx,X) & \Omega(\bx,\bx) \\
\end{bmatrix} \\
\hat{\Delta} &= \left[ \begin{bmatrix}
\Delta(X)  \\
\Delta(\bx) \\
\end{bmatrix}~~~~ D_{\Omega}^{-1}\begin{bmatrix}
\Omega(X,X) & \Omega(X,\bx) \\
\Omega(\bx,X) & \Omega(\bx,\bx) \\
\end{bmatrix}[W \;\; \mathbf{0}]^T \right]=\begin{bmatrix}
 \Delta(X)~~ & D_{\Omega(X,X)}^{-1}\Omega(X,X)W^T \\
\Delta(\bx)~~ & D_{\Omega(\bx,\bx)}^{-1}\Omega(\bx,X)W^T
\end{bmatrix} \\
\hat{\gamma} &= \tilde{\gamma},\\
\hat{\Gamma} &= \tilde{\Gamma},
\end{align*}
with $\tilde{\gamma},\tilde{\Gamma}$ from Lemma \ref{lemma:Affine}. By exploiting the marginalization properties of the SUN distribution, see Section \ref{sec:closure}, we obtain
\begin{align}
\label{eq:marginalization}
p\left( f(\bx) \mid W,f(X) \right) = SUN_{1,s+m_a}\left(\xi(\bx),\Omega(\bx,\bx),\begin{bmatrix}
\Delta(\bx)~ & D_{\Omega(\bx,\bx)}^{-1}\Omega(\bx,X)W^T
\end{bmatrix}, \tilde{\gamma}, \tilde{\Gamma}\right).
\end{align}

 \paragraph{Proposition~\ref{th:derivative}}
\begin{align}
 \frac{\partial}{ \partial \theta_i}\log\Phi_{m}(\tilde{\bgamma};~\tilde{\Gamma})&=
  \frac{1}{\Phi_{m}(\tilde{\bgamma};~\tilde{\Gamma})}
\iint_{-\infty}^{\tilde{\bgamma}}  \frac{\partial}{ \partial \theta_i}\left( \frac{1}{|2 \pi {\Gamma}|}e^{-\frac{1}{2}{\bf z}^T \tilde{\Gamma}^{-1} {\bf z}} d{\bf z}\right)\\
 &=-\frac{1}{2}\frac{1}{\Phi_{m}(\tilde{\bgamma};~\tilde{\Gamma})} \text{Tr}\left( \tilde{\Gamma}^{-1}\tilde{\Gamma}'_{\theta_i}\right)\iint_{-\infty}^{\tilde{\bgamma}}  \frac{1}{|2 \pi {\Gamma}|}e^{-\frac{1}{2}{\bf z}^T \tilde{\Gamma}^{-1} {\bf z}} d{\bf z}\\
&+\frac{1}{2}\frac{1}{\Phi_{m}(\tilde{\bgamma};~\tilde{\Gamma})} \iint_{-\infty}^{\tilde{\bgamma}} \text{Tr}\left(\tilde{\Gamma}^{-1}\tilde{\Gamma}'_{\theta_i}\tilde{\Gamma}^{-1} {\bf z}{\bf z}^T\right) \frac{1}{|2 \pi {\Gamma}|}e^{-\frac{1}{2}{\bf z}^T \tilde{\Gamma}^{-1} {\bf z}} d{\bf z}\\
 &=-\frac{1}{2}\text{Tr}\left( \tilde{\Gamma}^{-1}\tilde{\Gamma}'_{\theta_i}\right)+\frac{1}{2}\text{Tr}\left(\tilde{\Gamma}^{-1}\tilde{\Gamma}'_{\theta_i}\tilde{\Gamma}^{-1} N\right),
\end{align}

 where $\tilde{\Gamma}'_{\theta_i}=\frac{\partial}{ \partial \theta_i}\tilde{\Gamma}$
and $N= \frac{1}{\Phi_{m}(\tilde{\bgamma};~\tilde{\Gamma})}\iint_{-\infty}^{\tilde{\bgamma}} {\bf z}{\bf z}^T \frac{1}{|2 \pi {\Gamma}|}e^{-\frac{1}{2}{\bf z}^T \tilde{\Gamma}^{-1} {\bf z}} d{\bf z}$.
In the derivations, we have exploited that $\frac{\partial}{ \partial \theta_i} |\tilde{\Gamma}|= |\tilde{\Gamma}| Tr(\tilde{\Gamma}^{-1} \frac{\partial}{ \partial \theta_i}\tilde{\Gamma})$ and that $\frac{\partial}{ \partial \theta_i}\tilde{\Gamma}^{-1}=-\tilde{\Gamma}^{-1} \left(\frac{\partial}{ \partial \theta_i}\tilde{\Gamma}\right)\tilde{\Gamma}^{-1}$.

\bibliographystyle{apalike}
\bibliography{biblio}
\end{document}